\documentclass[10pt,twocolumn,letterpaper]{article}

\usepackage[utf8]{inputenc}
\usepackage{cvpr}
\usepackage{times}
\usepackage{graphicx}
\usepackage{amsmath,amsfonts,amssymb,amsthm}
\usepackage{xcolor}
\usepackage{caption}
\usepackage{subcaption}
\usepackage{algorithm}
\usepackage[noend]{algpseudocode}
\usepackage{xspace}
\usepackage{array}
\usepackage{cite}
\usepackage{soul}
\usepackage[pagebackref=true,breaklinks=true,letterpaper=true,colorlinks,bookmarks=false]{hyperref}
\usepackage{booktabs}
\usepackage{float}

\cvprfinalcopy
\def\cvprPaperID{1319}

\ifcvprfinal\fi

\graphicspath{ {gfx/} }

\newcommand{\figref}[1]{\Fig~\ref{#1}}
\newcommand{\secref}[1]{Section~\ref{#1}}

\newcommand{\eqnref}[1]{Eq.~\eqref{#1}}
\newcommand{\tabref}[1]{Table~\ref{#1}}

\makeatletter
\DeclareRobustCommand\onedot{\futurelet\@let@token\@onedot}
\def\@onedot{\ifx\@let@token.\else.\null\fi\xspace}
\def\eg{e.g\onedot} 
\def\ie{i.e\onedot}

\def\wrt{wrt\onedot}

\def\etal{et~al\onedot} 
\def\Fig{Fig\onedot}   
\makeatother

\newcommand{\boldparagraph}[1]{\vspace{0.2cm}\noindent{\bf #1:} }

\definecolor{darkgreen}{rgb}{0,0.7,0}

\DeclareMathOperator{\modulo}{mod~}
\DeclareMathOperator*{\pool}{pool\_voxels~}

\DeclareMathOperator{\octreeparent}{pa}
\DeclareMathOperator{\octreechild}{ch}
\DeclareMathOperator{\octreedataidx}{data\_idx}
\DeclareMathOperator{\octreeisset}{bit}

\DeclareMathOperator{\ocvoxeldepth}{vxd}

\DeclareMathOperator{\octreetotensor}{oc2ten}
\DeclareMathOperator{\tensortooctree}{ten2oc}

\def\Tnsr{T}
\def\Octr{O}
\def\Wgts{W}

\def\ocgriddepth{D}
\def\ocgridheight{H}
\def\ocgridwidth{W}

\begin{document}

\title{OctNet: Learning Deep 3D Representations at High Resolutions}

\author{Gernot Riegler$^1$ \qquad Ali Osman Ulusoy$^2$ \qquad Andreas Geiger$^{2,3}$\\
$^1$Institute for Computer Graphics and Vision, Graz University of Technology\\%
$^2$Autonomous Vision Group, MPI for Intelligent Systems T\"ubingen\\%
$^3$Computer Vision and Geometry Group, ETH Zürich\\%
{\tt\small riegler@icg.tugraz.at} \quad {\tt\small \{osman.ulusoy,andreas.geiger\}@tue.mpg.de}
}

\maketitle

\setlength\arraycolsep{1.5pt}

\begin{abstract}
We present OctNet, a representation for deep learning with sparse 3D data. In contrast to existing models, our representation enables 3D convolutional networks which are both deep and high resolution. Towards this goal, we exploit the sparsity in the input data to hierarchically partition the space using a set of unbalanced octrees where each leaf node stores a pooled feature representation. This allows to focus memory allocation and computation to the relevant dense regions and enables deeper networks without compromising resolution. We demonstrate the utility of our OctNet representation by analyzing the impact of resolution on several 3D tasks including 3D object classification, orientation estimation and point cloud labeling.
\end{abstract}

\vspace{-6pt}
\section{Introduction}
\label{sec:introduction}

Over the last several years, convolutional networks have lead to substantial performance gains in many areas of computer vision. 
In most of these cases, the input to the network is of two-dimensional nature, \eg, in image classification \cite{He2016CVPR}, object detection \cite{Ren2015NIPS} or semantic segmentation \cite{Ghiasi2016ECCV}. 
However, recent advances in 3D reconstruction \cite{Newcombe2011ISMAR} and graphics \cite{Huang2015SIGGRAPH} allow capturing and modeling large amounts of 3D data.
At the same time, large 3D repositories such as ModelNet \cite{Wu2015CVPR}, ShapeNet \cite{Chang2015ARXIV} or 3D Warehouse\footnote{\url{https://3dwarehouse.sketchup.com}} as well as databases of 3D object scans \cite{Choi2016ARXIV} are becoming increasingly available. 
These factors have motivated the development of convolutional networks that operate on 3D data.

\begin{figure}[t!]
  \rotatebox{90}{\hspace{1.2cm} \small{OctNet} \hspace{1cm} \small{Dense 3D ConvNet} \hspace{0.2cm} \small{Dense 3D ConvNet}}~
\begin{subfigure}[b]{0.31\linewidth}
  \includegraphics[width=\linewidth]{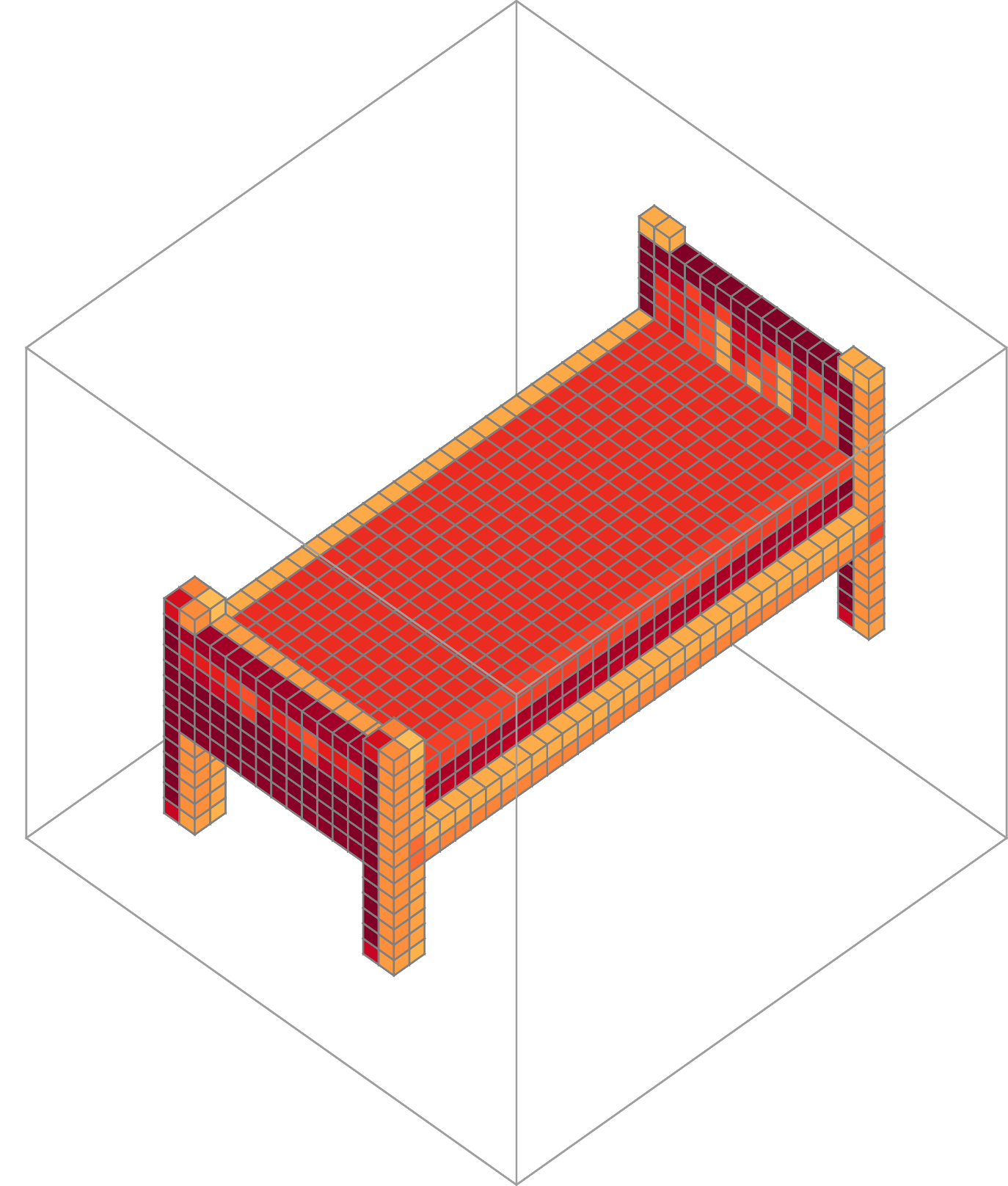}\vspace{0.1cm}
  \includegraphics[width=\linewidth]{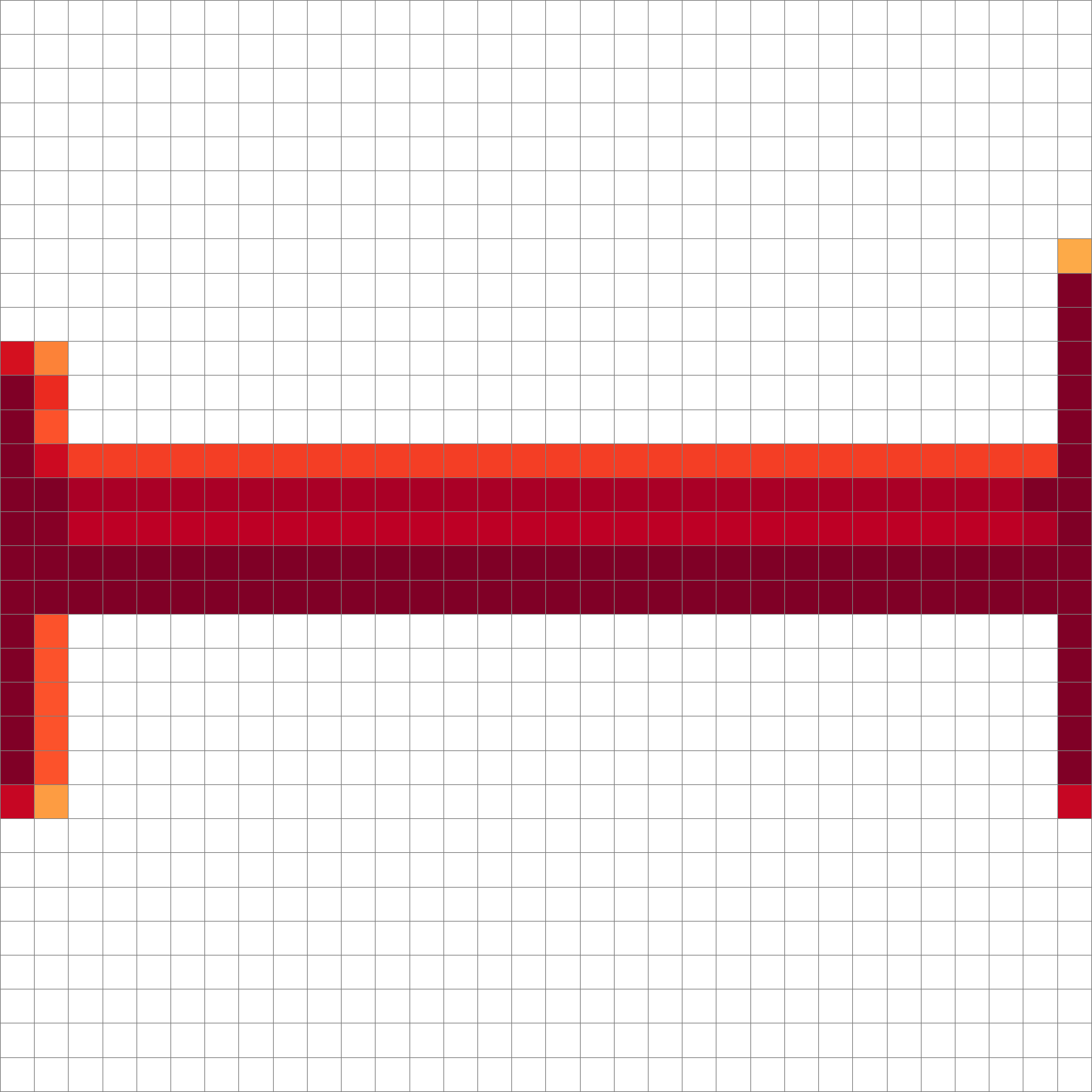}\vspace{0.1cm}
  \includegraphics[width=\linewidth]{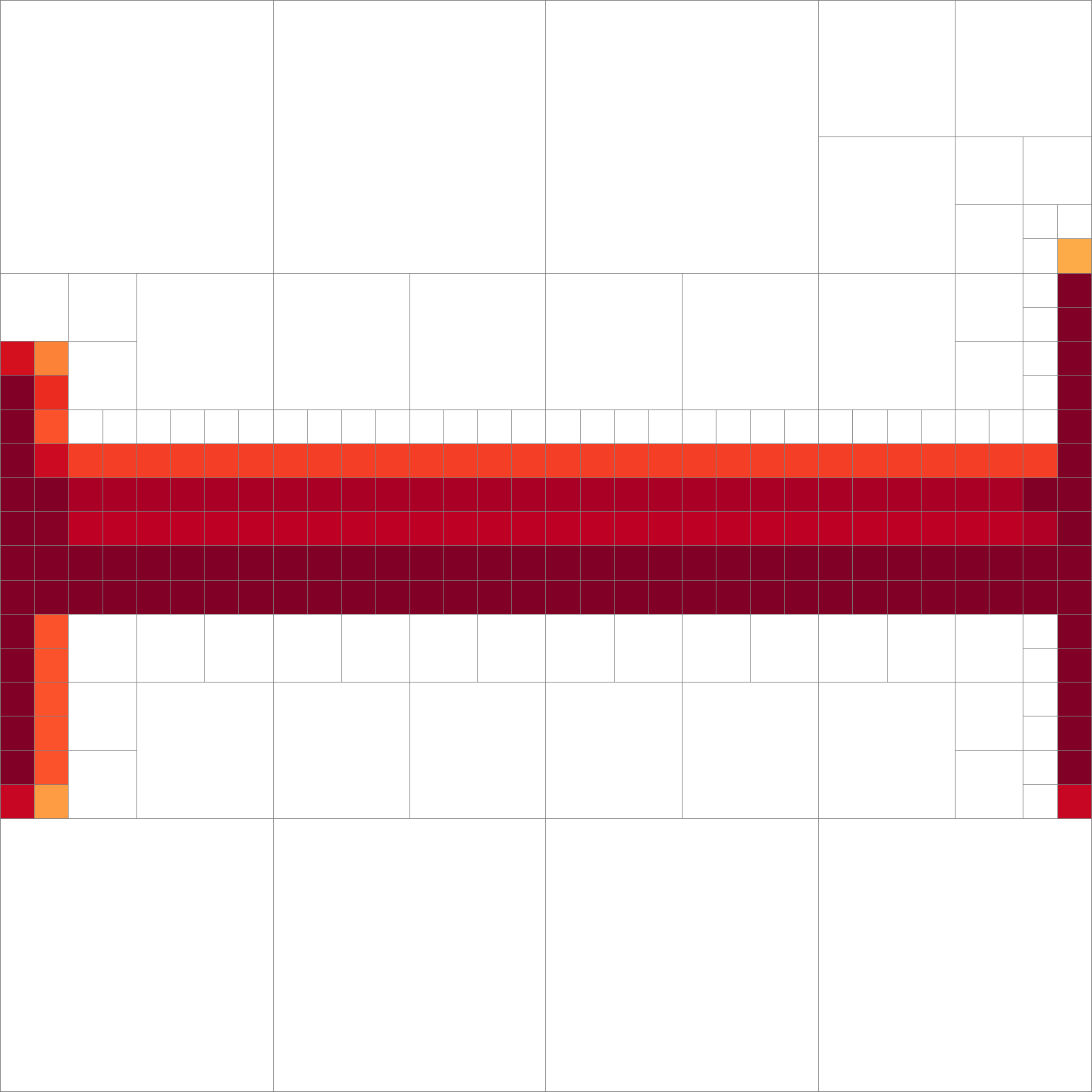}
  \caption{Layer 1: $32^3$}
\end{subfigure}
\begin{subfigure}[b]{0.31\linewidth}
  \includegraphics[width=\linewidth]{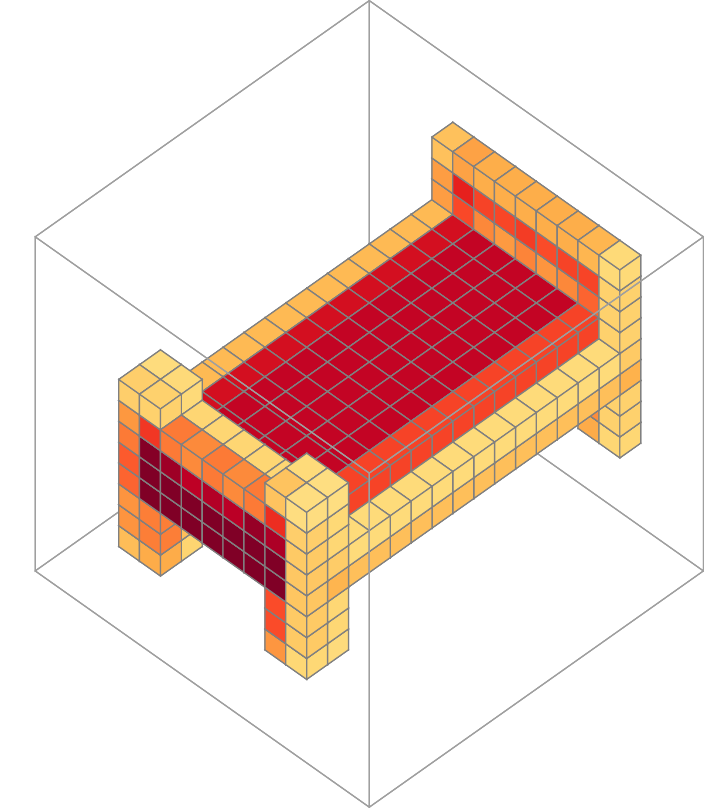}\vspace{0.1cm}
  \includegraphics[width=\linewidth]{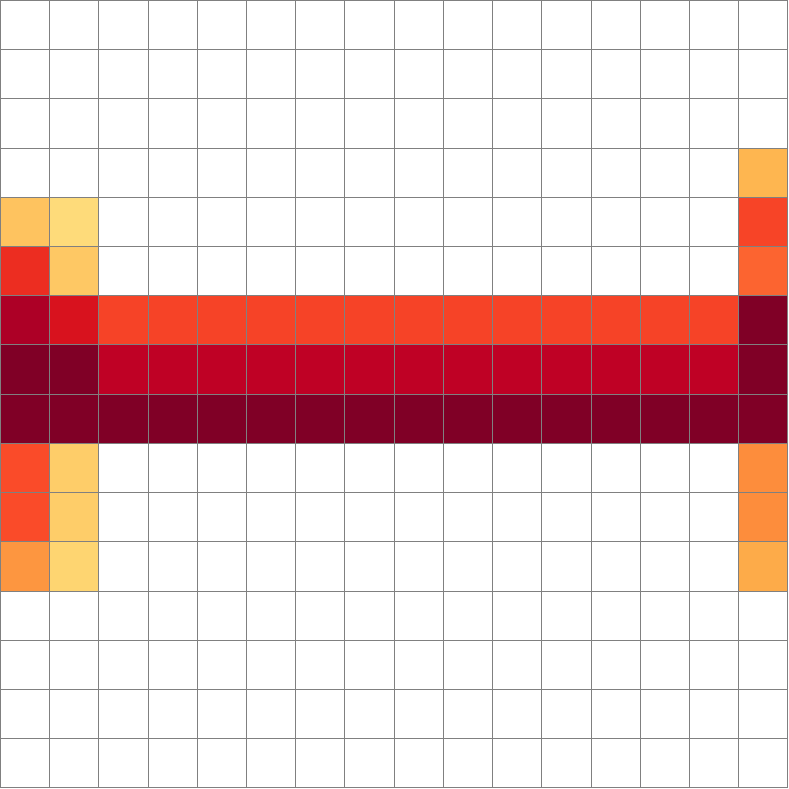}\vspace{0.1cm}
  \includegraphics[width=\linewidth]{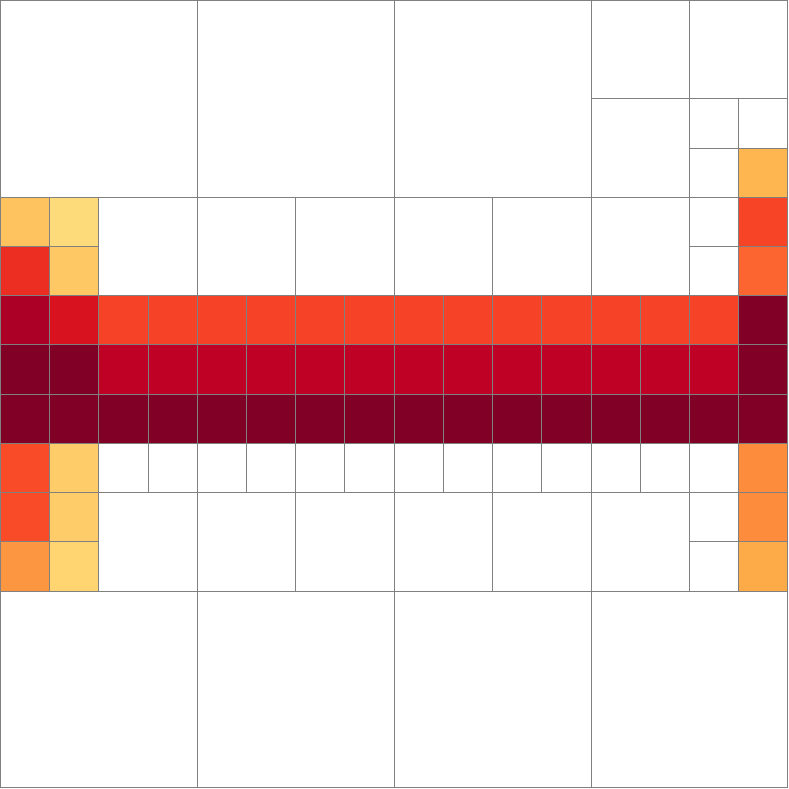}
  \caption{Layer 2: $16^3$}
\end{subfigure}
\begin{subfigure}[b]{0.31\linewidth}
  \includegraphics[width=\linewidth]{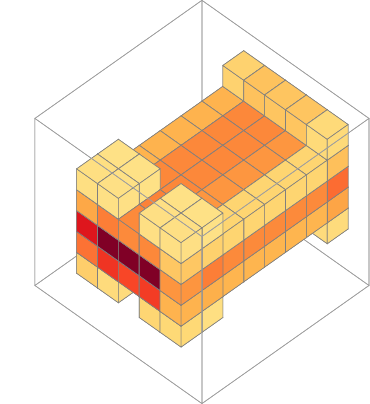}\vspace{0.1cm}
  \includegraphics[width=\linewidth]{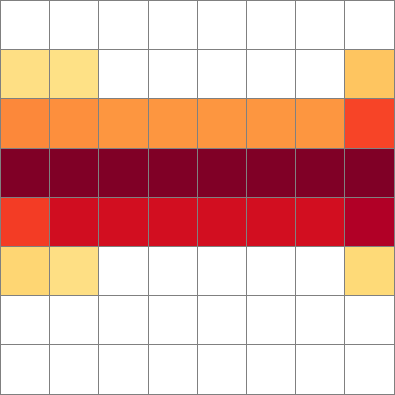}\vspace{0.1cm}
  \includegraphics[width=\linewidth]{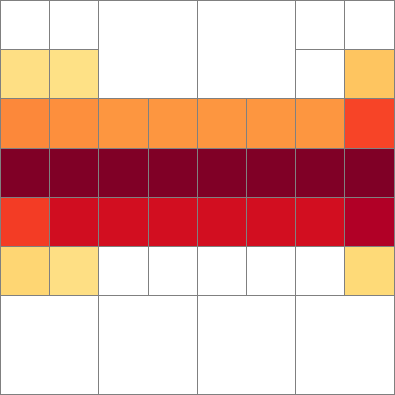}
  \caption{Layer 3: $8^3$}
\end{subfigure}
\vspace{-18pt}
\caption{{\bf Motivation.}
For illustration purposes, we trained a dense convolutional network to classify 3D shapes from~\cite{Wu2015CVPR}. 
Given a voxelized bed as input, we show the maximum response across all feature maps at intermediate layers (a-c) of the network before pooling.
Higher activations are indicated with darker colors.
Voxels with zero activation are not displayed.
The first row visualizes the responses in 3D while the second row shows a 2D slice.
Note how voxels close to the object contour respond more strongly than voxels further away. 
We exploit the sparsity in our data by allocating memory and computations using a space partitioning data structure (bottom row).
}
\label{fig:sparse_activations}
\vspace{-18pt}
\end{figure}

Most existing 3D network architectures \cite{Choy2016ECCV,Maturana2015IROS,Qi2016CVPR,Wu2015CVPR} replace the 2D pixel array by its 3D analogue, \ie, a dense and regular 3D voxel grid, and process this grid using 3D convolution and pooling operations.
However, for dense 3D data, computational and memory requirements grow \emph{cubically} with the resolution. 
Consequently, existing 3D networks %
are limited to low 3D resolutions, typically in the order of $30^3$ voxels.
To fully exploit the rich and detailed geometry of our 3D world, however, much higher resolution networks are required.

In this work, we build on the observation that 3D data is often sparse in nature, \eg, point clouds, or meshes, resulting in wasted computations when applying 3D convolutions na\"{i}vely. 
We illustrate this in \figref{fig:sparse_activations} for a 3D classification example.
Given the 3D meshes of~\cite{Wu2015CVPR} we voxelize the input at a resolution of $64^3$ and train a simple 3D convolutional network to minimize a classification loss.
We depict the maximum of the responses across all feature maps at different layers of the network.
It is easy to observe that high activations occur only near the object boundaries.

Motivated by this observation, we propose {\it OctNet}, a 3D convolutional network that exploits this sparsity property. Our OctNet hierarchically partitions the 3D space into a set of unbalanced octrees \cite{Miller2011GPGPU}.
Each octree splits the 3D space according to the density of the data.
More specifically, we recursively split octree nodes that contain a data point in its domain, \ie, 3D points, or mesh triangles, stopping at the finest resolution of the tree.
Therefore, leaf nodes vary in size, \eg, an empty leaf node may comprise up to $8^3=512$ voxels for a tree of depth 3 and each leaf node in the octree stores a pooled summary of all feature activations of the voxel it comprises.
The convolutional network operations are directly defined on the structure of these trees.
Therefore, our network dynamically focuses computational and memory resources, depending on the 3D structure of the input.
This leads to a significant reduction in computational and memory requirements which allows for deep learning at high resolutions.
Importantly, we also show how essential network operations (convolution, pooling or unpooling) can be efficiently implemented on this new data structure.

We demonstrate the utility of the proposed OctNet on three different problems involving three-dimensional data: 
3D classification, 3D orientation estimation of unknown object instances and semantic segmentation of 3D point clouds. 
In particular, we show that the proposed OctNet enables significant higher input resolutions compared to dense inputs due to its lower memory consumption, while achieving identical performance compared to the equivalent dense network at lower resolutions.
At the same time we gain significant speed-ups at resolutions of $128^3$ and above.
Using our OctNet, we investigate the impact of high resolution inputs \wrt accuracy on the three tasks and demonstrate that higher resolutions are particularly beneficial for orientation estimation and semantic point cloud labeling.
Our code is available from the project website\footnote{\url{https://github.com/griegler/octnet}}.

\section{Related Work}
\label{sec:related}

While 2D convolutional networks have proven very successful in extracting information from images \cite{He2016CVPR,Ren2015NIPS,Ghiasi2016ECCV,Zamir2016ECCV,
Tatarchenko2016ECCV,Xie2016ECCV,Su2015ICCV,Flynn2015ARXIV,Wu2016ECCV}, there exists comparably little work on processing three-dimensional data. In this Section, we review existing work on dense and sparse models.

\boldparagraph{Dense Models}
Wu \etal \cite{Wu2015CVPR} trained a deep belief network on shapes discretized to a $30^3$ voxel grid for object classification, shape completion and next best view prediction.
Maturana \etal \cite{Maturana2015IROS} proposed VoxNet, a feed-forward convolutional network for classifying $32^3$  voxel volumes from \mbox{RGB-D} data.
In follow-up work, Sedaghat \etal \cite{Sedaghat2016ARXIV} showed that introducing an auxiliary orientation loss increases classification performance over the original VoxNet.
Similar models have also been exploited for semantic point cloud labeling \cite{Huang2016ICPR} and scene context has been integrated in \cite{Zhang2016ARXIV}.

Recently, generative models \cite{Rezende2016ARXIV} and auto-encoders \cite{Sharma2016ARXIV,Brock2016ARXIV} have demonstrated impressive performance in learning low-dimensional object representations from collections of low-resolution ($32^3$) 3D shapes. 
Interestingly, these low-dimensional representations can be directly inferred from a single image \cite{Girdhar2016ECCV} or a sequence of images \cite{Choy2016ECCV}.

Due to computational and memory limitations, all aforementioned methods are only able to process and generate shapes at a very coarse resolution, typically in the order of $30^3$ voxels. 
Besides, when high-resolution outputs are desired, \eg, for labeling 3D point clouds, inefficient sliding-window techniques with a limited receptive field must be adopted \cite{Huang2016ICPR}.
Increasing the resolution na\"{i}vely \cite{Song2015ARXIV,Cicek2016ARXIV,Milletari2016ARXIV} reduces the depth of the networks and hence their expressiveness.
In contrast, the proposed OctNets allow for training deep architectures at significant higher resolutions.

\boldparagraph{Sparse Models}
There exist only few network architectures which explicitly exploit sparsity in the data.
As these networks do not require exhaustive dense convolutions they have the potential of handling higher resolutions.

Engelcke \etal \cite{Engelcke2016ARXIV} proposed to calculate convolutions at sparse input locations by pushing values to their target locations. This has the potential to reduce the number of convolutions but does not reduce the amount of memory required. Consequently, their work considers only very shallow networks with up to three layers.

A similar approach is presented in \cite{Graham2014ARXIV,Graham2015BMVC} where sparse convolutions are reduced to matrix operations.
Unfortunately, the model only allows for $2\times 2$ convolutions and results in indexing and copy overhead which prevents processing volumes of larger resolution (the maximum resolution considered in \cite{Graham2014ARXIV,Graham2015BMVC}  is $80^3$ voxels). Besides, each layer decreases sparsity and thus increases the number of operations, even at a single resolution. In contrast, the number of operations remains constant in our model.

Li \etal \cite{Li2016ARXIV} proposed field probing networks which sample 3D data at sparse points before feeding them into fully connected layers. While this reduces memory and computation, it does not allow for exploiting the distributed computational power of convolutional networks as field probing layers can not be stacked, convolved or pooled.

Jampani \etal \cite{Jampani2016CVPR} introduced bilateral convolution layers (BCL) which map sparse inputs into permutohedral space where learnt convolutional filters are applied. 
Their work is related to ours with respect to efficiently exploiting the sparsity in the input data.
However, in contrast to BCL our method is specifically targeted at 3D convolutional networks and can be immediately dropped in as a replacement in existing network architectures.

\section{Octree Networks}
\label{sec:method}

To decrease the memory footprint of convolutional networks operating on sparse 3D data, we propose an adaptive space partitioning scheme which focuses computations on the relevant regions.
As mathematical operations of deep networks, especially convolutional networks, are best understood on regular grids, we restrict our attention to data structures on 3D voxel grids.
One of the most popular space partitioning structures on voxel grids are octrees~\cite{Meagher1982CGIP} which have been widely adopted due to their flexible and hierarchical structure. Areas of application include depth fusion \cite{Kehl2016ARXIV}, image rendering \cite{Laine2011VCG} and 3D reconstruction \cite{Ulusoy2015THREEDV}.
In this paper, we propose 3D convolutional networks on octrees to learn representations from high resolution 3D data.

An octree partitions the 3D space by recursively subdividing it into octants.
By subdividing only the cells which contain relevant information (\eg, cells crossing a surface boundary or cells containing one or more 3D points) storage can be allocated adaptively. Densely populated regions are modeled with high accuracy (\ie, using small cells) while empty regions are summarized by large cells in the octree.

Unfortunately, vanilla octree implementations \cite{Meagher1982CGIP} have several drawbacks that hamper its application in deep networks. 
While octrees reduce the memory footprint of the 3D representation, most versions do not allow for efficient access to the underlying data. 
In particular, octrees are typically implemented using pointers, where each node contains a pointer to its children. 
Accessing an arbitrary element (or the neighbor of an element) in the octree requires a traversal starting from the root until the desired cell is reached.
Thus, the number of memory accesses is equal to the depth of the tree.
This becomes increasingly costly for deep, \ie, high-resolution, octrees.
Convolutional network operations such as convolution or pooling require frequent access to neighboring elements.
It is thus critical to utilize an octree design that allows for fast data access.

We tackle these challenges by leveraging a hybrid grid-octree data structure which we describe in \secref{sec:hybrid_grid_octree}. In \secref{sec:network_operations}, we show how 3D convolution and pooling operations can be implemented efficiently on this data structure.

\subsection{Hybrid Grid-Octree Data Structure}
\label{sec:hybrid_grid_octree}

\begin{figure}
\centering
\begin{subfigure}[b]{0.23\textwidth}
\includegraphics[width=\textwidth]{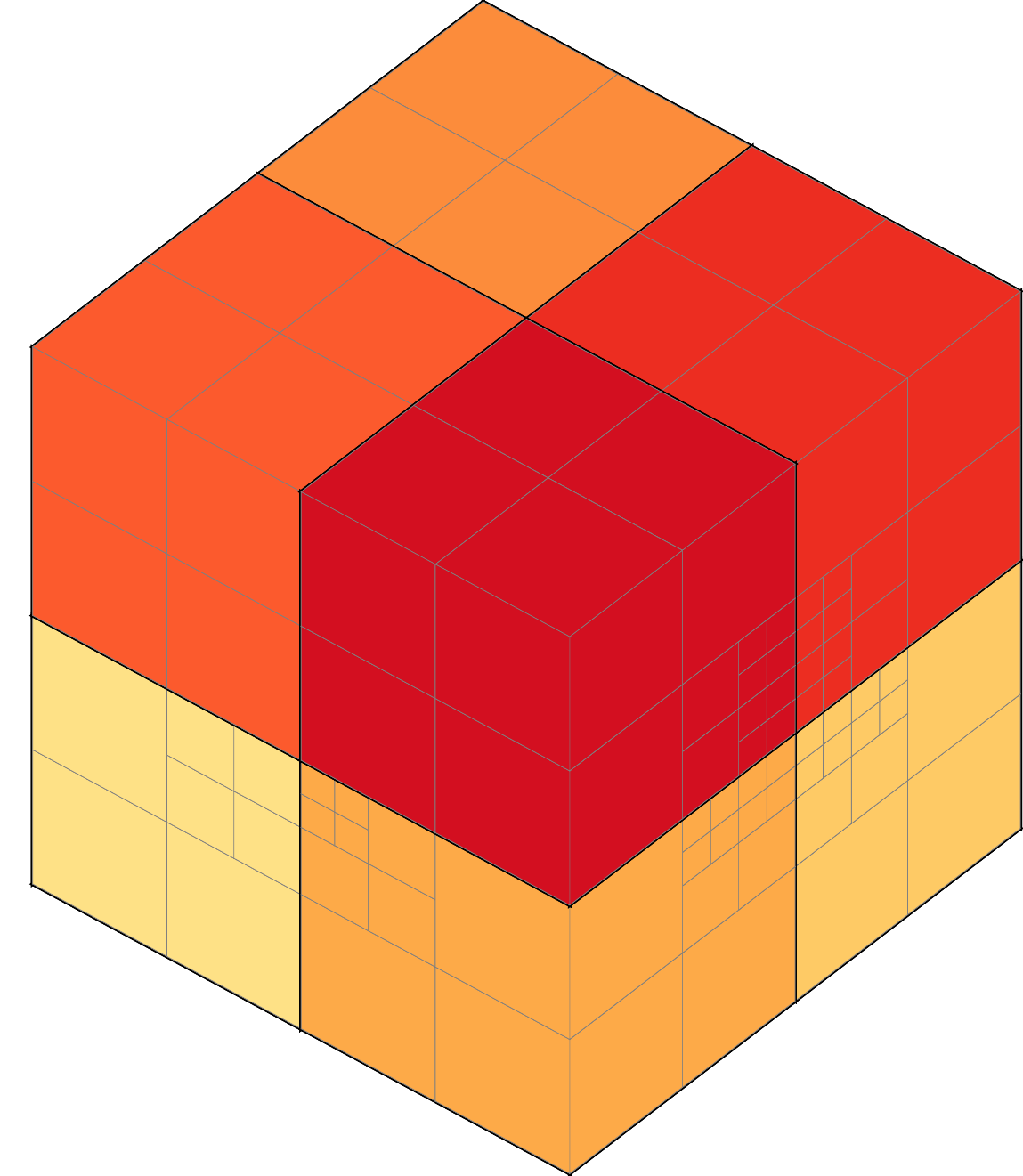}
\end{subfigure}
\begin{subfigure}[b]{0.23\textwidth}
\includegraphics[width=\textwidth]{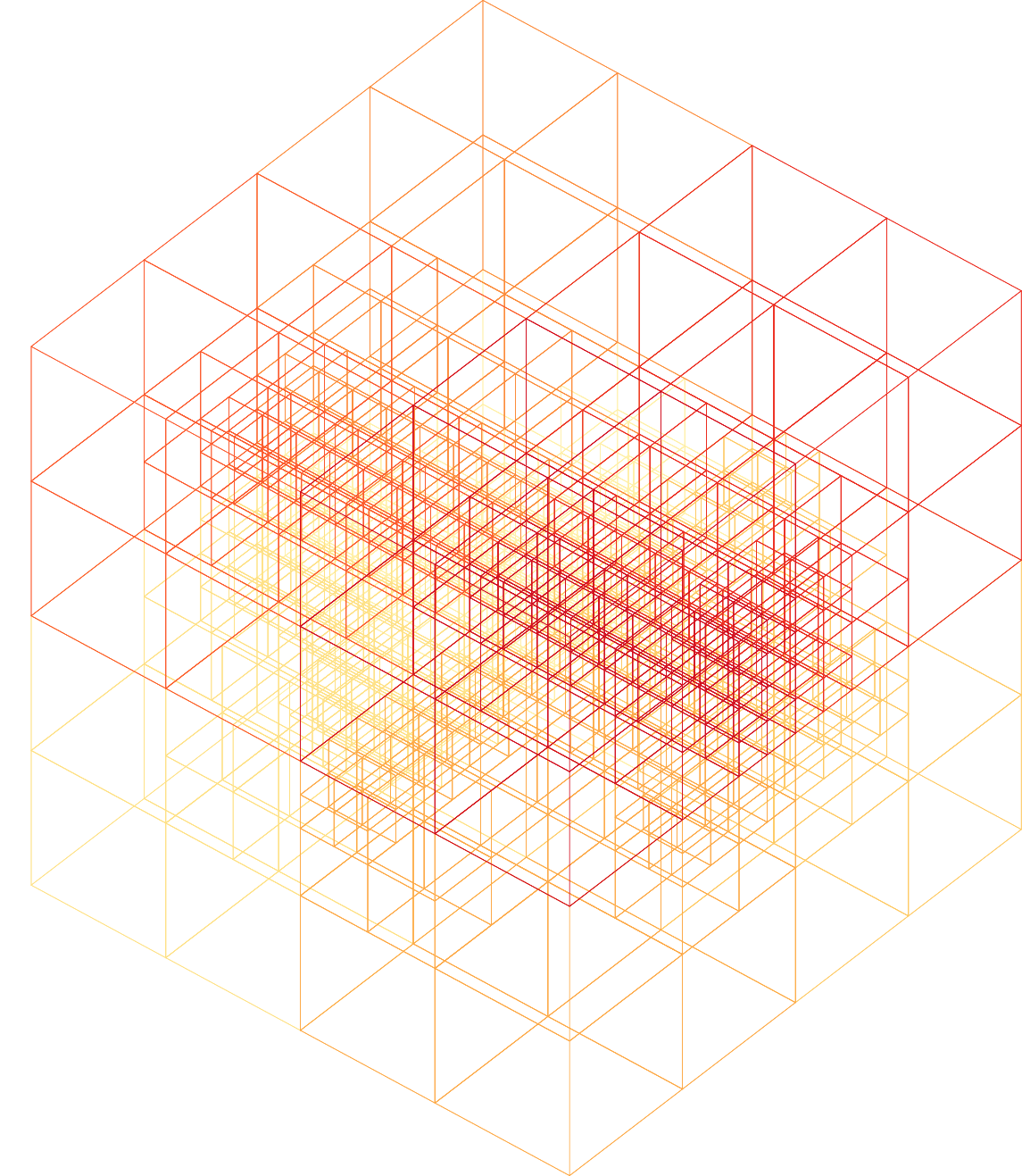}
\end{subfigure}
\caption{{\bf Hybrid Grid-Octree Data Structure.} This example illustrates a hybrid grid-octree consisting of $8$ shallow octrees indicated by different colors. Using $2$ shallow octrees in each dimension with a maximum depth of $3$ leads to a total resolution of $16^3$ voxels.}
\label{fig:hybrid_grid_octree}
\vspace{-0.3cm}
\end{figure}

\begin{figure}
  \centering
\definecolor{bit_color_0}{RGB}{254,202,101}
\definecolor{bit_color_1}{RGB}{252,140,59}
\definecolor{bit_color_2}{RGB}{236,45,33}
  \begin{subfigure}[b]{0.15\textwidth}
    \includegraphics[width=\textwidth]{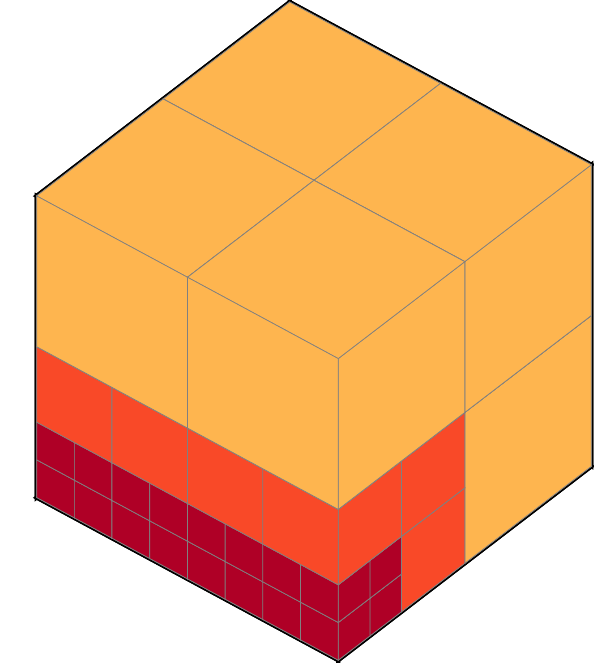}
    \caption{Shallow Octree}
  \end{subfigure}
  \begin{subfigure}[b]{0.28\textwidth}
    \includegraphics[width=\textwidth]{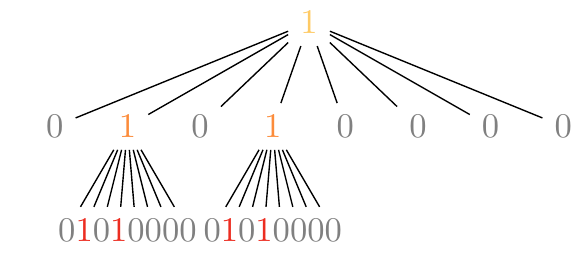}
    \caption{Bit-Representation}
  \end{subfigure}
  \caption{
    {\bf Bit Representation.} Shallow octrees can be efficiently encoded using bit-strings.
    Here, the bit-string\\ {{\color{bit_color_0}1}\hspace{0.3em}{\color{gray}0}{\color{bit_color_1}1}{\color{gray}0}{\color{bit_color_1}1}{\color{gray}0}{\color{gray}0}{\color{gray}0}{\color{gray}0}\hspace{0.3em}{\color{gray}0}{\color{gray}0}{\color{gray}0}{\color{gray}0}{\color{gray}0}{\color{gray}0}{\color{gray}0}{\color{gray}0}\hspace{0.15em}{\color{gray}0}{\color{bit_color_2}1}{\color{gray}0}{\color{bit_color_2}1}{\color{gray}0}{\color{gray}0}{\color{gray}0}{\color{gray}0}\hspace{0.15em}{\color{gray}0}{\color{gray}0}{\color{gray}0}{\color{gray}0}{\color{gray}0}{\color{gray}0}{\color{gray}0}{\color{gray}0}\hspace{0.15em}{\color{gray}0}{\color{bit_color_2}1}{\color{gray}0}{\color{bit_color_2}1}{\color{gray}0}{\color{gray}0}{\color{gray}0}{\color{gray}0}\hspace{0.15em}{\color{gray}0...}} defines the octree in (a). 
    The corresponding tree is shown in (b). 
    The color of the voxels corresponds to the split level.
  }
  \label{fig:octree_bit_representation}
  \vspace{-0.5cm}
\end{figure}

The above mentioned problems with the vanilla octree data structure increase with the octree depth.
Instead of representing the entire high resolution 3D input with a single unbalanced octree, we leverage a hybrid grid-octree structure similar to the one proposed by Miller \etal \cite{Miller2011GPGPU}.
The key idea is to restrict the maximal depth of an octree to a small number, \eg, three, and place several such shallow octrees along a regular grid (\figref{fig:hybrid_grid_octree}).
While this data structure may not be as memory efficient as the standard octree, significant compression ratios can still be achieved.
For instance, a single shallow octree that does not contain input data stores only a single vector, instead of $8^3=512$ vectors for all voxels at the finest resolution at depth $3$.

An additional benefit of a collection of shallow octrees is that their structure can be encoded very efficiently using a bit string representation which further lowers access time and allows for efficient GPGPU implementations \cite{Miller2011GPGPU}.
Given a shallow octree of depth $3$, we use $73$ bit to represent the complete tree. 
The first bit with index $0$ indicates, if the root node is split, or not.
Further, bits $1$ to $8$ indicate if one of the child nodes is subdivided and bits $9$ to $72$ denote splits of the grandchildren, see \figref{fig:octree_bit_representation}.
A tree depth of $3$ gives a good trade-off between memory consumption and computational efficiency. Increasing the octree depth results in an exponential growth in the required bits to store the tree structure and further increases the cell traversal time.

Using this bit-representation, a single voxel in the shallow octree is fully characterised by its bit index.
This index determines the depth of the voxel in the octree and therefore also the voxel size.
Instead of using pointers to the parent and child nodes, simple arithmetic can be used to retrieve the corresponding indices of a voxel with bit index $i$:
\begin{align}
  \octreeparent(i) &= \left\lfloor\frac{i - 1}{8}\right\rfloor \,, \\
  \octreechild(i) &= 8 \cdot i + 1 \,.
\end{align}
In contrast to \cite{Miller2011GPGPU}, we associate a data container (for storing features vectors) with all leaf nodes of each shallow tree. 
We allocate the data of a shallow octree in a contiguous data array.
The offset associated with a particular voxel in this array can be computed as follows:
\begin{align}
\begin{split}
\octreedataidx(i) = & \underbrace{8 \sum_{j=0}^{\octreeparent(i)-1} \octreeisset(j) + 1}_{\text{\#nodes above i}} - \underbrace{\sum_{j=0}^{i-1} \octreeisset(j)}_{\text{\#split nodes pre i}} \\
&+ \underbrace{\modulo(i-1, 8)}_{\text{offset}} \,.
\end{split}
\end{align}
Here, $\modulo$ denotes the modulo operator and $\octreeisset$ returns the tree bit-string value at $i$. See supp.\ document for an example.
Both sum operations can be efficiently implemented using bit counting intrinsics (\texttt{popcnt}). 
The data arrays of all shallow octrees are concatenated into a single contiguous data array during training and testing to reduce I/O latency.

\subsection{Network Operations}
\label{sec:network_operations}

Given the hybrid grid-octree data structure introduced in the previous Section, we now discuss the efficient implementation of network operations on this data structure.
We will focus on the most common operations in convolutional networks~\cite{He2016CVPR,Ren2015NIPS,Ghiasi2016ECCV}: convolution, pooling and unpooling.
Note that point-wise operations, like activation functions, do not differ in their implementation as they are independent of the data structure.

Let us first introduce the notation which will be used throughout this Section.
$\Tnsr_{i,j,k}$ denotes the value of a 3D tensor $\Tnsr$ at location $(i,j,k)$.
Now assume a hybrid grid-octree structure with $\ocgriddepth \times \ocgridheight \times \ocgridwidth$ unbalanced shallow octrees of maximum depth $3$.
Let $\Octr[i,j,k]$ denote the value of the smallest cell in this structure which comprises the voxel $(i,j,k)$.
Note that in contrast to the tensor notation, $\Octr[i_1,j_1,k_1]$ and $\Octr[i_2,j_2,k_2]$ with $i_1 \neq i_2 \vee j_1 \neq j_2 \vee k_1 \neq k_2$ may refer to the same voxel in the hybrid grid-octree, depending on the size of the voxels.
We obtain the index of the shallow octree in the grid via $(\lfloor \tfrac{i}{8} \rfloor, \lfloor \tfrac{j}{8} \rfloor, \lfloor \tfrac{k}{8} \rfloor)$ and the local index of the voxel at the finest resolution in that octree by $(\modulo(i,8), \modulo(j,8), \modulo(k,8))$.

Given this notation, the mapping from a grid-octree $\Octr$ to a tensor $\Tnsr$ with compatible dimensions is given by
\begin{align}
\octreetotensor: \Tnsr_{i,j,k} = \Octr[i,j,k] \,. %
\end{align}
Similarly, the reverse mapping is given by
\begin{align}
\tensortooctree: \Octr[i,j,k] = \pool_{(\bar{i},\bar{j},\bar{k})\in \Omega[{i,j,k}]}(\Tnsr_{\bar{i},\bar{j},\bar{k}}) \,,
\end{align} %
where $\pool(\cdot)$ is a pooling function (\eg, average- or max-pooling) which pools all voxels in $T$ over the smallest grid-octree cell comprising location $(i,j,k)$, denoted by $\Omega[{i,j,k}]$.
This pooling is necessary as a single voxel in $\Octr$ can cover up to $8^3=512$ elements of $\Tnsr$, depending on its size $|\Omega[{i,j,k}]|$. 

\noindent {\bf Remark: }With the two functions defined above, we could wrap any network operation $f$ defined on 3D tensors via
\begin{align}
g(\Octr) = \tensortooctree(f(\octreetotensor(\Octr))) \,.
\end{align}
However, this would require a costly conversion from the memory efficient grid-octrees to a regular 3D tensor and back.
Besides, storing a dense tensor in memory limits the maximal resolution.
We therefore define our network operations directly on the hybrid grid-octree data structure.

\paragraph{Convolution}
\begin{figure}
\centering
\begin{subfigure}[b]{0.23\textwidth}
\includegraphics[width=\textwidth]{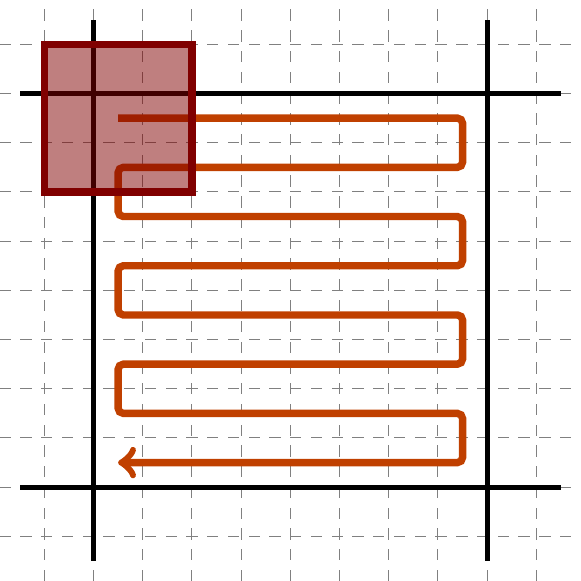}
\caption{Standard Convolution}
\end{subfigure}
\begin{subfigure}[b]{0.23\textwidth}
\includegraphics[width=\textwidth]{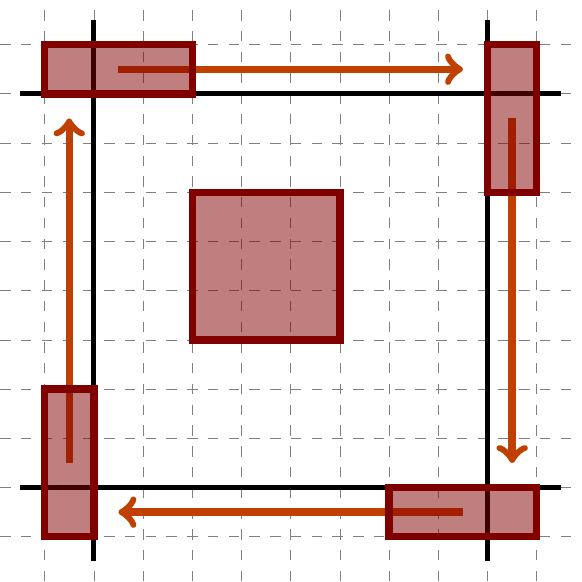}
\caption{Efficient Convolution}
\end{subfigure}
\caption{{\bf Convolution.}
This figure illustrates the convolution of a $3^3$ kernel (red) with a $8^3$ grid-octree cell (black).
Only 2 of the 3 dimensions are shown.
A na\"{i}ve implementation evaluates the kernel at every location $(i,j,k)$ within a grid-octree cell as shown in (a).
This results in ${\sim}14$k multiplications for this example.
In contrast, (b) depicts our efficient implementation of the same operation which requires only ${\sim}3$k multiplications.
As all $8^3$ voxels inside the grid-octree cell are the same value, the convolution kernel inside the cell needs to be evaluated only once.
Voxels at the cell boundary need to integrate information from neighboring cells.
This can be efficiently implemented by summing truncated kernels.
See our supp.\ document for details.
}
\label{fig:efficient_octree_conv}
\end{figure}

The convolution operation is the most important, but also the most computational expensive operation in deep convolutional networks.
For a single feature map, convolving a 3D tensor $T$ with a 3D convolution kernel $\Wgts \in \mathbb{R}^{L \times M \times N}$ can be written as
\begin{align}
  \Tnsr^\mathsf{out}_{i,j,k} = \sum_{l=0}^{L-1} \sum_{m=0}^{M-1} \sum_{n=0}^{N-1} \Wgts_{l,m,n} \cdot \Tnsr^\mathsf{in}_{\hat{i}, \hat{j}, \hat{k}} \,,
  \label{eq:tensor_convolution}
\end{align}
with $\hat{i} = i-l+\lfloor L/2 \rfloor$, $\hat{j} = j-m+\lfloor M/2 \rfloor$, $\hat{k} = k-n+\lfloor N/2 \rfloor$.
Similarly, the convolutions on the grid-octree data structure are defined as
\begin{eqnarray}
\Octr^\mathsf{out}[i,j,k] &=& \pool_{(\bar{i},\bar{j},\bar{k})\in \Omega[{i,j,k}]}(\Tnsr_{\bar{i},\bar{j},\bar{k}})  \label{eq:octree_convolution}\\
T_{i,j,k} &=& \sum_{l=0}^{L-1} \sum_{m=0}^{M-1} \sum_{n=0}^{N-1} \Wgts_{l,m,n} \cdot \Octr^\mathsf{in}[\hat{i}, \hat{j}, \hat{k}]\nonumber \,.
\end{eqnarray}

\noindent While this calculation yields the same result as the tensor convolution in \eqnref{eq:tensor_convolution} with the $\octreetotensor,\tensortooctree$ wrapper, we are now able to define a computationally more efficient convolution operator.
Our key observation is that for small convolution kernels and large voxels, $T_{i,j,k}$ is constant within a small margin of the voxel due to its constant support $\Octr^\mathsf{in}[\hat{i}, \hat{j}, \hat{k}]$.
Thus, we only need to compute the convolution within the voxel once, followed by convolution along the surface of the voxel where the support changes due to adjacent voxels taking different values (\figref{fig:efficient_octree_conv}).
This minimizes the number of calculations by a factor of $4$ for voxels of size $8^3$, see supp.\ material for a detailed derivation.
At the same time, it enables a better caching mechanism.

\paragraph{Pooling}
\begin{figure}
\centering
\begin{subfigure}[b]{0.23\textwidth}
\includegraphics[width=0.8\textwidth]{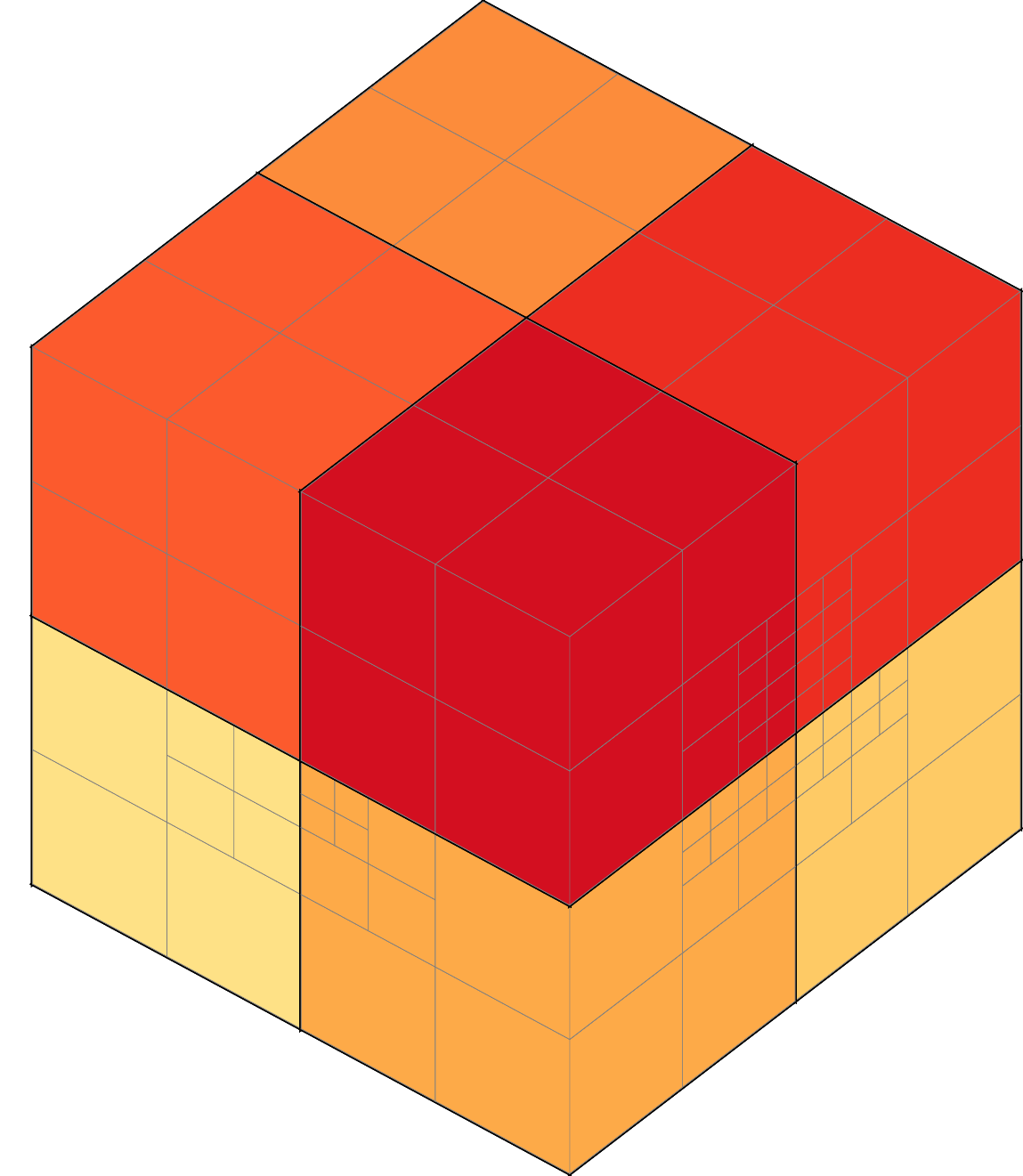}
\caption{Input}
\end{subfigure}
\begin{subfigure}[b]{0.1\textwidth}
\includegraphics[width=0.8\textwidth]{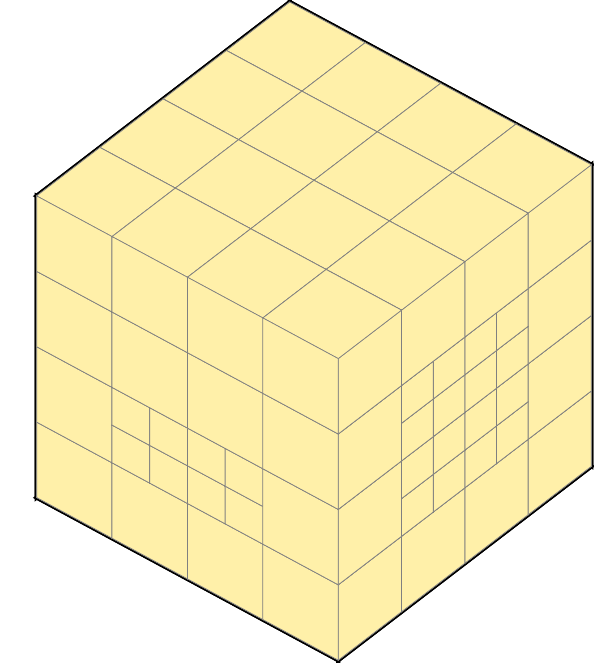}
\caption{Output}
\end{subfigure}
\caption{{\bf Pooling}. The $2^3$ pooling operation on the grid-octree structure combines 8 neighbouring shallow octrees (a) into one shallow octree (b).
The size of each voxel is halved and copied to the new shallow octree structure.
Voxels on the finest resolution are pooled.
Different shallow octrees are depicted in different colors.}
\label{fig:octree_pooling}
\vspace{-0.5cm}
\end{figure}

Another important operation in deep convolutional networks is pooling.
Pooling reduces the spatial resolution of the input tensor and aggregates higher-level information for further processing, thereby increasing the receptive field and capturing context.
For instance, strided $2^3$ max-pooling divides the input tensor $\Tnsr^\mathsf{in}$ into $2^3$ non-overlapping regions and computes the maximum value within each region.
Formally, we have
\begin{align}
  \Tnsr^\mathsf{out}_{i,j,k} = \max_{l,m,n \in [0,1]}\left(\Tnsr^\mathsf{in}_{2i+l,2j+m,2k+n}\right) \,,
\end{align}
where $\Tnsr^\mathsf{in} \in \mathbb{R}^{2D \times 2H \times 2W}$ and $\Tnsr^\mathsf{out} \in \mathbb{R}^{D \times H \times W}$.

To implement pooling on the grid-octree data structure
we reduce the number of shallow octrees.
For an input grid-octree $\Octr^\mathsf{in}$ with $2\ocgriddepth \times 2\ocgridheight \times 2\ocgridwidth$ shallow octrees, the output $\Octr^\mathsf{out}$ contains $\ocgriddepth \times \ocgridheight \times \ocgridwidth$ shallow octrees.
Each voxel of $\Octr^\mathsf{in}$ is halved in size and copied one level deeper in the shallow octree.
Voxels at depth $3$ in $\Octr^\mathsf{in}$ are pooled.
This can be formalized as
\begin{align}
  &\Octr^\mathsf{out}[i,j,k] = 
  \begin{cases}
    \Octr^\mathsf{in}[2i,2j,2k] & \mathrm{if} \ocvoxeldepth(2i,2j,2k) < 3 \\
    P & \mathrm{else}
  \end{cases} \nonumber\\
  &P = \max_{l,m,n \in [0,1]}(\Octr^\mathsf{in}[2i+l,2j+m,2k+n]) \,,
\end{align}
where $\ocvoxeldepth(\cdot)$ computes the depth of the indexed voxel in the shallow octree.
A visual example is depicted in \figref{fig:octree_pooling}.

\paragraph{Unpooling}
\begin{figure}
\centering
\begin{subfigure}[b]{0.1\textwidth}
\includegraphics[width=0.8\textwidth]{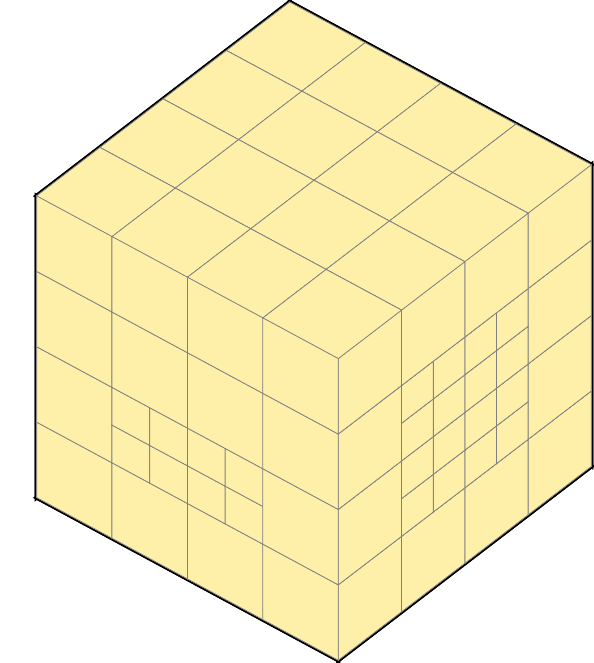}
\caption{Input}
\end{subfigure}
\begin{subfigure}[b]{0.23\textwidth}
\includegraphics[width=0.8\textwidth]{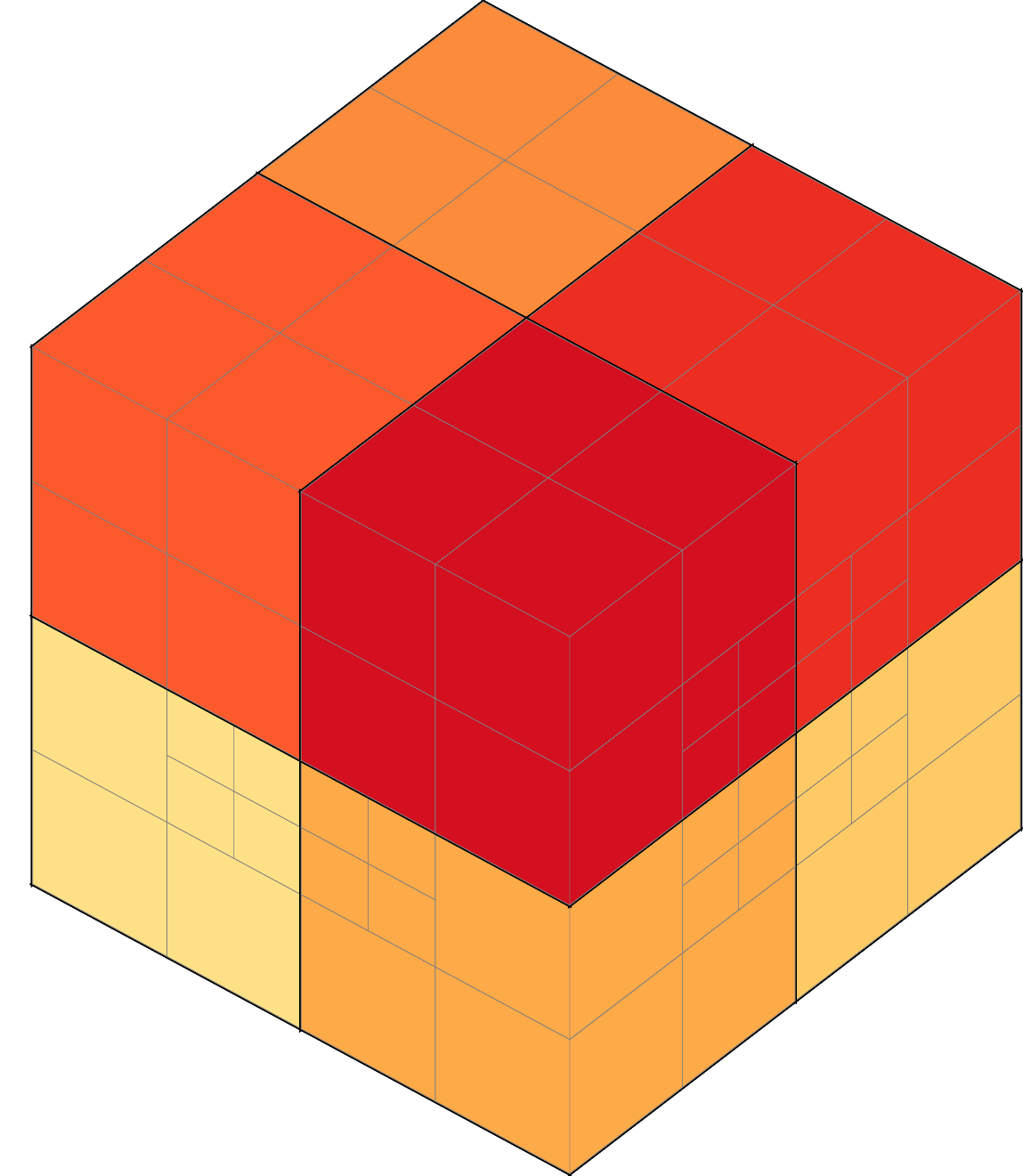}
\caption{Output}
\end{subfigure}
\caption{{\bf Unpooling.}
The $2^3$ unpooling operation transforms a single shallow octree of depth $d$ as shown in (a) into $8$ shallow octrees of depth $d-1$, illustrated in (b). For each node at depth zero one shallow octree is spawned. All other voxels double in size.
Different shallow octrees are depicted in different colors.}
  \label{fig:octree_unpooling}
  \vspace{-0.5cm}
\end{figure}

For several tasks such as semantic segmentation, the desired network output is of the same size as the network input.
While pooling is crucial to increase the receptive field size of the network and capture context, it loses spatial resolution. To increase the resolution of the network, U-shaped network architectures have become popular \cite{Cicek2016ARXIV,Badrinarayanan2015ARXIV} which 
encode information using pooling operations and increase the resolution in a decoder part using unpooling or deconvolution layers \cite{Zeiler2011ICCV}, possibly in combination with skip-connections \cite{He2016CVPR,Dosovitskiy2015ICCV} to increase precision.
The simplest unpooling strategy uses nearest neighbour interpolation and can be formalized on dense input $\Tnsr^\mathsf{in} \in \mathbb{R}^{D \times H \times W}$ and output $\Tnsr^\mathsf{out} \in \mathbb{R}^{2D \times 2H \times 2W}$ tensors as follows:
\begin{align}
  \Tnsr^\mathsf{out}_{i,j,k} = \Tnsr^\mathsf{in}_{\lfloor i/2 \rfloor,\lfloor j/2 \rfloor,\lfloor k/2 \rfloor} \,.
\end{align}
Again, we can define the analogous operation on the hybrid grid-octree data structure by 
\begin{align}
  \Octr^\mathsf{out}[i,j,k] = \Octr^\mathsf{in}[\lfloor i/2 \rfloor,\lfloor j/2 \rfloor,\lfloor k/2 \rfloor] \,.
\end{align}
This operation also changes the data structure: 
The number of shallow octrees increases by a factor of $8$, as each node at depth $0$ spawns a new shallow octree.
All other nodes double their size.
Thus, after this operation the tree depth is decreased.
See \figref{fig:octree_unpooling} for a visual example of this operation.

\noindent {\bf Remark:} To capture fine details, voxels can be split again at the finest resolution according to the original octree of the corresponding pooling layer.
This allows us to take full advantage of skip connections. We follow this approach in our semantic 3D point cloud labeling experiments.

\section{Experimental Evaluation}
\label{sec:results}

In this Section we leverage our OctNet representation to investigate the impact of input resolution on three different 3D tasks: 3D shape classification, 3D orientation estimation and semantic segmentation of 3D point clouds.
To isolate the effect of resolution from other factors we consider simple network architectures.
Orthogonal techniques like data augmentation, joint 2D/3D modeling or ensemble learning are likely to further improve the performance of our models.

\vspace{-1em}
\paragraph{Implementation}
We implemented the grid-octree data structure, all layers including the necessary forward and backward functions, as well as utility methods to create the data structure from point clouds and meshes, as a stand-alone C++/CUDA library. %
This allows the usage of our code within all existing deep learning frameworks.
For our experimental evaluation we used the Torch\footnote{\url{http://torch.ch/}} framework. %

\subsection{3D Classification}

We use the popular ModelNet10 dataset \cite{Wu2015CVPR} for the 3D shape classification task. 
The dataset contains $10$ shape categories and consists of $3991$ 3D shapes for training and $908$ 3D shapes for testing.
Each shape is provided as a triangular mesh, oriented in a canonical pose. 
We convert the triangle meshes to dense respective grid-octree occupancy grids, where a voxel is set to $1$ if it intersects the mesh.
We scale each mesh to fit into a 3D grid of $(N-P)^3$ voxels, where $N$ is the number of voxels in each dimension of the input grid and $P=2$ is a padding parameter.

We first study the influence of the input resolution on memory usage, runtime and classification accuracy.
Towards this goal, we create a series of networks of different input resolution from $8^3$ to $256^3$ voxels.
Each network consists of several blocks which reduce resolution by half until we reach a resolution of $8^3$. 
Each block comprises two convolutional layers ($3^3$ filters, stride $1$) and one max-pooling layer ($2^3$ filters, stride $2$). 
The number of feature maps in the first block is $8$ and increases by $6$ with every block.
After the last block we add a fully-connected layer with $512$ units and a final output layer with $10$ units.
Each convolutional layer and the first fully-connected layer are followed by a rectified linear unit \cite{Krizhevsky2012NIPS} as activation function and the weights are initialized as described in \cite{He2015ICCV}. 
We use the standard cross-entropy loss for training and train all networks for $20$ epochs with a batch size of $32$ using Adam \cite{Kingma2015ICLR}.
The initial learning rate is set to $0.001$ and we decrease the learning rate by a factor of $10$ after $15$ epochs.

\begin{figure}[t!]
  \center
  \begin{subfigure}[b]{0.23\textwidth}
    \includegraphics[width=\textwidth]{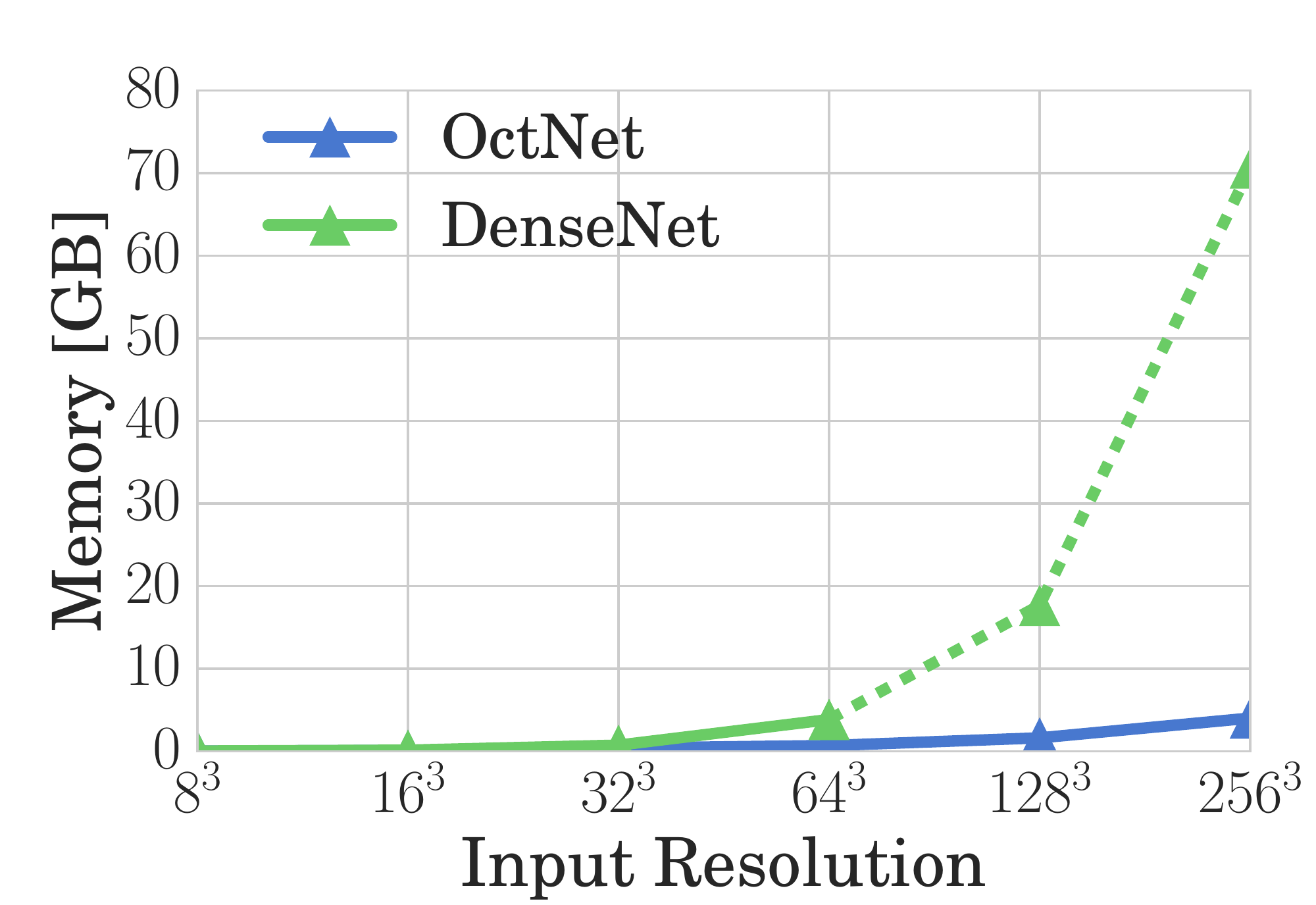}
    \caption{Memory}
    \label{fig:results_modelnet_memory}
  \end{subfigure}
  \begin{subfigure}[b]{0.23\textwidth}
    \includegraphics[width=\textwidth]{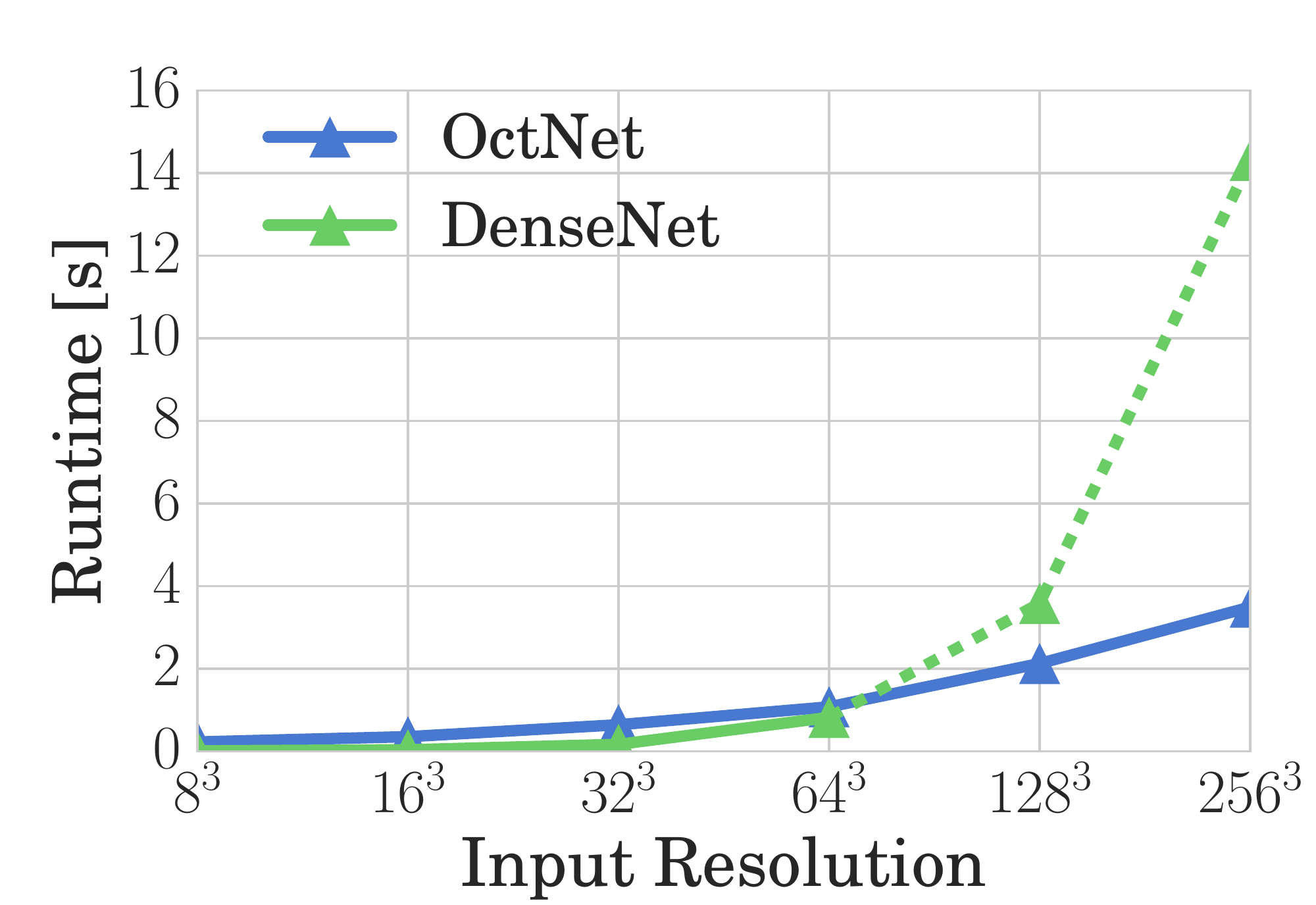}
    \caption{Runtime}
    \label{fig:results_modelnet_runtime}
  \end{subfigure}
  \begin{subfigure}[b]{0.23\textwidth}
    \includegraphics[width=\textwidth]{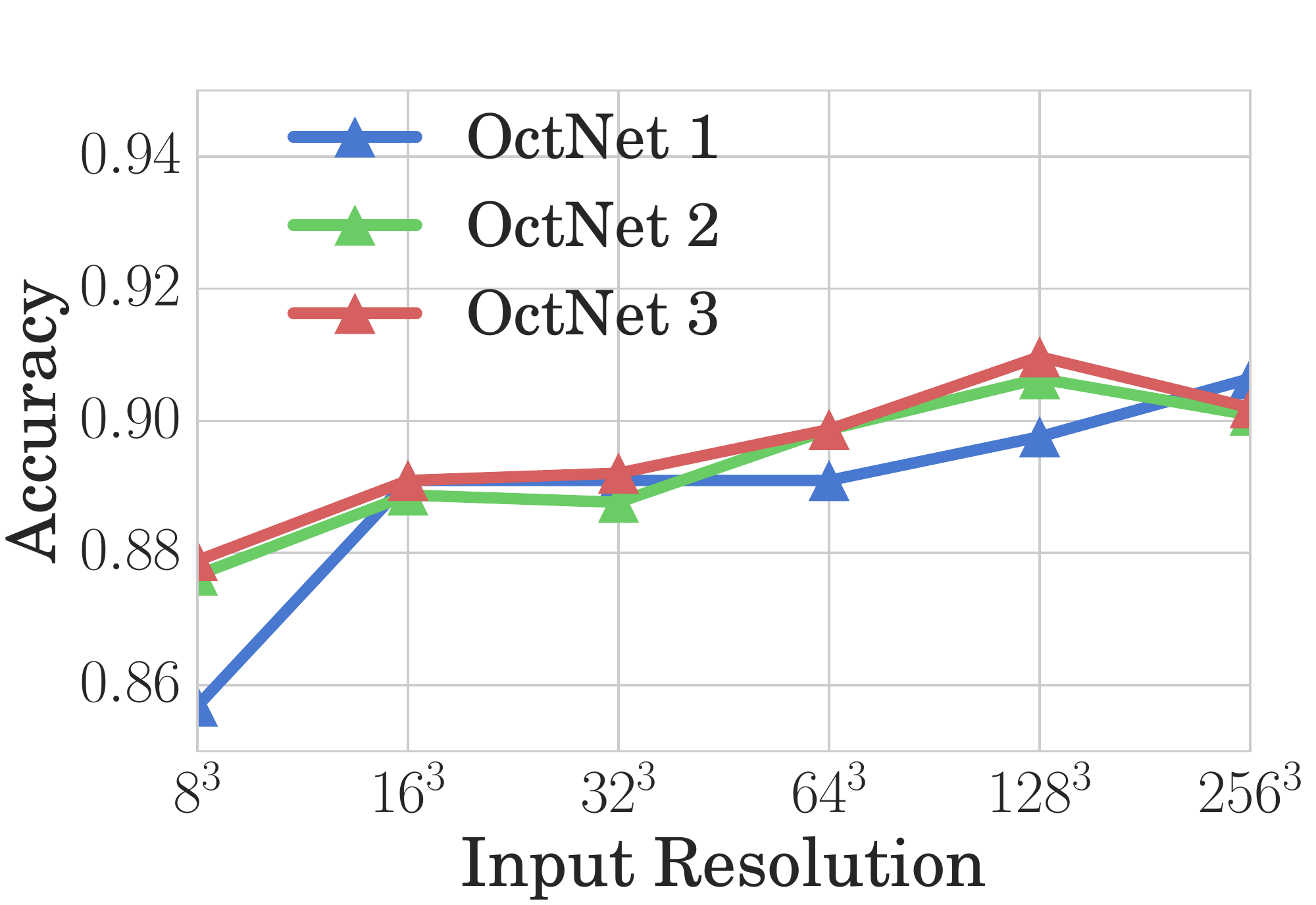}
    \caption{Accuracy}
    \label{fig:results_modelnet_unfair}
  \end{subfigure}
  \begin{subfigure}[b]{0.23\textwidth}
    \includegraphics[width=\textwidth]{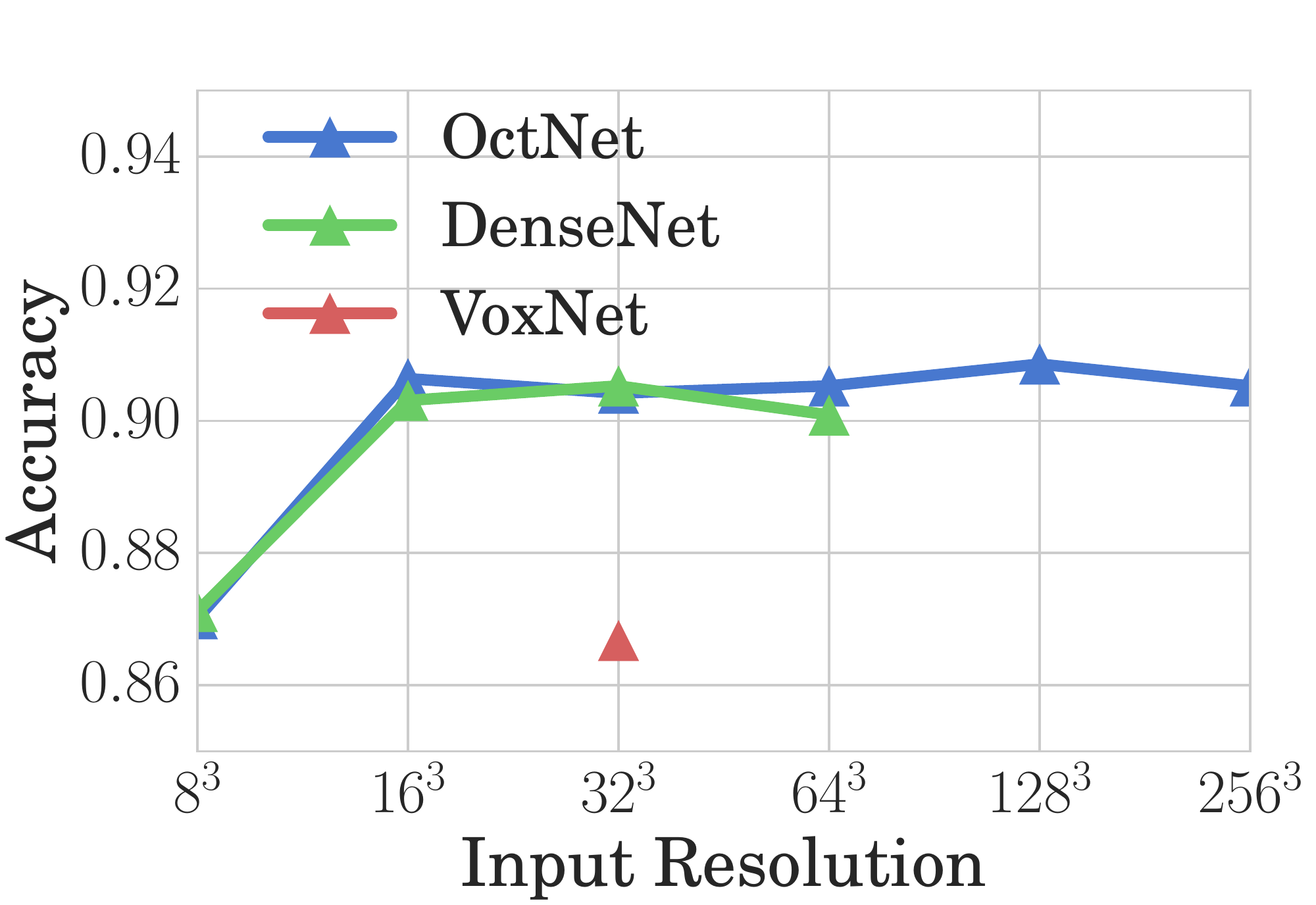}
    \caption{Accuracy}
    \label{fig:results_modelnet_fair}
  \end{subfigure}
  \vspace{-0.5em}
  \caption{{\bf Results on ModelNet10 Classification Task.}}
  \label{fig:results_modelnet}
  \vspace{-1.65em}
\end{figure}

Overall, we consider three different types of networks: the original VoxNet architecture of Maturana \etal \cite{Maturana2015IROS} which operates on a fixed $32^3$ voxel grid, the proposed OctNet and a dense version of it which we denote ``DenseNet'' in the following.
While performance gains can be obtained using orthogonal approaches such as network ensembles \cite{Brock2016ARXIV} or a combination of 3D and 2D convolutional networks \cite{Su2015ICCV,Hegde2016ARXIV}, in this paper we deliberately focus on ``pure'' 3D convolutional network approaches to isolate the effect of resolution from other influencing factors.

\figref{fig:results_modelnet} shows our results. 
First, we compare the memory consumption and run-time of our OctNet \wrt the dense baseline approach, see \figref{fig:results_modelnet_memory} and \ref{fig:results_modelnet_runtime}.
Importantly, OctNets require significantly less memory and run-time for high input resolutions compared to dense input grids. 
Using a batch size of $32$ samples, our OctNet easily fits in a modern GPU's memory (12GB) for an input resolution of $256^3$.
In contrast, the corresponding dense model fits into the memory only for resolutions ${\le}64^3$.
A more detailed analysis of the memory consumption wrt.\ the sparsity in the data is provided in the supp.\ document.
OctNets also run faster than their dense counterparts for resolutions ${>}64^3$.
For resolutions ${\le}64^3$, OctNets run slightly slower due to the overhead incurred by the grid-octree representation and processing.

\begin{figure}
\captionsetup[subfigure]{labelformat=empty}
  \center
  \rotatebox{90}{\hspace{1.2cm} $64^3$ \hspace{1.1cm} $32^3$ \hspace{1.1cm} $16^3$ \hspace{1.1cm} $8^3$}~
  \begin{subfigure}[b]{0.1\textwidth}
    \center
    \includegraphics[height=0.07\textheight]{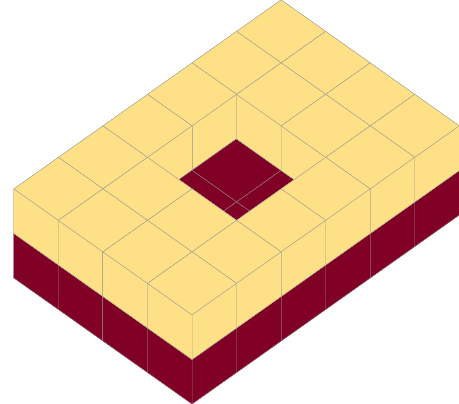}\\%
    \includegraphics[height=0.075\textheight]{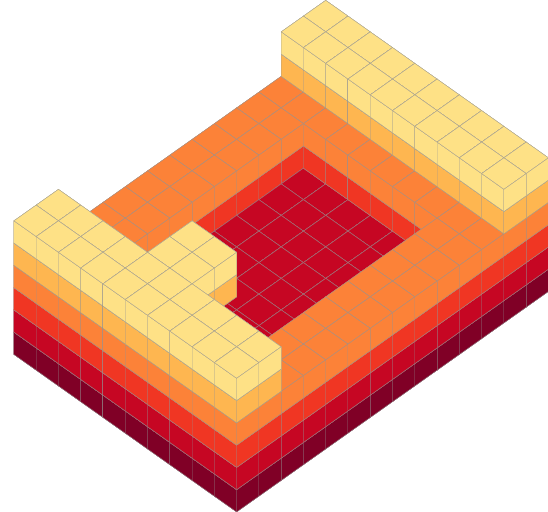}\\%
    \includegraphics[height=0.075\textheight]{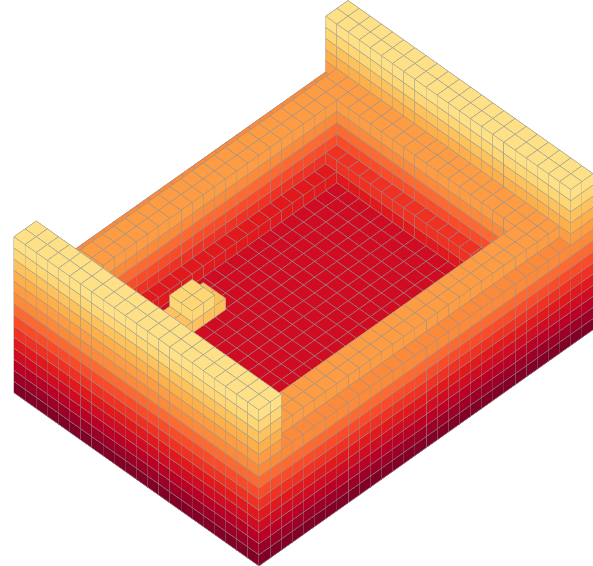}\\%
    \includegraphics[height=0.075\textheight]{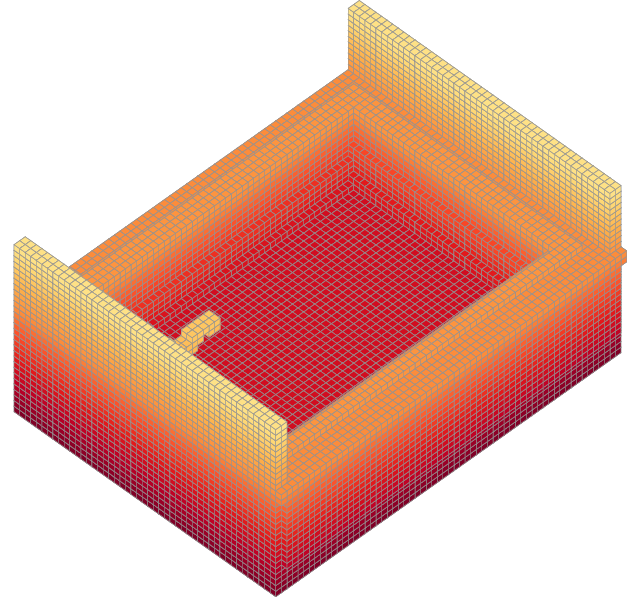}%
    \caption{Bathtub}
  \end{subfigure}
  \begin{subfigure}[b]{0.1\textwidth}
    \center
    \includegraphics[height=0.075\textheight]{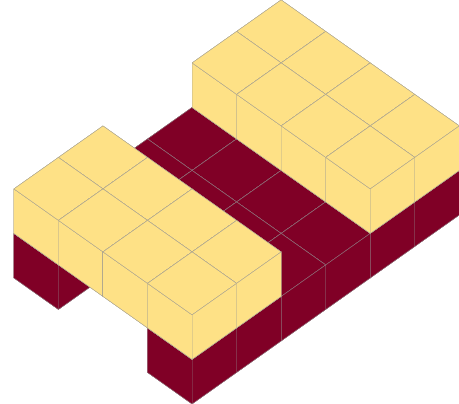}\\%
    \includegraphics[height=0.075\textheight]{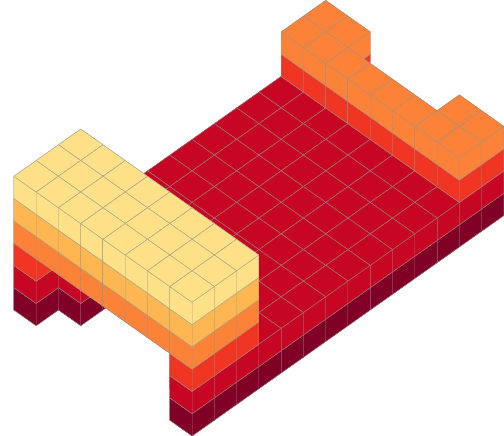}\\%
    \includegraphics[height=0.075\textheight]{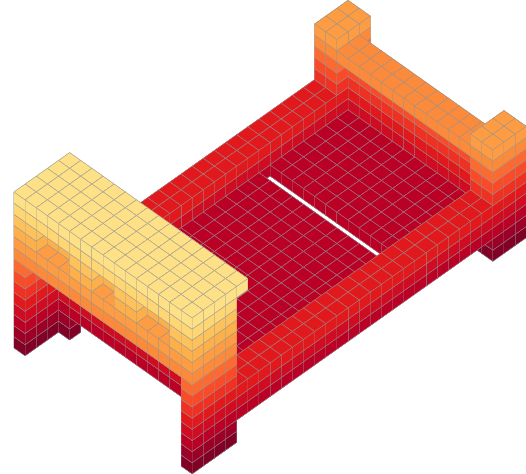}\\%
    \includegraphics[height=0.075\textheight]{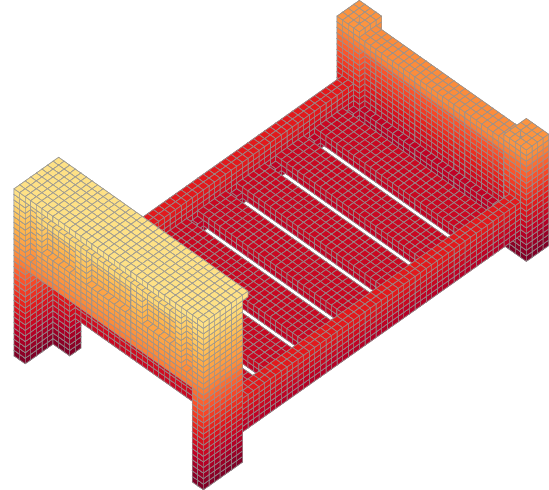}%
    \caption{Bed}
  \end{subfigure}
  \begin{subfigure}[b]{0.1\textwidth}
    \center
    \includegraphics[height=0.075\textheight]{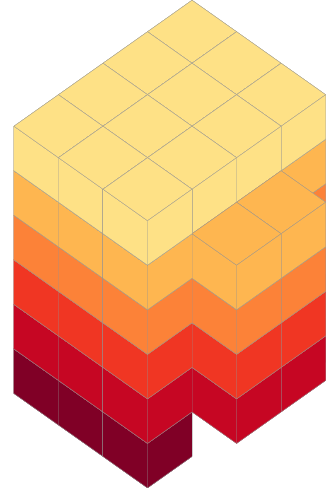}\\%
    \includegraphics[height=0.075\textheight]{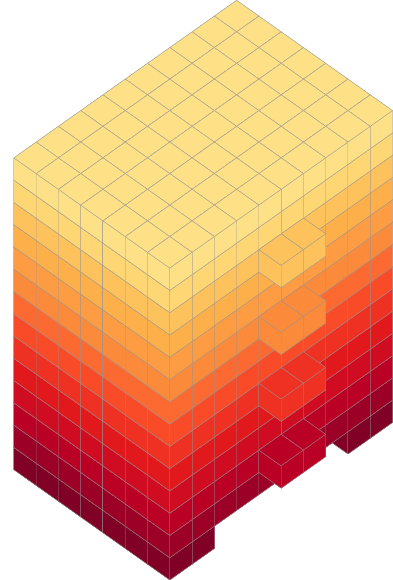}\\%
    \includegraphics[height=0.075\textheight]{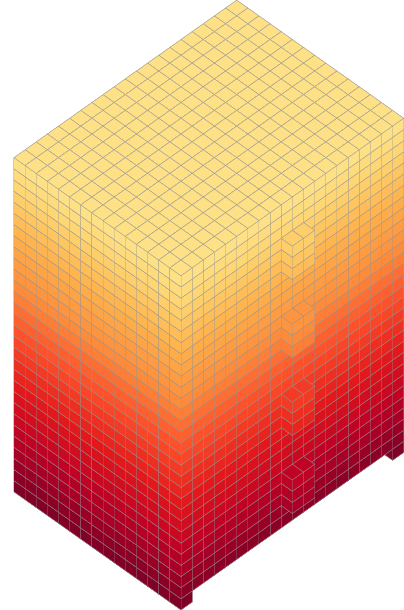}\\%
    \includegraphics[height=0.075\textheight]{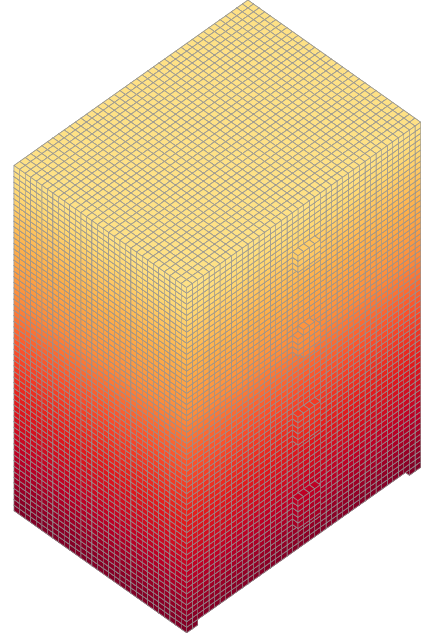}%
    \caption{Dresser}
  \end{subfigure}
  \begin{subfigure}[b]{0.1\textwidth}
    \center
    \includegraphics[height=0.075\textheight]{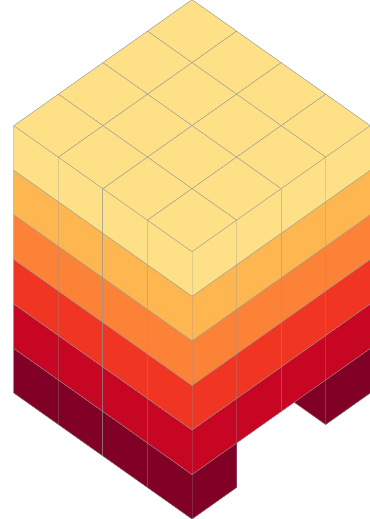}\\%
    \includegraphics[height=0.075\textheight]{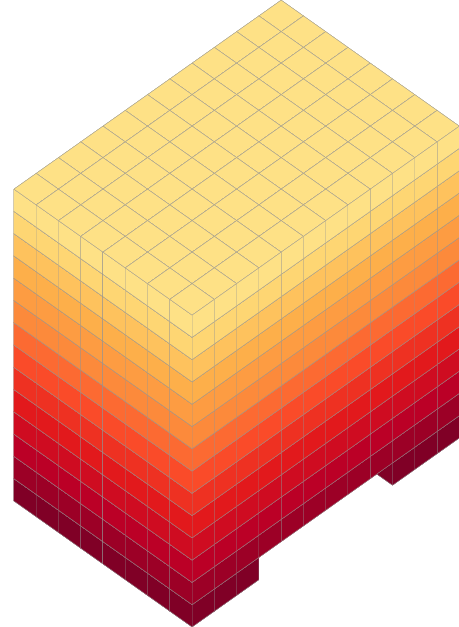}\\%
    \includegraphics[height=0.075\textheight]{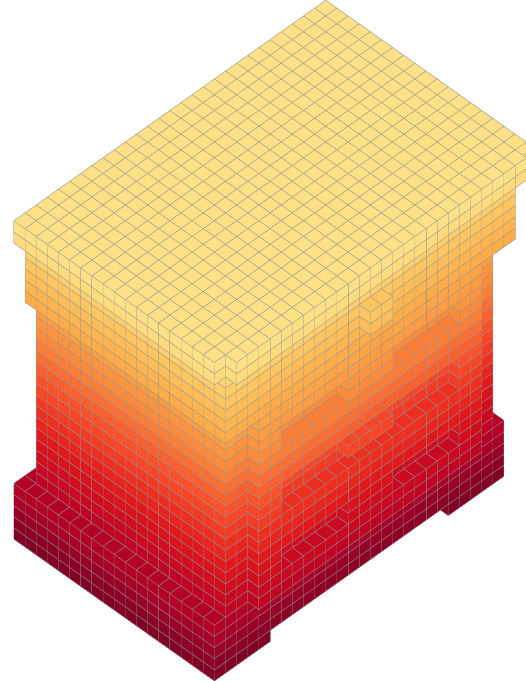}\\%
    \includegraphics[height=0.075\textheight]{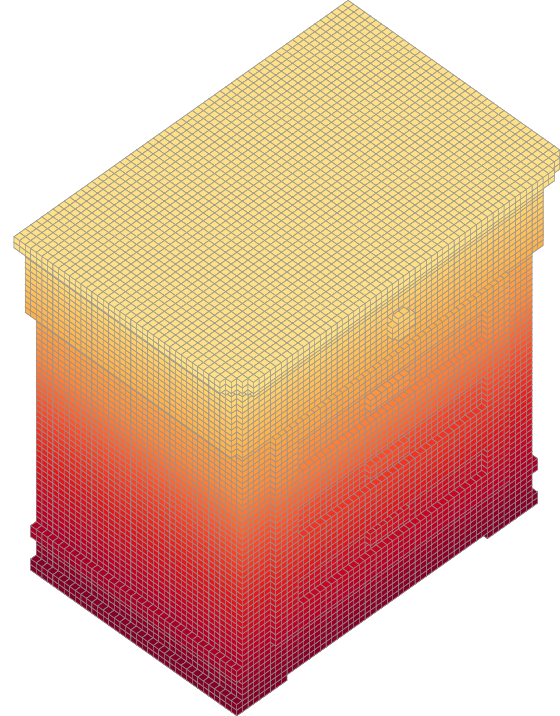}%
    \caption{N. Stand}
  \end{subfigure}
  \caption{{\bf Voxelized 3D Shapes from ModelNet10.}}
  \label{fig:modelnet10_inputs}
  \vspace{-0.5cm}
\end{figure}

Leveraging our OctNets, we now compare the impact of input resolution with respect to classification accuracy.
\figref{fig:results_modelnet_unfair} shows the results of different OctNet architectures where we keep the number of convolutional layers per block fixed to 1, 2 and 3.
\figref{fig:results_modelnet_fair} shows a comparison of accuracy with respect to DenseNet and VoxNet when keeping the capacity of the model, \ie, the number of parameters, constant by removing max-pooling layers from the beginning of the network.
We first note that despite its pooled representation, OctNet performs on par with its dense equivalent.
This confirms our initial intuition (\figref{fig:sparse_activations}) that sparse data allows for allocating resources adaptively without loss in performance.
Furthermore, both models outperform the shallower VoxNet architecture, indicating the importance of network depth.

\begin{figure}[t!]
\captionsetup[subfigure]{labelformat=empty}
  \begin{subfigure}[b]{0.48\linewidth}
    \includegraphics[width=\textwidth]{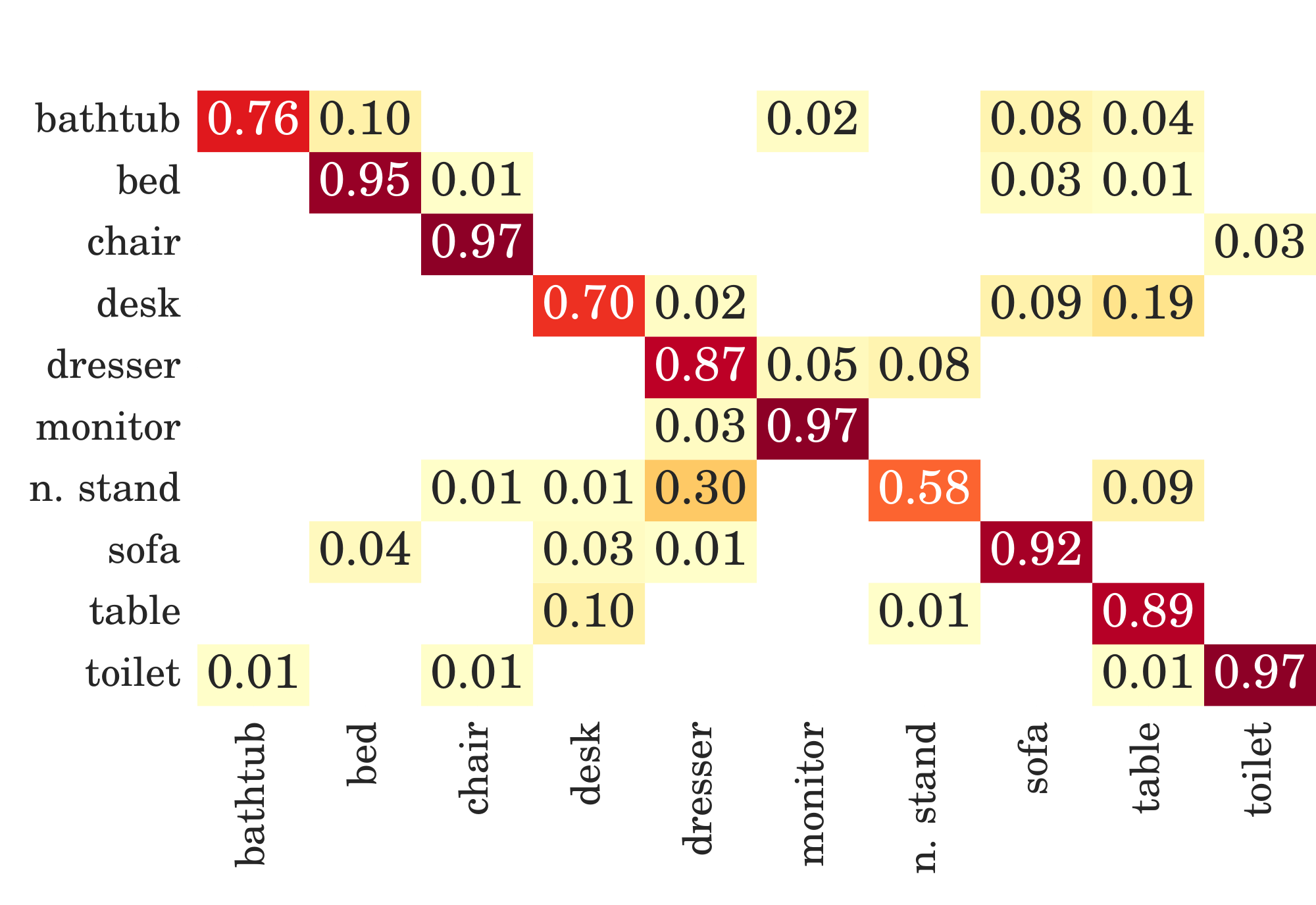}
    \caption{\qquad$8^3$}
  \end{subfigure}
  \begin{subfigure}[b]{0.48\linewidth}
    \includegraphics[width=\textwidth]{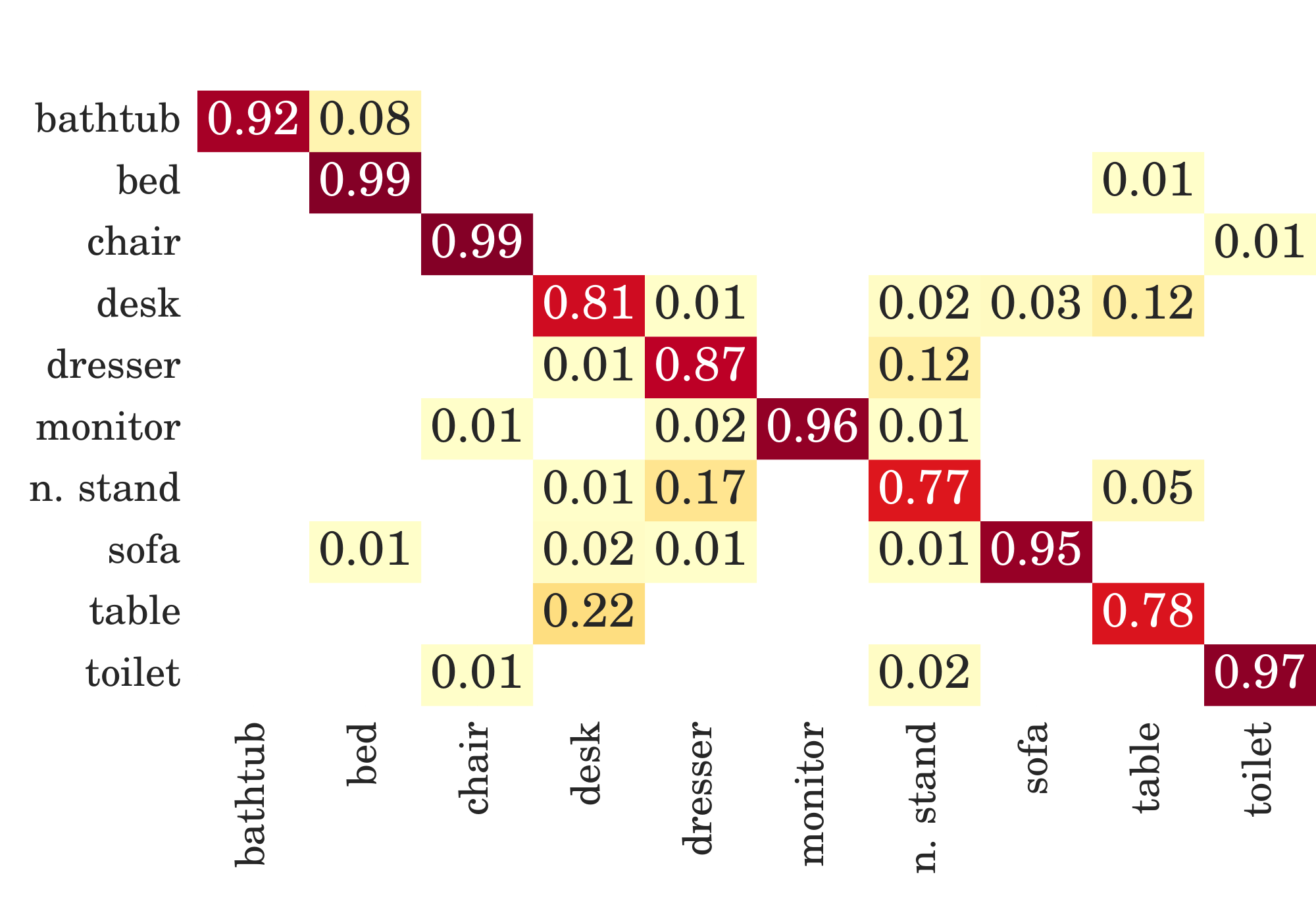}
    \caption{\qquad$32^3$}
  \end{subfigure}
  \caption{
    {\bf Confusion Matrices on ModelNet10.}
  }
  \label{fig:results_modelnet_confusion}
\vspace{-0.5cm}
\end{figure}

Regarding classification accuracy we observed improvements for lower resolutions but diminishing returns beyond an input resolution of $32^3$ voxels.
Taking a closer look at the confusion matrices in \figref{fig:results_modelnet_confusion}, we observe that higher input resolution helps for some classes, \eg, \textit{bathtub}, while others remain ambiguous independently of the resolution, \eg, \textit{dresser} vs.\ \textit{night stand}.
We visualize this lack of discriminative power by showing voxelized representations of 3D shapes from the ModelNet10 database \figref{fig:modelnet10_inputs}.
While bathtubs look similar to beds (or sofas, tables) at low resolution they can be successfully distinguished at higher resolutions.
However, a certain ambiguity between dresser and night stand remains.

\subsection{3D Orientation Estimation}

In this Section, we investigate the importance of input resolution on 3D orientation estimation. 
Most existing approaches to 3D pose estimation \cite{Brachmann2014ECCV,Brachmann2016CVPR,Wohlhart2015CVPR,Rios-Cabrera2013ICCV,Tejani2014ECCV} assume that the true 3D shape of the {\it object instance} is known. 
To assess the generalization ability of 3D convolutional networks, we consider a slightly different setup where only the {\it object category} is known.
After training a model on a hold-out set of 3D shapes from a single category, we test the ability of the model to predict the 3D orientation of unseen 3D shapes from the same category.

More concretely, given an instance of an object category with unknown pose, the goal is to estimate the rotation with respect to the canonical pose.
We utilize the 3D shapes from the \textit{chair} class of the ModelNet10 dataset and rotate them randomly between $\pm 15^\circ$ around each axis.
We use the same network architectures and training protocol as in the classification experiment, except that the networks regress orientations.
We use unit quaternions to represent 3D rotations and train our networks with an Euclidean loss. For small angles, this loss is a good approximation to the rotation angle $\phi = \arccos(2 \langle q_1, q_2 \rangle^2 - 1)$ between quaternions $q_1, q_2$.

\begin{figure}
  \center
  \begin{subfigure}[b]{0.48\linewidth}
    \includegraphics[width=\linewidth]{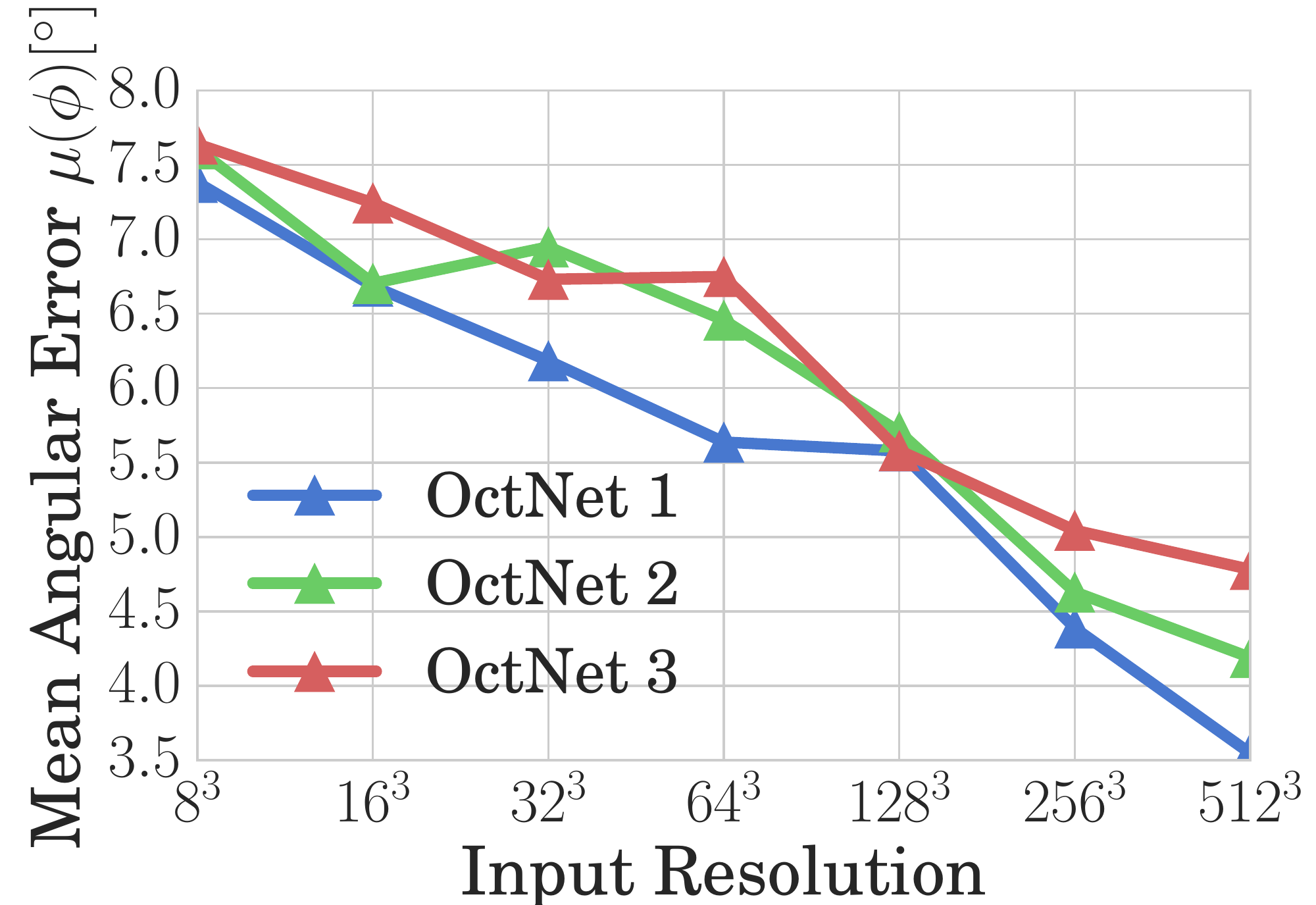}
    \caption{Mean Angular Error}
    \label{fig:orientation_modelnet_chairs_mse_unfair}
  \end{subfigure}
  \begin{subfigure}[b]{0.48\linewidth}
    \includegraphics[width=\linewidth]{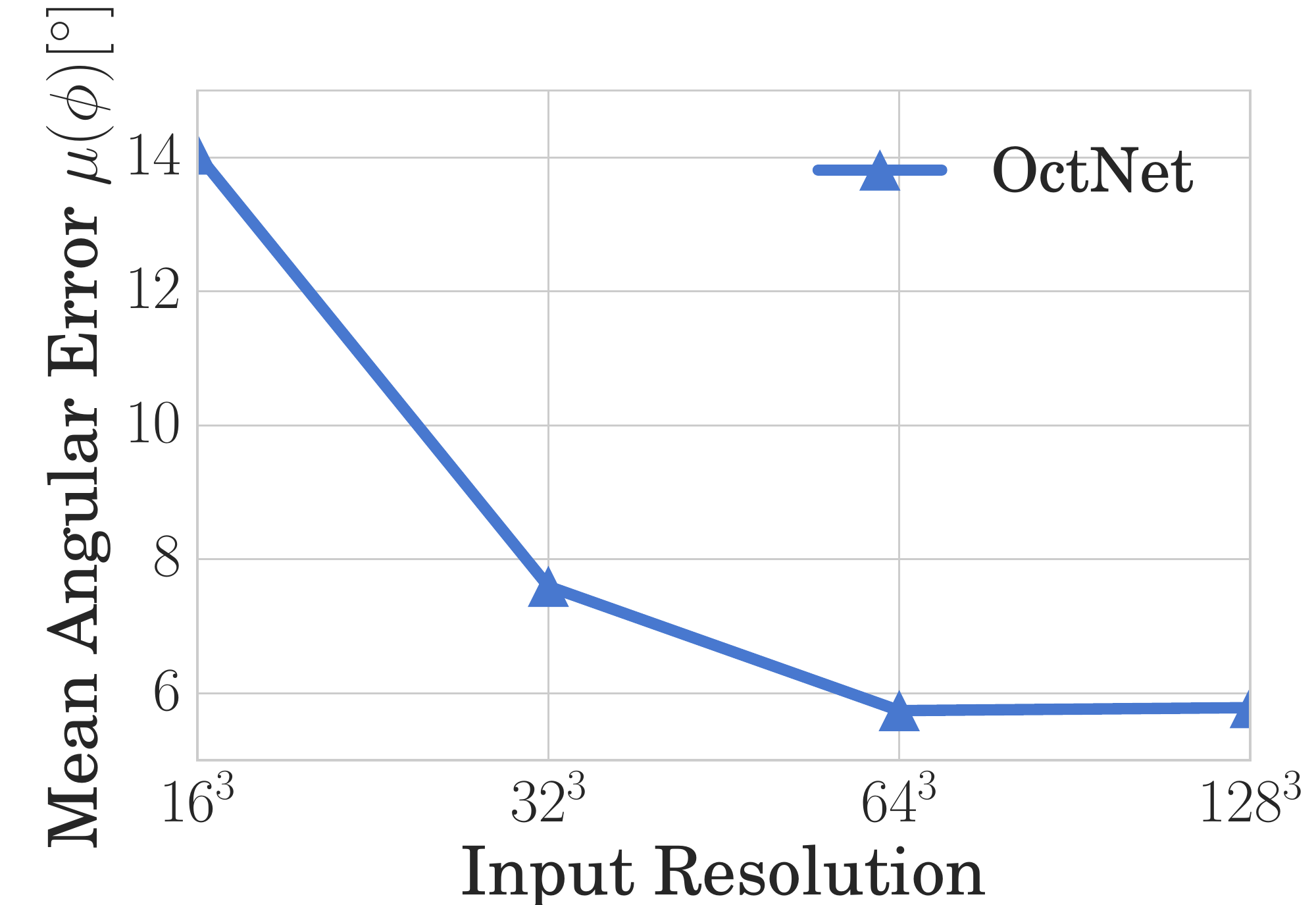}
    \caption{Mean Angular Error}
    \label{fig:orientation_modelnet_chairs_mse_fair}
  \end{subfigure}
  \caption{{\bf Orientation Estimation on ModelNet10.}}
  \label{fig:orientation_modelnet_chairs}
  \vspace{-0.5cm}
\end{figure}

\begin{figure}
\captionsetup[subfigure]{labelformat=empty}
  \center
  \begin{subfigure}[b]{0.2\linewidth}
    \center
    \includegraphics[width=\linewidth]{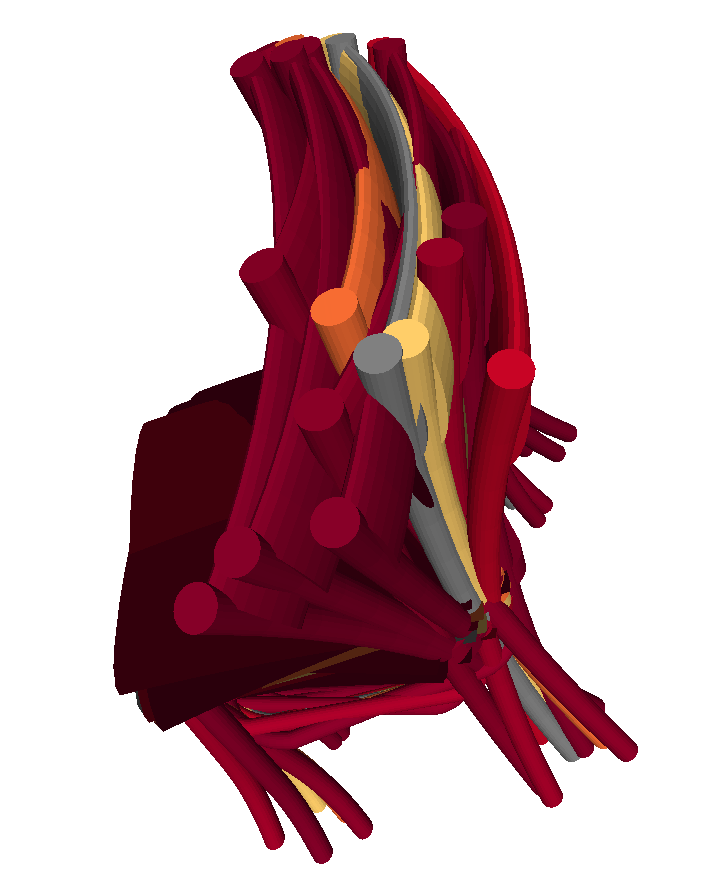}
    \includegraphics[width=\linewidth]{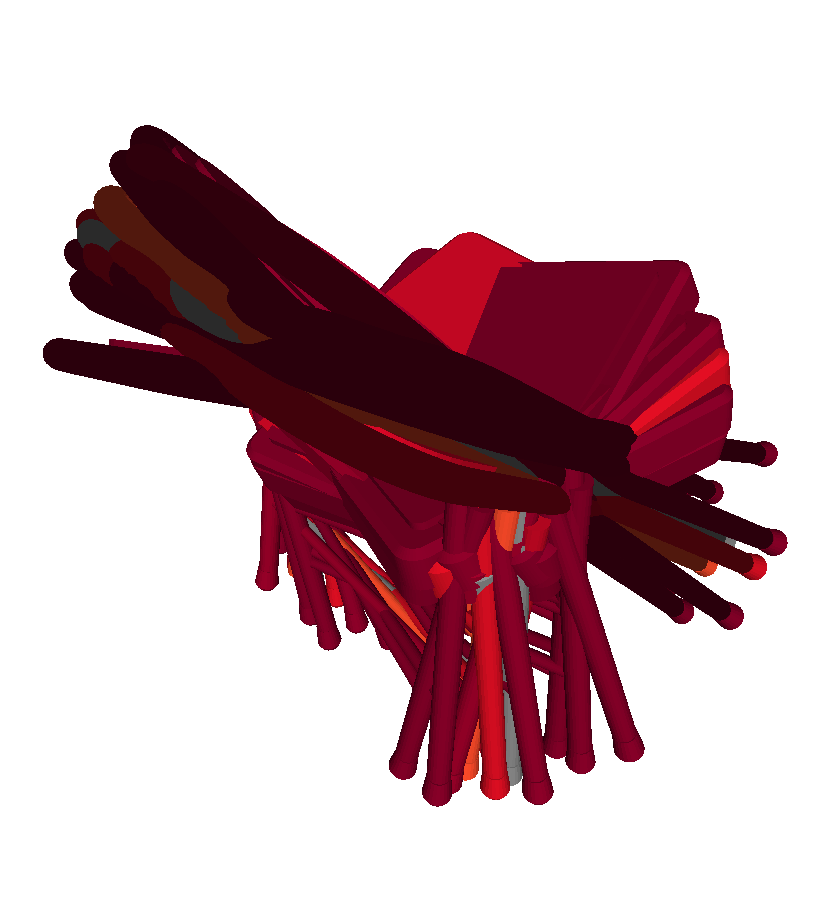}
    \includegraphics[width=\linewidth]{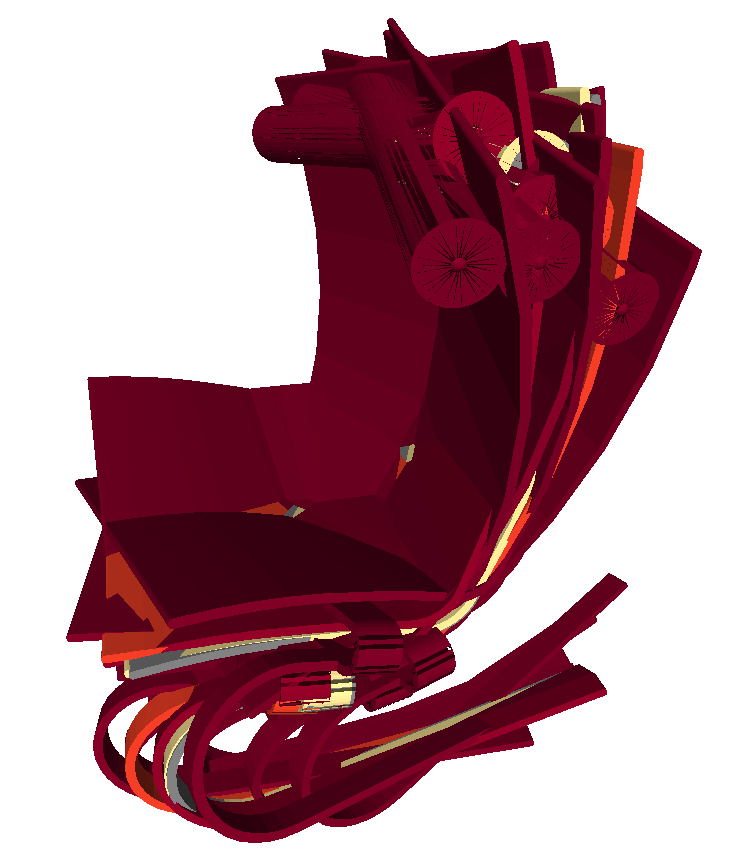}
    \caption{$16^3$}
  \end{subfigure}~
  \begin{subfigure}[b]{0.2\linewidth}
    \center
    \includegraphics[width=\linewidth]{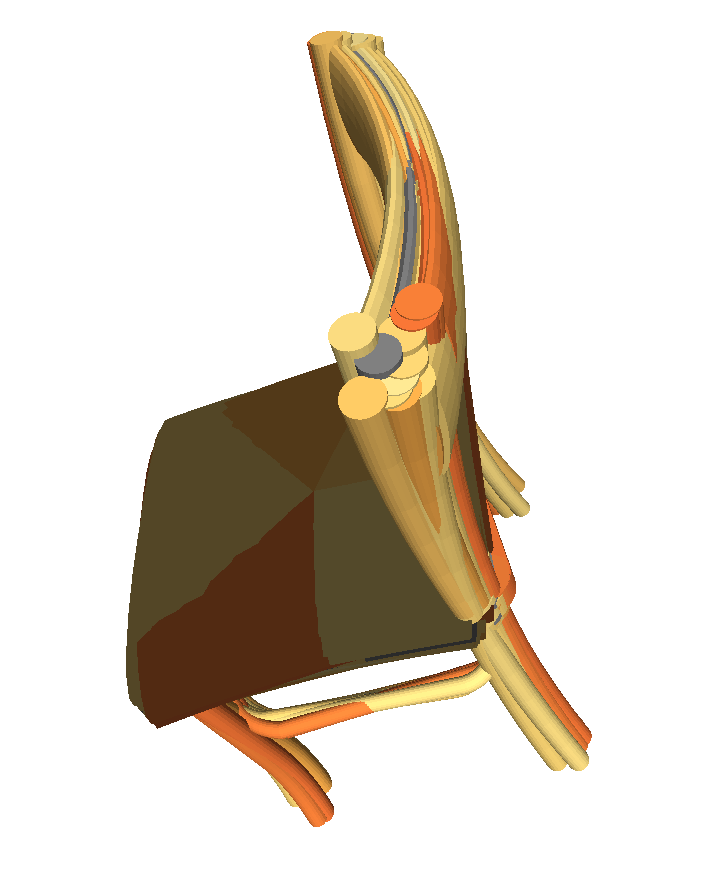}
    \includegraphics[width=\linewidth]{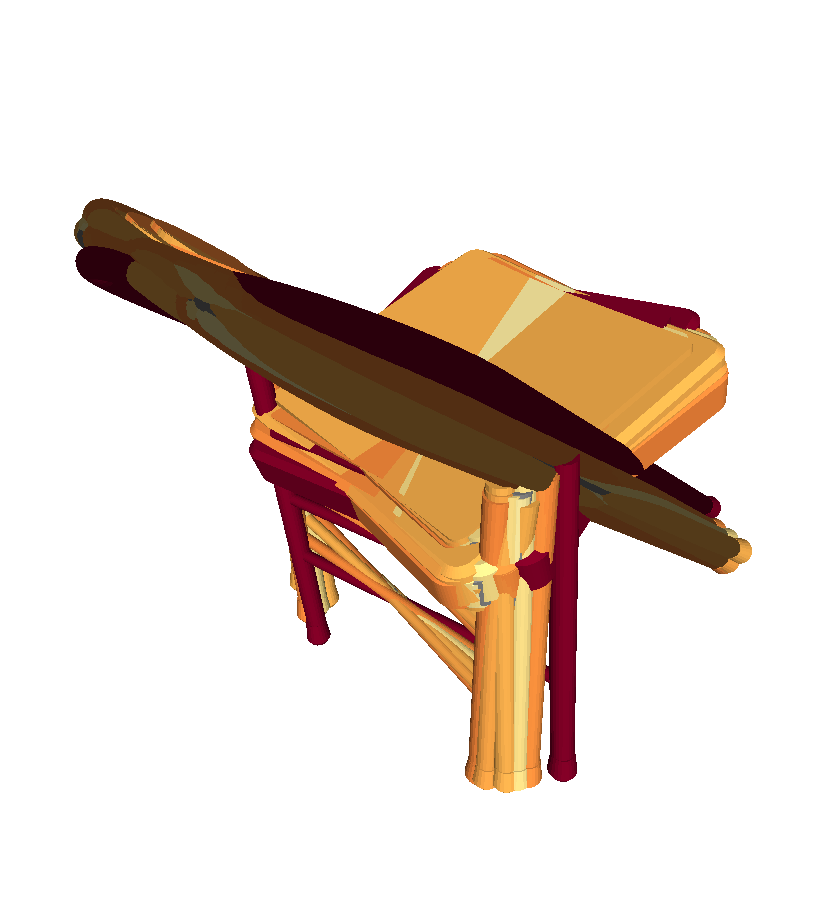}
    \includegraphics[width=\linewidth]{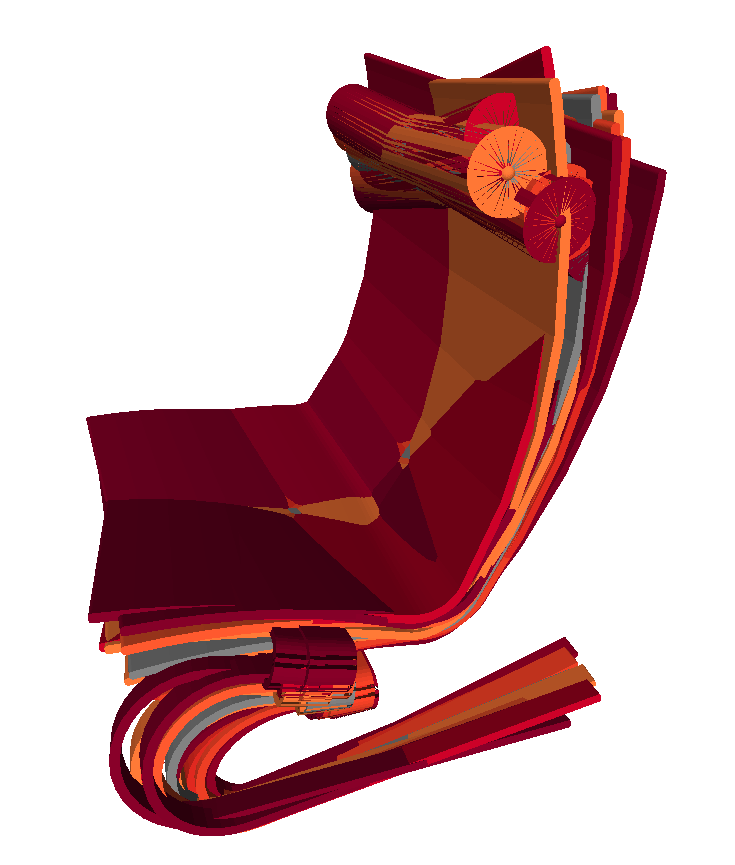}
    \caption{$32^3$}
  \end{subfigure}~
  \begin{subfigure}[b]{0.2\linewidth}
    \center
    \includegraphics[width=\linewidth]{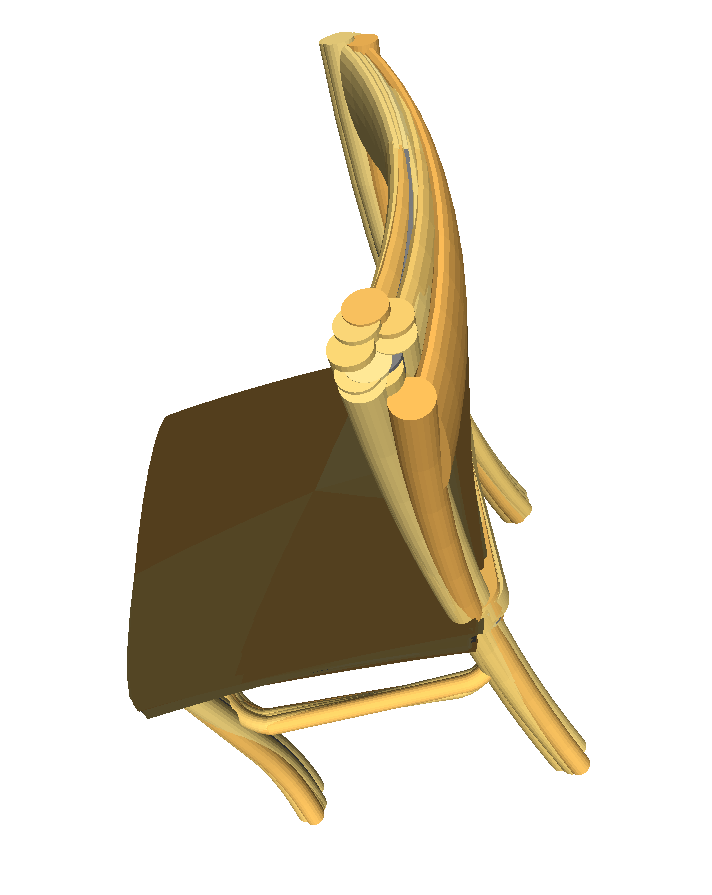}
    \includegraphics[width=\linewidth]{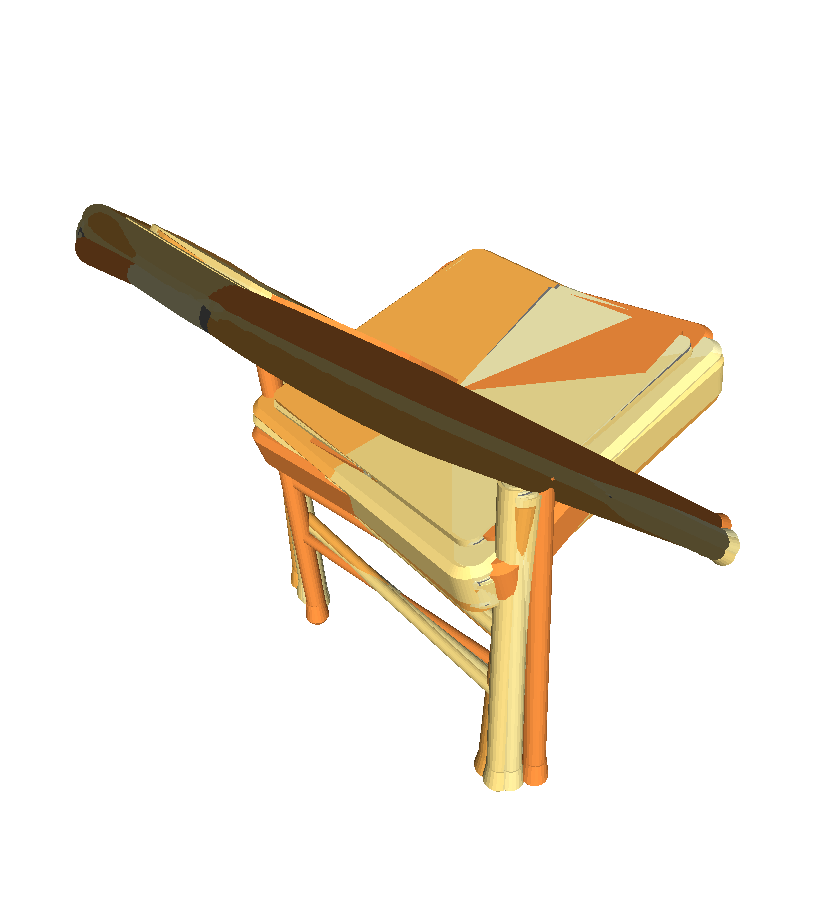}
    \includegraphics[width=\linewidth]{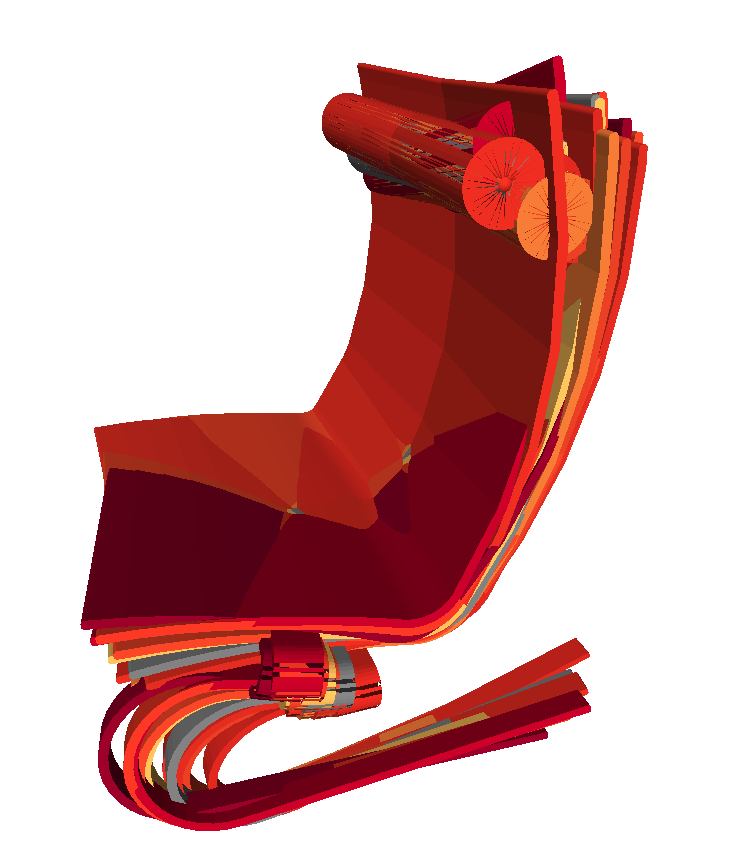}
    \caption{$64^3$}
  \end{subfigure}~
  \begin{subfigure}[b]{0.2\linewidth}
    \center
    \includegraphics[width=\linewidth]{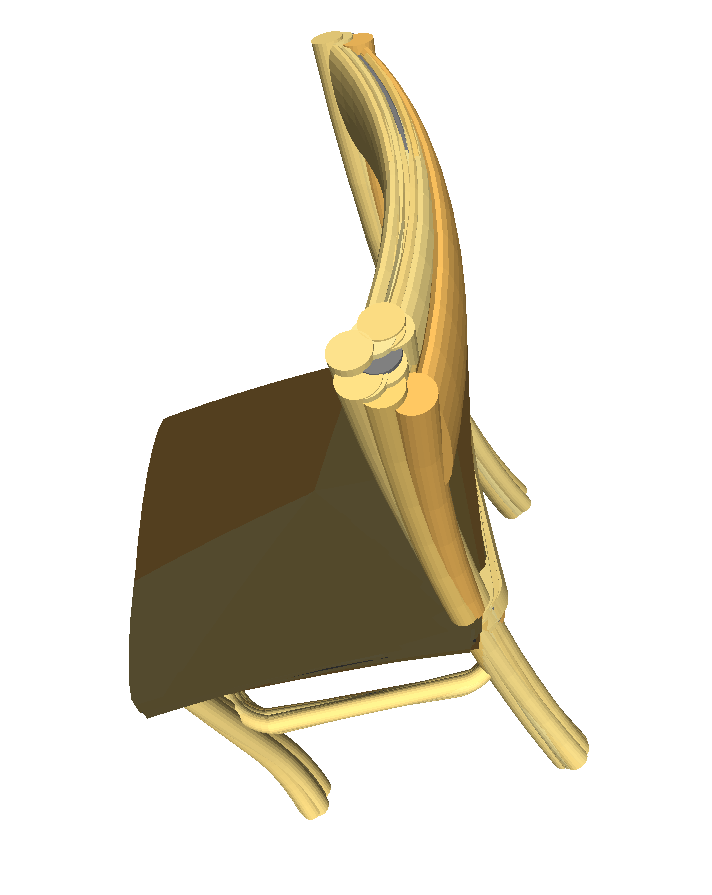}
    \includegraphics[width=\linewidth]{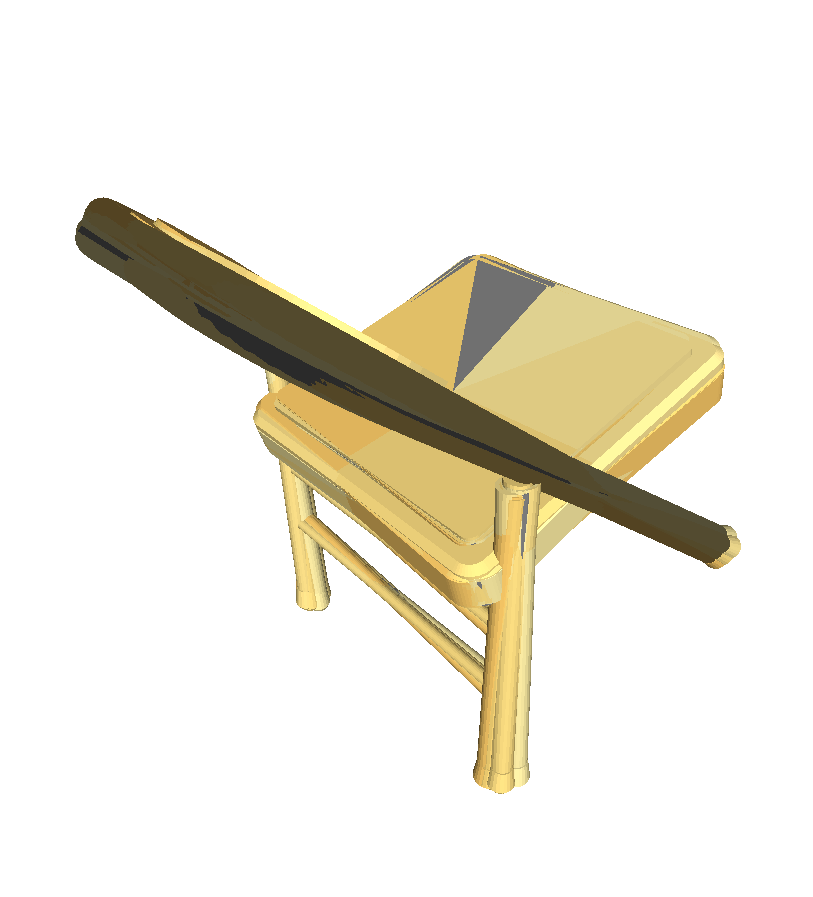}
    \includegraphics[width=\linewidth]{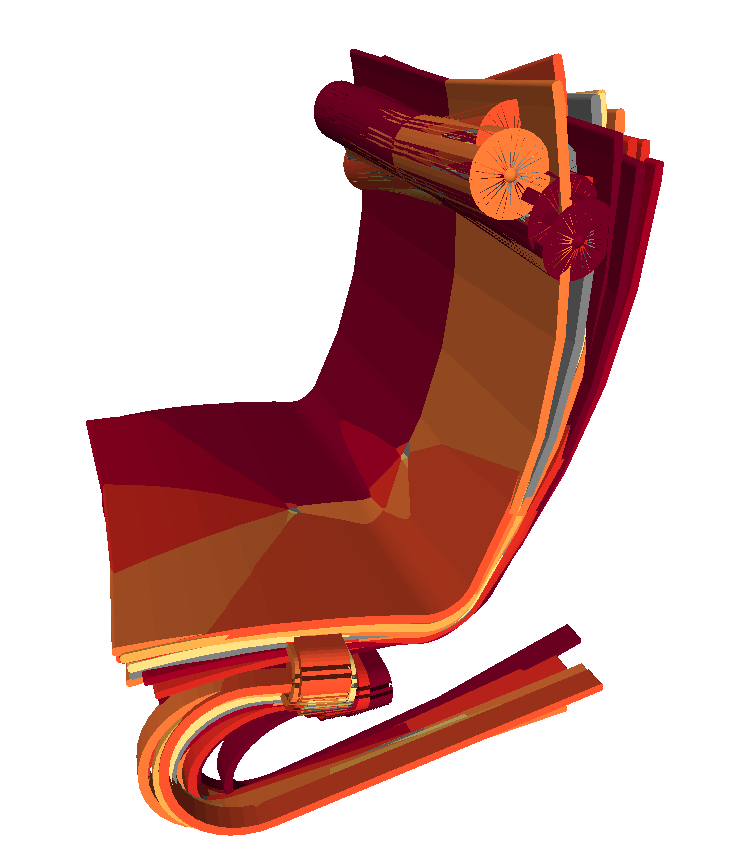}
    \caption{$128^3$}
  \end{subfigure}
  \vspace{-0.5em}
  \caption{
    {\bf Orientation Estimation on ModelNet10.}
    This figure illustrates 10 rotation estimates for 3 chair instances while varying the input resolution from $16^3$ to $128^3$.
    Darker colors indicate larger deviations from the ground truth.
  }
  \label{fig:qualitative_orientation}
  \vspace{-1.5em}
\end{figure}

\figref{fig:orientation_modelnet_chairs} shows our results using the same naming convention as in the previous Section.
We observe that fine details are more important compared to the classification task.
For the OctNet 1-3 architectures we observe a steady increase in performance, while for networks with constant capacity across resolutions (\figref{fig:orientation_modelnet_chairs_mse_fair}), performance levels beyond $128^3$ voxels input resolution.
Qualitative results of the latter experiment are shown in \figref{fig:qualitative_orientation}.
Each row shows 10 different predictions for two randomly selected chair instance over several input resolutions, ranging from $16^3$ to $128^3$.
Darker colors indicate larger errors which occur more frequently at lower resolutions.
In contrast, predictions at higher network resolutions cluster around the true pose.
Note that learning a dense 3D representation at a resolution of $128^3$ voxels or beyond would not be feasible.

\subsection{3D Semantic Segmentation}

In this Section, we evaluate the proposed OctNets on the problem of labeling 3D point cloud with semantic information.
We use the RueMonge2014 dataset~\cite{Riemenschneider2014ECCV} that provides a colored 3D point cloud of several Haussmanian style facades, comprising ${\sim}1$ million 3D points in total.
The labels are \textit{window}, \textit{wall}, \textit{balcony}, \textit{door}, \textit{roof}, \textit{sky} and \textit{shop}. 

For this task, we train a U-shaped network \cite{Badrinarayanan2015ARXIV,Cicek2016ARXIV} on three different input resolutions, $64^3$, $128^3$ and $256^3$, where the voxel size was selected such that the height of all buildings fits into the input volume.
We first map the point cloud into the grid-octree structure.
For all leaf nodes which contain more than one point, we average the input features and calculate the majority vote of the ground truth labels for training.
As features we use the binary voxel occupancy, the RGB color, the normal vector and the height above ground.
Due to the small number of training samples, we augment the data for this task by applying small rotations.

Our network architecture comprises an encoder and a decoder part.
The encoder part consists of four blocks which comprise $2$ convolution layers ($3^3$ filters, stride $1$) followed by one max-pooling layer each.
The decoder consists of four blocks which comprise $2$ convolutions ($3^3$ filters, stride $1$) followed by a guided unpooling layer as discussed in the previous Section.
Additionally, after each unpooling step all features from the last layer of the encoder at the same resolution are concatenated to provide high-resolution details.
All networks are trained with a per voxel cross entropy loss using Adam \cite{Kingma2015ICLR} and a learning rate of $0.0001$.

\tabref{tab:semantic_segmentation} compares the proposed OctNet to several state of the art approaches on the facade labeling task following the extended evaluation protocol of \cite{Gadde2016ARXIV}.
The 3D points of the test set are assigned the label of the corresponding grid-octree voxels. 
As evaluation measures we use \textit{overall pixel accuracy} $\tfrac{\text{TP}}{\text{TP} + \text{FN}}$ over all 3D points, \textit{average class accuracy}, and \textit{intersection over union} $\tfrac{\text{TP}}{\text{TP} + \text{FN} + \text{FP}}$ over all classes. Here, FP, FN and TP denote false positives, false negatives and true positives, respectively.

\begin{figure}
   \center
  \begin{subfigure}[b]{0.45\linewidth}
    \includegraphics[width=\linewidth]{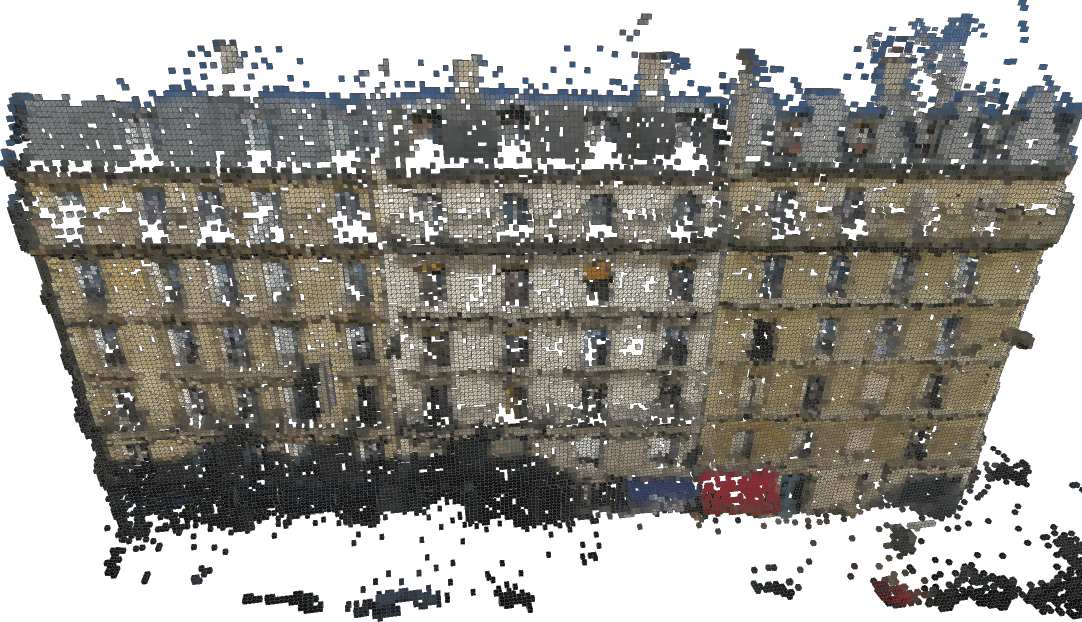}
    \caption{Voxelized Input}
   \end{subfigure}
  \begin{subfigure}[b]{0.45\linewidth}
    \includegraphics[width=\linewidth]{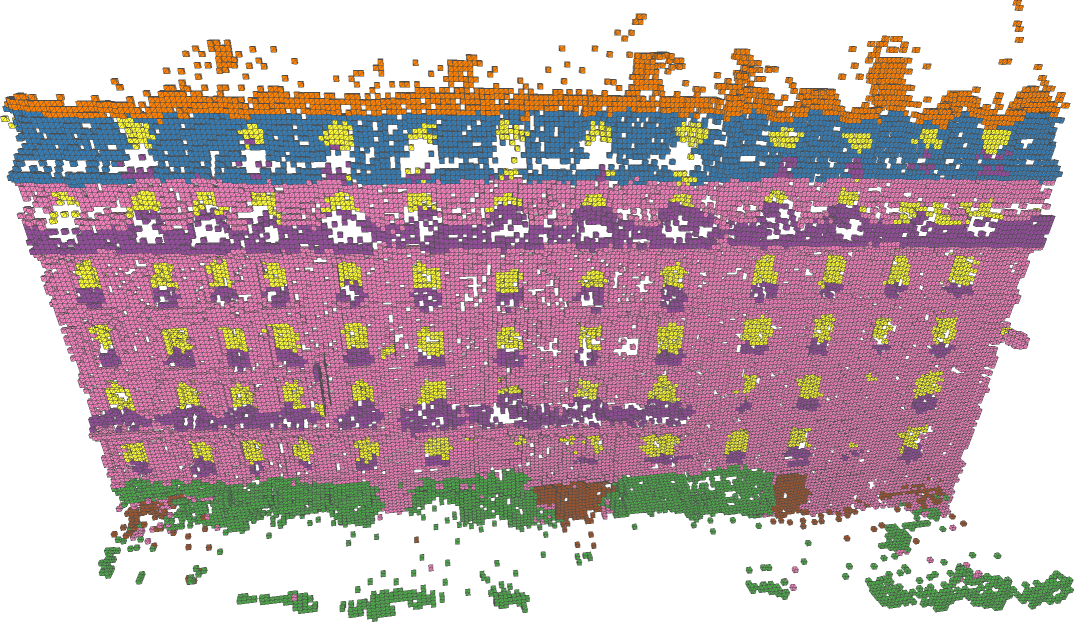}
    \caption{Voxel Estimates}
  \end{subfigure}
  \begin{subfigure}[b]{0.45\linewidth}
     \includegraphics[width=\linewidth]{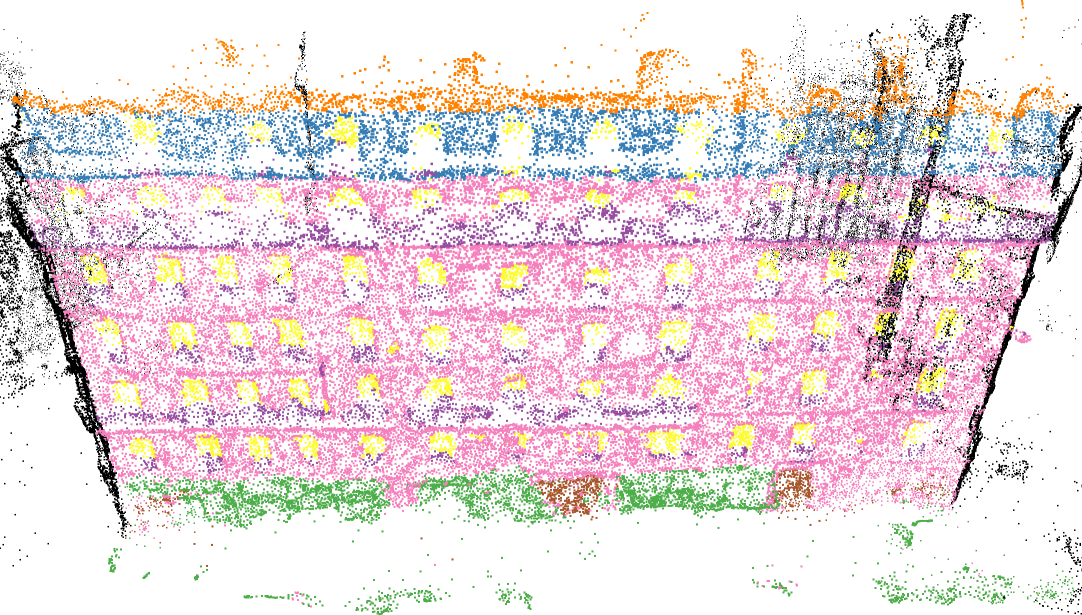}
    \caption{Estimated Point Cloud}
   \end{subfigure}
  \begin{subfigure}[b]{0.45\linewidth}
     \includegraphics[width=\linewidth]{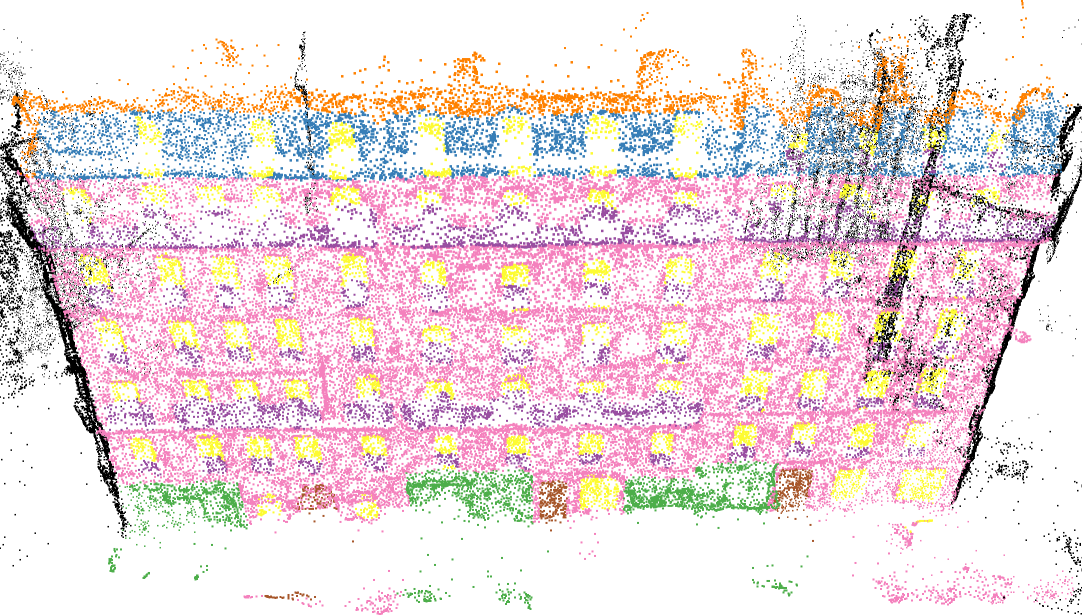}
     \caption{Ground Truth Point Cloud}
   \end{subfigure}
   \vspace{-0.5em}
   \caption{{\bf OctNet $256^3$ Facade Labeling Results.}}
   \label{fig:qualitative_varcity}
   \vspace{-1.5em}
 \end{figure}

Our results clearly show that increasing the input resolution is essential to obtain state-of-the-art results, as finer details vanish at coarser resolutions. Qualitative results for one facade are provided in \figref{fig:qualitative_varcity}. Further results are provided in the supp.\ document.

\begin{table}
  \center
  {\small
  \begin{tabular}{l c c c}
    \toprule
                                                         & Average       & Overall       & IoU \\
    \midrule
    Riemenschneider \etal \cite{Riemenschneider2014ECCV} & -             & -             & 42.3 \\
    Martinovic \etal \cite{Martinovic2015CVPR}           & -             & -             & 52.2 \\
    Gadde \etal \cite{Gadde2016ARXIV}                    & 68.5          & 78.6          & 54.4 \\
    \midrule
    OctNet $64^3$                                        & 60.0          & 73.6          & 45.6 \\
    OctNet $128^3$                                       & 65.3          & 76.1          & 50.4 \\
    OctNet $256^3$                                       & \textbf{73.6} & \textbf{81.5} & \textbf{59.2} \\
    \bottomrule
  \end{tabular}
  }
  \caption{{\bf Semantic Segmentation on RueMonge2014.}}
  \label{tab:semantic_segmentation}
   \vspace{-1.2em}
\end{table}

\section{Conclusion and Future Work}

We presented OctNet, a novel 3D representation which makes deep learning with high-resolution inputs tractable.
We analyzed the importance of high resolution inputs on several 3D learning tasks, such as object categorization, pose estimation and semantic segmentation. 
Our experiments revealed that for ModelNet10 classification low-resolution networks prove sufficient while high input (and output) resolution matters for 3D orientation estimation and 3D point cloud labeling.
We believe that as the community moves from low resolution object datasets such as ModelNet10 to high resolution large scale 3D data, OctNet will enable further improvements.
One particularly promising avenue for future research is in learning representations for multi-view 3D reconstruction where the ability to process high resolution voxelized shapes is of crucial importance.

{\small
\bibliographystyle{ieee}
\bibliography{bibliography_long,bibliography}
}

\setcounter{section}{0}
\renewcommand\thesection{\Alph{section}}
\newcommand{\suppsection}{\subsection}
\clearpage
\onecolumn
\section{Appendix}
\suppsection{Introduction}
In the following supplemental material we present details regarding our operations on the hybrid grid-octree data structure, additional experimental results and all details on the used network architectures.
In \secref{sec:data_index} we provide details on the data index used to efficiently access data in the grid-octree data structure.
\secref{sec:efficient_convolution} yields more insights into the efficient implementation of the convolution operation on octrees.
We present further quantitative and qualitative results in \secref{sec:additional_results} and in \secref{sec:network_architectures} we specify all details of the network architectures used in our experiments.

\suppsection{Data Index}
\label{sec:data_index}
An important implementation detail of the hybrid grid-octree data structure is the memory alignment for fast data access.
We store all data associated with the leaf nodes of a shallow octree in a compact contiguous array. 
Thus, we need a fast way to compute the offset in this data array for any given voxel.
In the main text we presented the following equation:
\begin{align}
  \octreedataidx(i) = \underbrace{8 \sum_{j=0}^{\octreeparent(i)-1} \octreeisset(j) + 1}_{\text{\#nodes above i}} - \underbrace{\sum_{j=0}^{i-1} \octreeisset(j)}_{\text{\#split nodes pre i}} + \underbrace{\modulo(i-1, 8)}_{\text{offset}} \,.
\end{align}
As explained in the main text the whole octree structure is stored as a bit-string and a voxel is uniquely identified by the bit index $i$, \ie, the index within the bit string.
The data is aligned breadth-first and only the leaf nodes have data associated. 
Consequently, the first part of the equation above counts the number of split and leaf nodes up to the voxel with bit index $i$.
The second term subtracts the number of split nodes before the particular voxel as data is only associated with leaf nodes. 
Finally, we need to get the offset within the voxel's neighborhood. 
This is done by the last term of the equation.

\begin{figure*}
  \center
  \begin{subfigure}[t]{0.24\linewidth}
    \includegraphics[width=\linewidth]{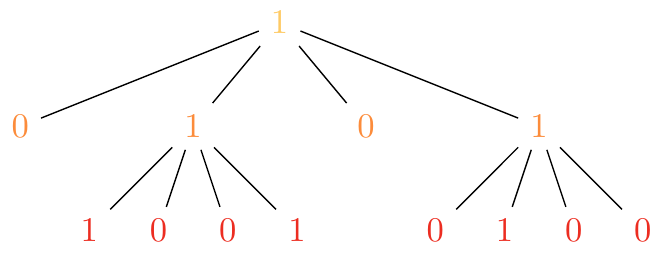}
    \caption{Bit-String}
    \label{fig:data_idx_bit_string}
  \end{subfigure}
  \begin{subfigure}[t]{0.24\linewidth}
    \includegraphics[width=\linewidth]{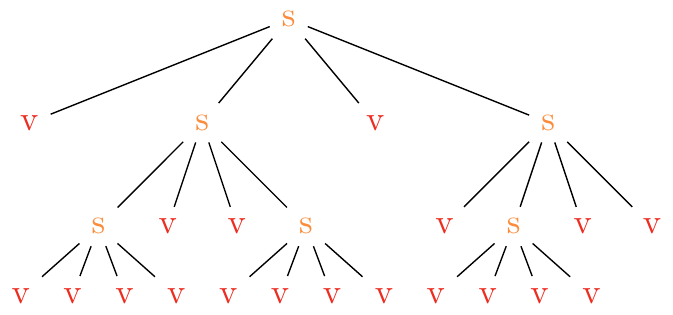}
    \caption{Split and Leaf Nodes}
    \label{fig:data_idx_split_leaf}
  \end{subfigure}
  \begin{subfigure}[t]{0.24\linewidth}
    \includegraphics[width=\linewidth]{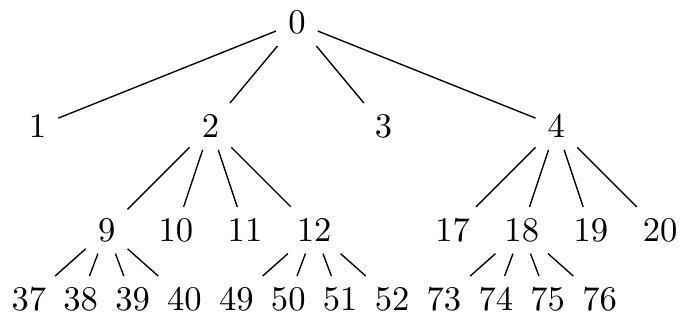}
    \caption{Bit Index}
    \label{fig:data_idx_bit_idx}
  \end{subfigure}
  \begin{subfigure}[t]{0.24\linewidth}
    \includegraphics[width=\linewidth]{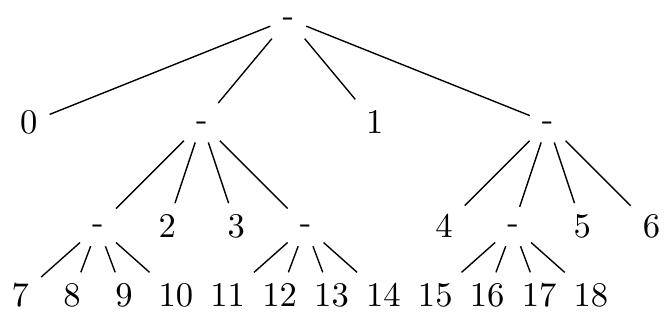}
     \caption{Data Index}
    \label{fig:data_ids_data_idx}
  \end{subfigure}
  \caption{{\bf Data Index.}}
  \label{fig:data_idx}
\end{figure*}

Let us illustrate this with a simple example:
For ease of visualization we will consider a quadtree.
Hence, each voxel can be split into $4$ instead of $8$ children. 
The equation for the offset changes to
\begin{align}
  \octreedataidx_4(i) = \underbrace{4 \sum_{j=0}^{\octreeparent_4(i)-1} \octreeisset(j) + 1}_{\text{\#nodes above i}} - \underbrace{\sum_{j=0}^{i-1} \octreeisset(j)}_{\text{\#split nodes pre i}} + \underbrace{\modulo(i-1, 4)}_{\text{offset}} \label{eq:data_idx_4}\,,
\end{align}
with 
\begin{align}
  \octreeparent_4(i) &= \left\lfloor\frac{i - 1}{4}\right\rfloor \label{eq:parent_4}\,. 
\end{align}
Now consider the following bit string for instance: $1\,0101\,0000\,1001\,0000\,0100$.
According to our definition, this bit string corresponds to the tree structure visualized in \figref{fig:data_idx_bit_string} and \ref{fig:data_idx_split_leaf}, where \textit{s} indicates a split node and \textit{v} a leaf node with associated data.
In \figref{fig:data_idx_bit_idx} we show the bit indices for all nodes. 
Note that the leaf nodes at depth $3$ do not need to be stored in the bit string as this information is implicit.
Finally, the data index for all leaf nodes is visualized in \figref{fig:data_ids_data_idx}.
Now we can verify equation~\eqref{eq:data_idx_4} using a simple example.
Assume the bit index $51$:
The parent bit index is given by equation~\eqref{eq:parent_4} as $12$.
To compute the data index we first count the number of nodes before $49$ as it is the first node within its siblings (first term of equation), which is $17$.
Next, we count the number of split nodes up to $49$ (second term of equation), which is $6$.
Finally, we look up the position $51$ within its siblings (last term of equation), which is $2$.
Combining those three terms yields the data index $17-6+2=13$.

\begin{figure*}
  \center
  \begin{subfigure}[b]{0.24\linewidth}
    \center
    \includegraphics[height=0.175\textheight]{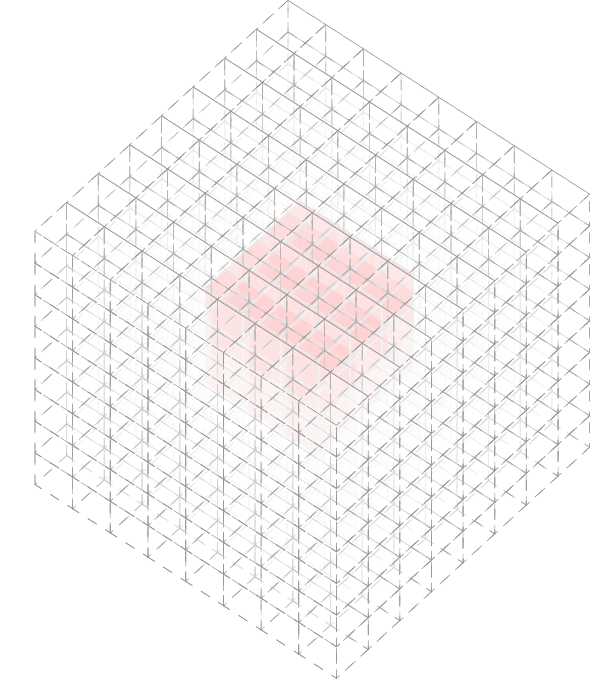}
    \caption{Constant}
    \label{fig:eff_conv_const}
  \end{subfigure}
  \begin{subfigure}[b]{0.24\linewidth}
    \center
    \includegraphics[height=0.175\textheight]{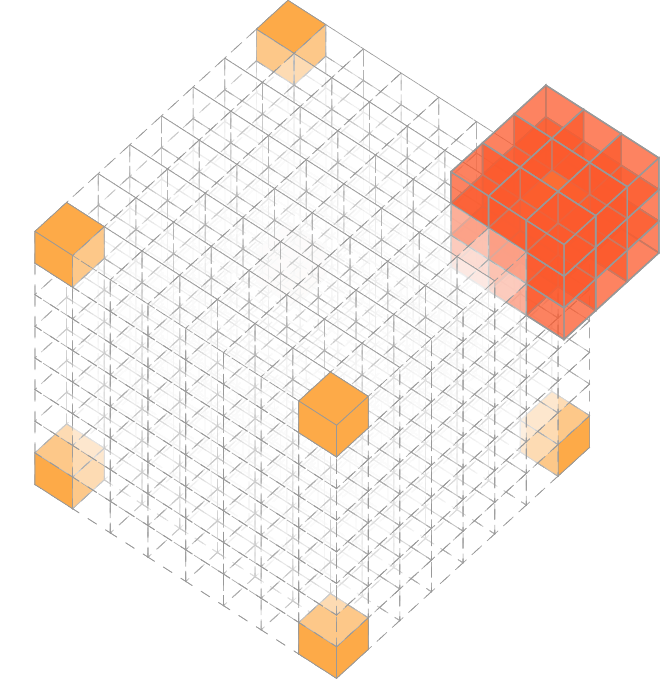}
    \caption{Corners}
    \label{fig:eff_conv_corner}
  \end{subfigure}
  \begin{subfigure}[b]{0.24\linewidth}
    \center
    \includegraphics[height=0.175\textheight]{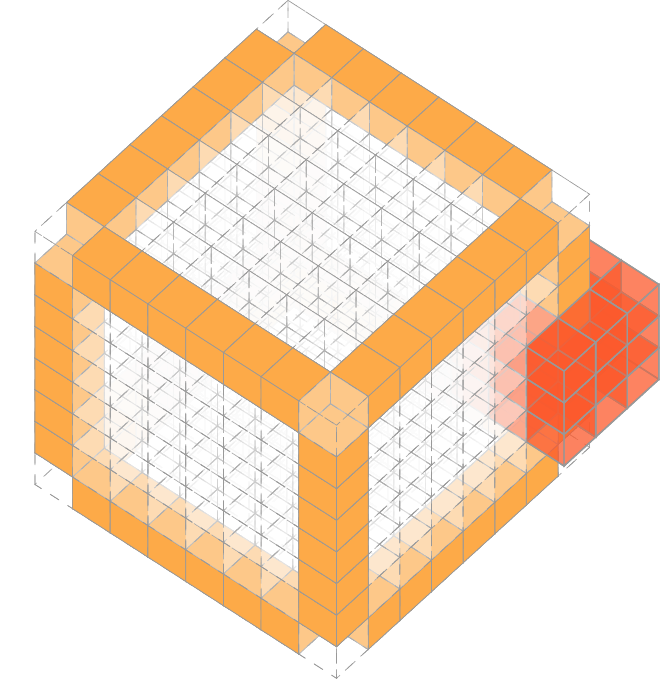}
    \caption{Edges}
    \label{fig:eff_conv_edge}
  \end{subfigure}
  \begin{subfigure}[b]{0.24\linewidth}
    \center
    \includegraphics[height=0.175\textheight]{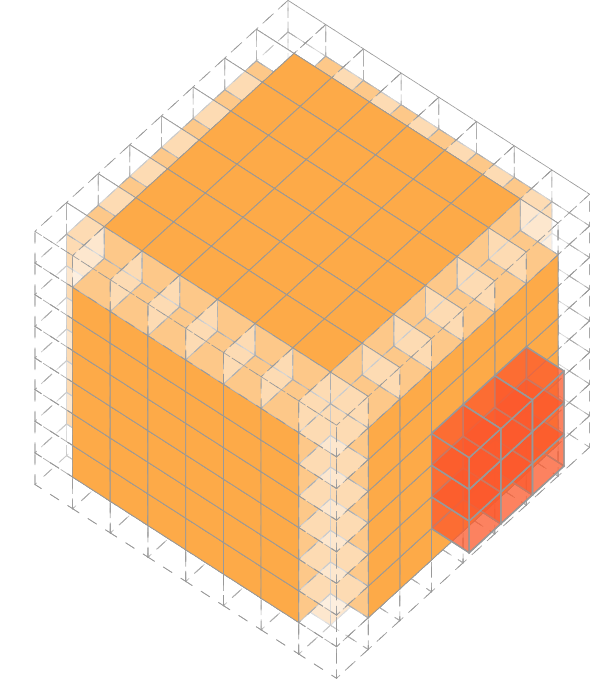}
     \caption{Faces}
    \label{fig:eff_conv_face}
  \end{subfigure}
  \caption{{\bf Efficient Convolution.}}
  \label{fig:efficient_conv}
\end{figure*}

\suppsection{Efficient Convolution}
\label{sec:efficient_convolution}
In the main text we discussed that the convolution for larger octree cells and small convolution kernels can be efficiently implemented.
A na\"{i}ve implementation applies the convolution kernel at every location $(i,j,k)$ comprised by the cell $\Omega[i,j,k]$.
Therefore, for an octree cell of size $8^3$ and a convolution kernel kernel of $3^3$ this would require $8^3 \cdot 3^3 =13,824$ multiplications.
However, we can implement this calculation much more efficiently as depicted in \figref{fig:efficient_conv}.
We observe that the value inside the cell of size $8^3$ is constant. 
Thus, we only need to evaluate the convolution once inside this cell and multiply the result with the size of the cell $8^3$, see \figref{fig:eff_conv_const}.
Additionally, we only need to evaluate a truncated versions of the kernel on the corners, edges and faces of the voxel, see Fig.~\ref{fig:eff_conv_corner}-d.
This implementation is more efficient, as we need only $27$ multiplications for the constant part, $8 \cdot 19$ multiplications for the corners, $12 \cdot 6 \cdot 15$ multiplications for the edges, and $6 \cdot 6^2 \cdot 9$ multiplications for the faces of the voxel.
In total this yields $3203$ multiplications, or $23.17\%$ of the multiplications required by the na\"{i}ve implementation.

\suppsection{Additional Results}
\label{sec:additional_results}
In this Section we show additional quantitative and qualitative results for 3D shape classification, 3D orientation estimation and semantic 3D point labeling.

\subsection{3D Classification}
\begin{figure*}
  \center
  \begin{subfigure}[b]{0.24\linewidth}
    \includegraphics[width=\linewidth]{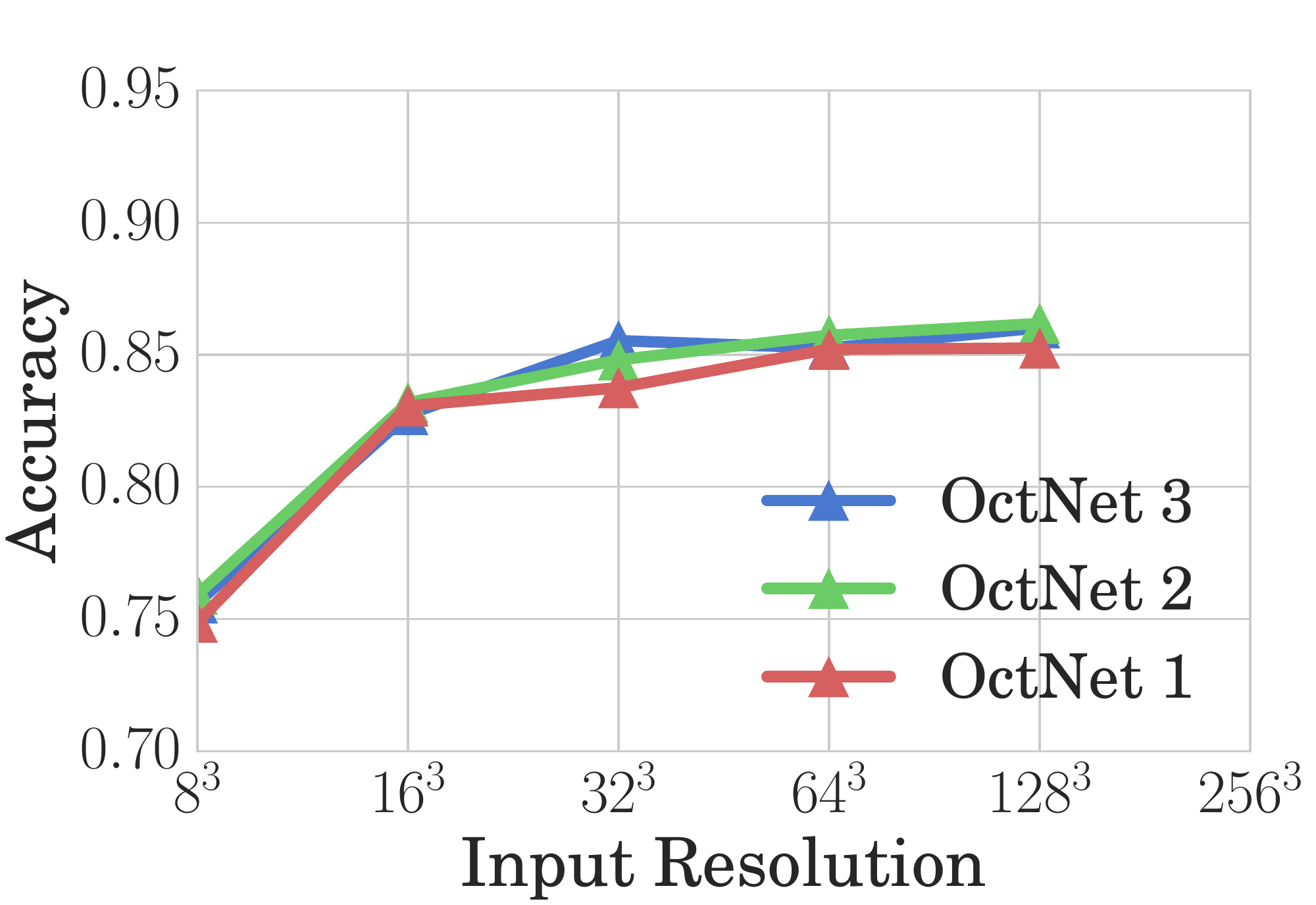}
    \caption{Accuracy}
  \end{subfigure}
  \begin{subfigure}[b]{0.24\linewidth}
    \includegraphics[width=\linewidth]{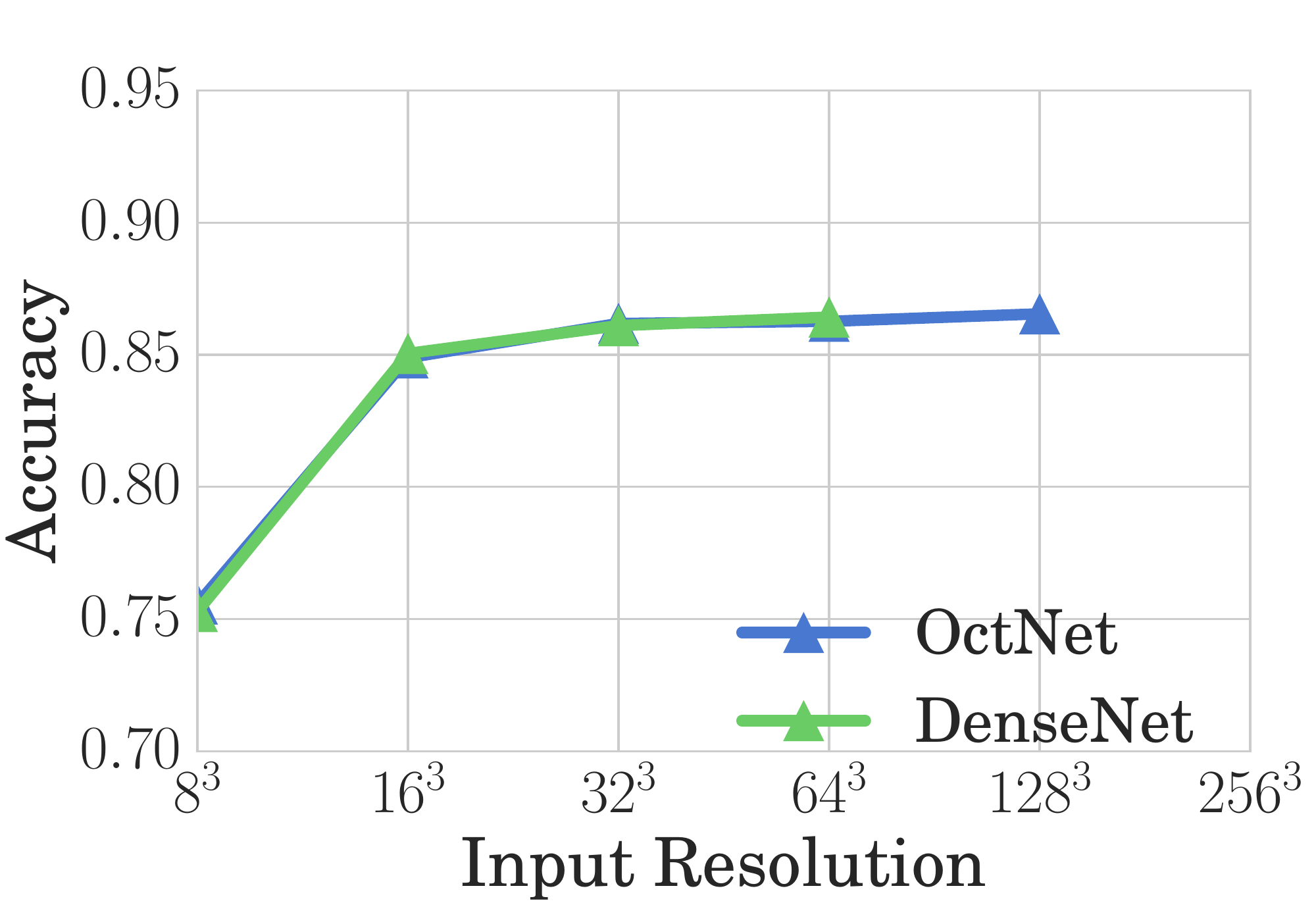}
    \caption{Accuracy}
  \end{subfigure}
  \begin{subfigure}[b]{0.24\linewidth}
    \includegraphics[width=\linewidth]{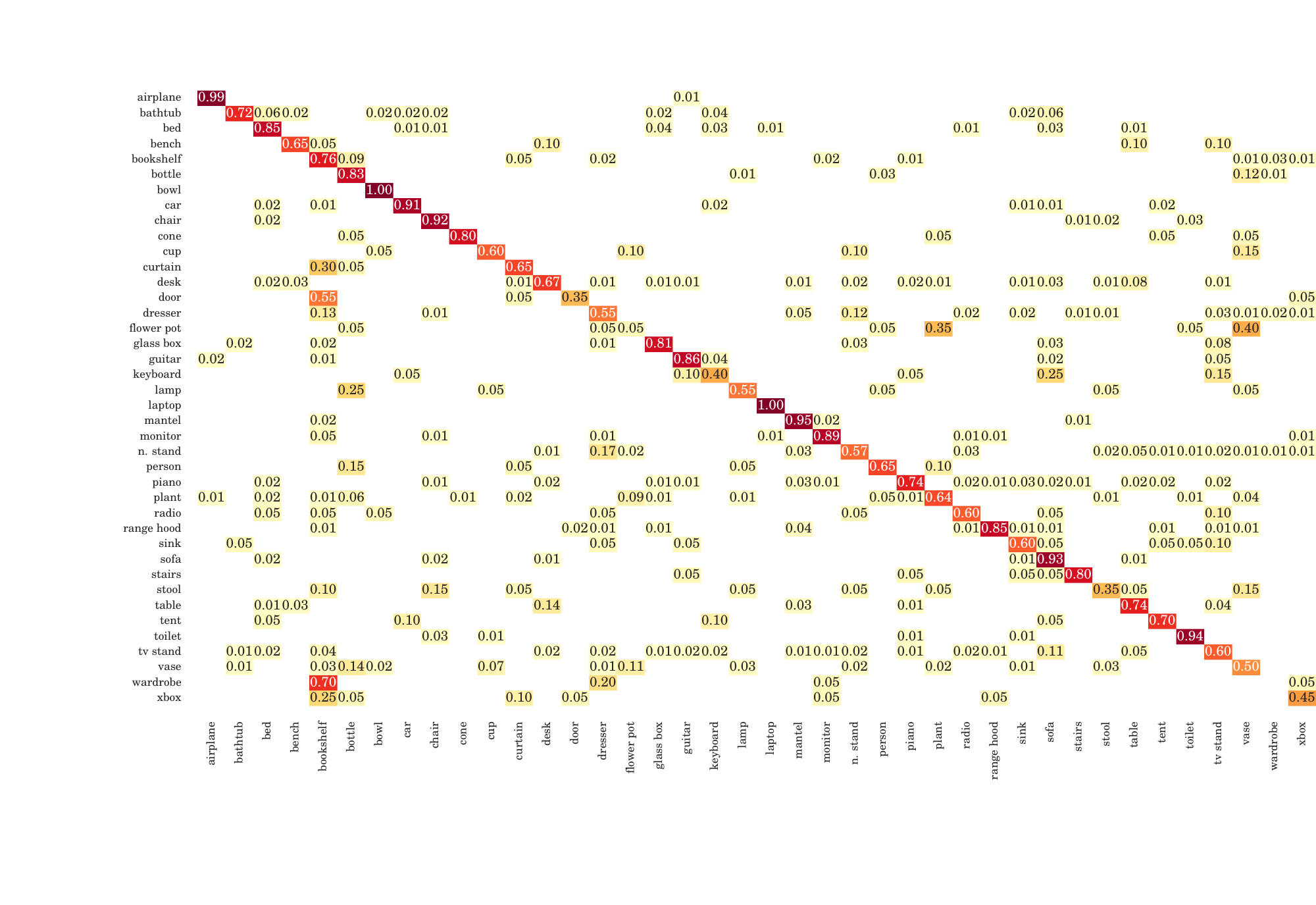}
    \caption{Confusion Matrix $8^3$}
  \end{subfigure}
  \begin{subfigure}[b]{0.24\linewidth}
    \includegraphics[width=\linewidth]{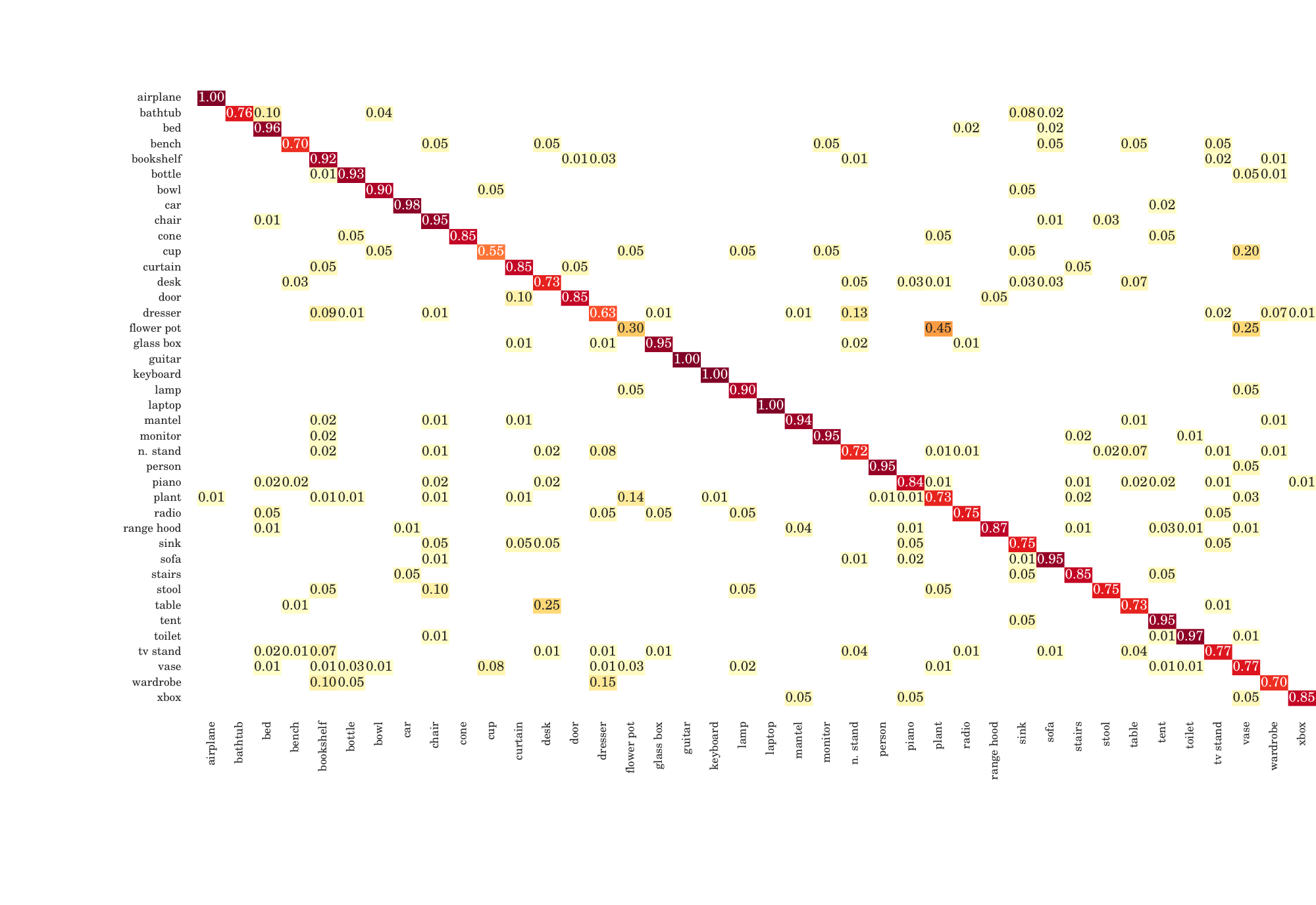}
     \caption{Confusion Matrix $32^3$}
  \end{subfigure}
  \caption{{\bf ModelNet40 results.}}
  \label{fig:modelnet_40}
\end{figure*}

In the main text of our work we analyzed the runtime and memory consumption of OctNet compared with the equivalent dense networks on ModelNet10~\cite{Wu2015CVPR}. 
Additionally, we demonstrated that without further data augmentation, ensemble learning, or more sophisticated architectures the accuracy saturates at an input resolution of about $16^3$, when keeping the number of network parameters fixed across all resolutions.
In this Section we show the same experiment on ModelNet40~\cite{Wu2015CVPR}.
The results are summarized in \figref{fig:modelnet_40}.
In contrast to ModelNet10, we see an increase in accuracy up to an input resolution of $32^3$.
Beyond this resolution the classification performance does not further improve.
Note that the only form of data augmentation we used in this experiment was rotation around the up-vector as the 3D models in this dataset vary in pose. We conclude that object classification on the ModelNet40 dataset is more challenging than on the ModelNet10 dataset, but both datasets are relatively easy in the sense that details do not matter as much as in the datasets used for our other experiments.

\begin{figure}
\begin{subfigure}[b]{0.32\linewidth}
  \includegraphics[width=\linewidth]{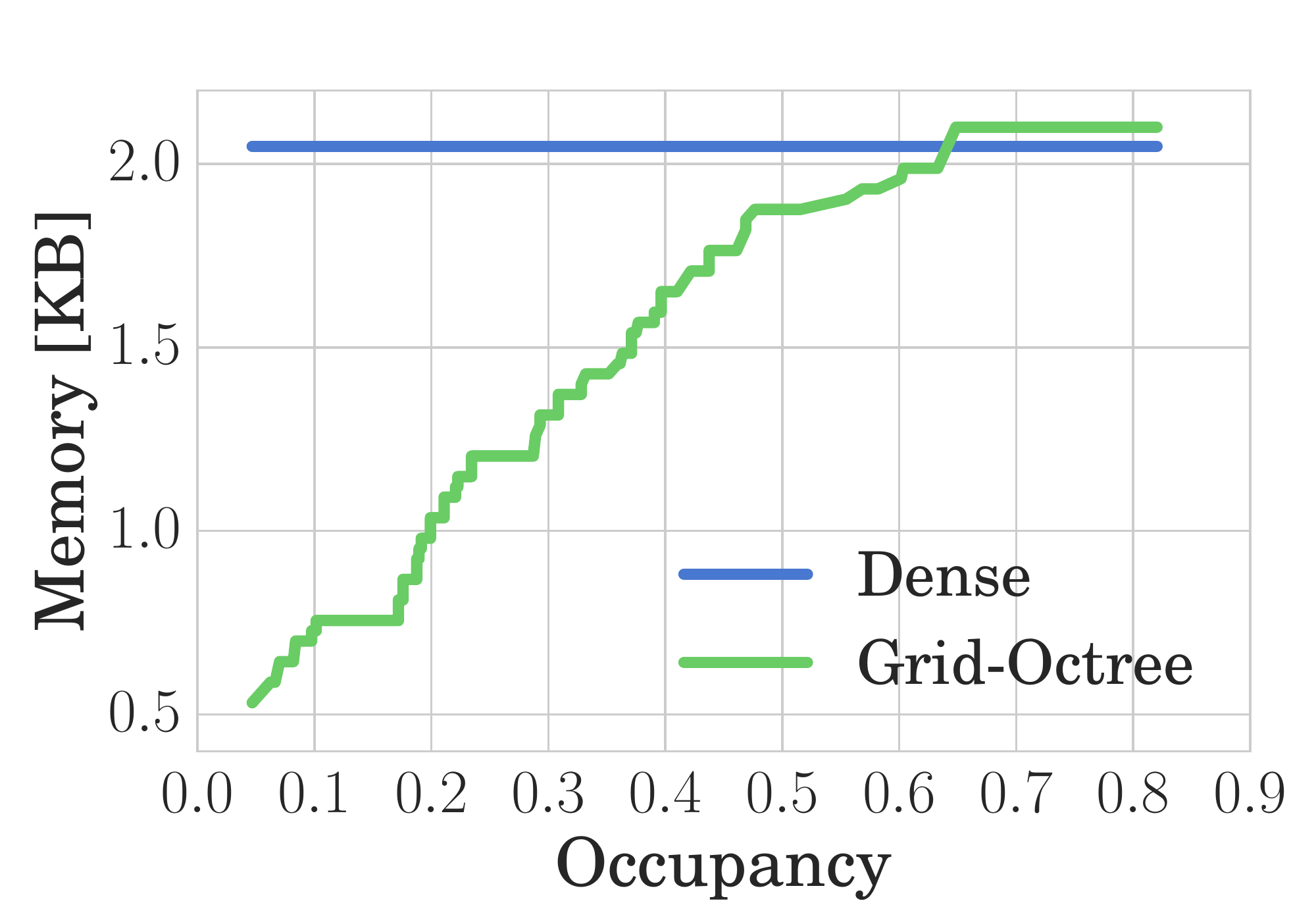}
  \caption*{{$8^3$}}
\end{subfigure}
\begin{subfigure}[b]{0.32\linewidth}
  \includegraphics[width=\linewidth]{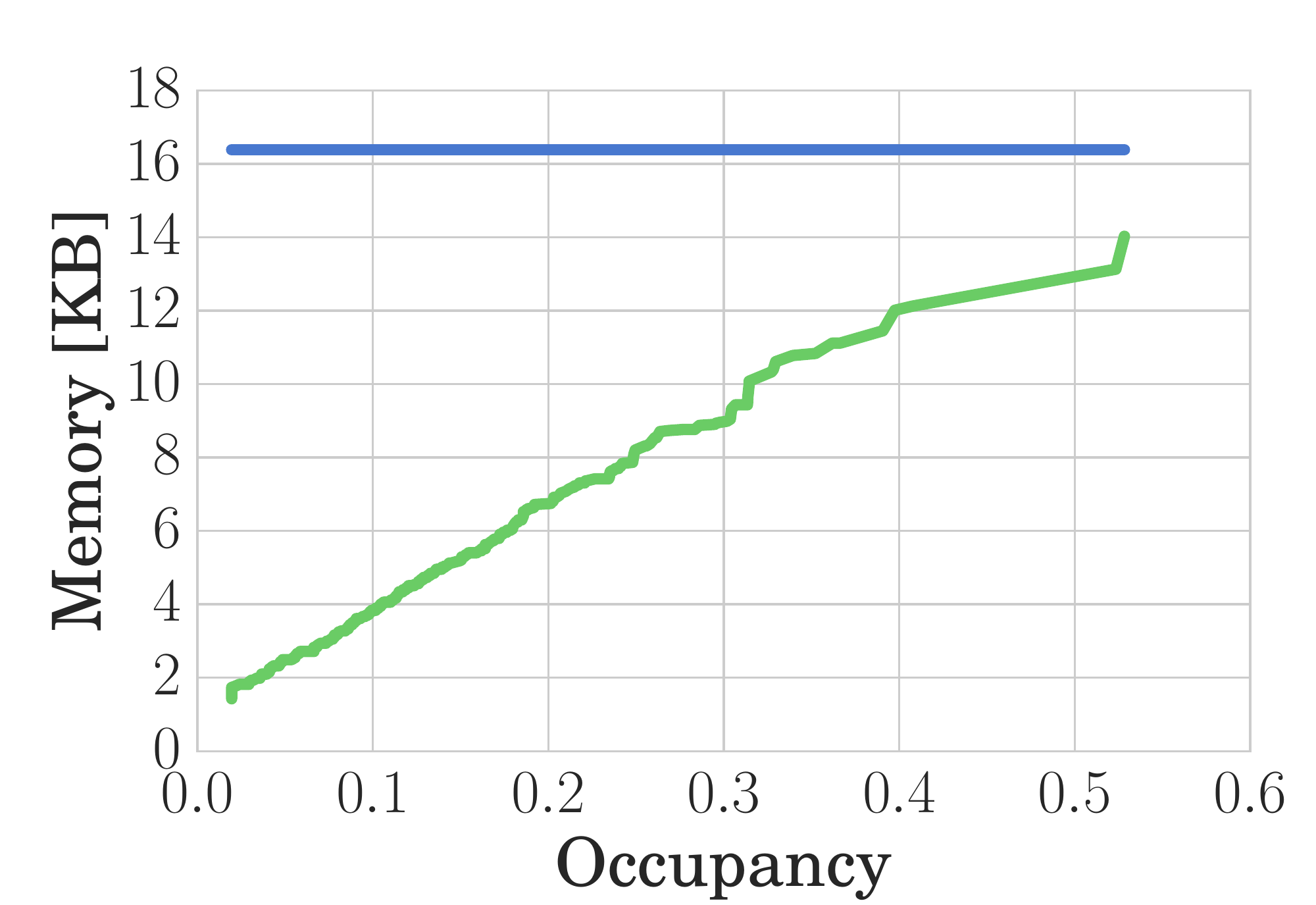}
  \caption*{{$16^3$}}
\end{subfigure}
\begin{subfigure}[b]{0.32\linewidth}
  \includegraphics[width=\linewidth]{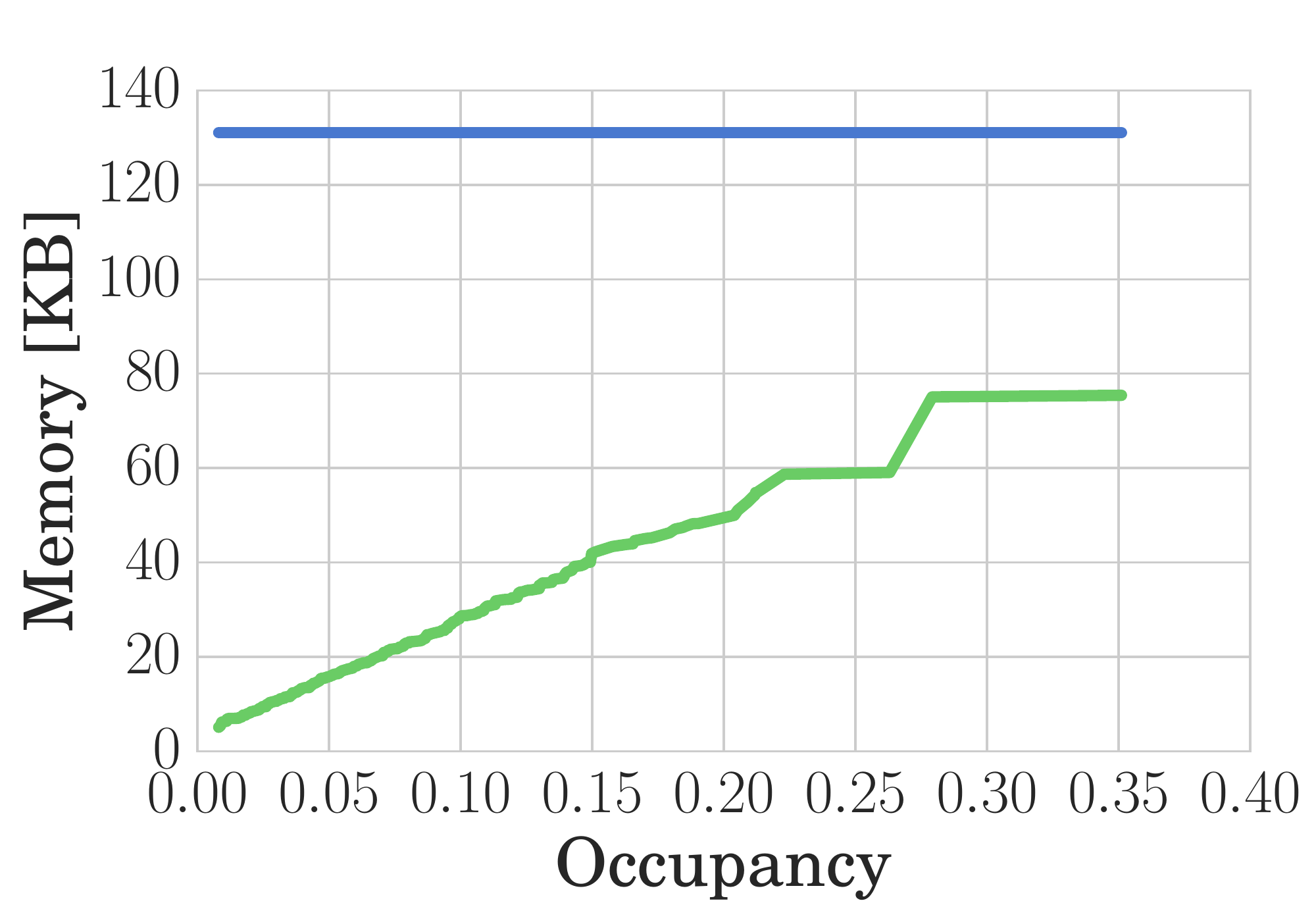}
  \caption*{{$32^3$}}
\end{subfigure}
\begin{subfigure}[b]{0.32\linewidth}
  \includegraphics[width=\linewidth]{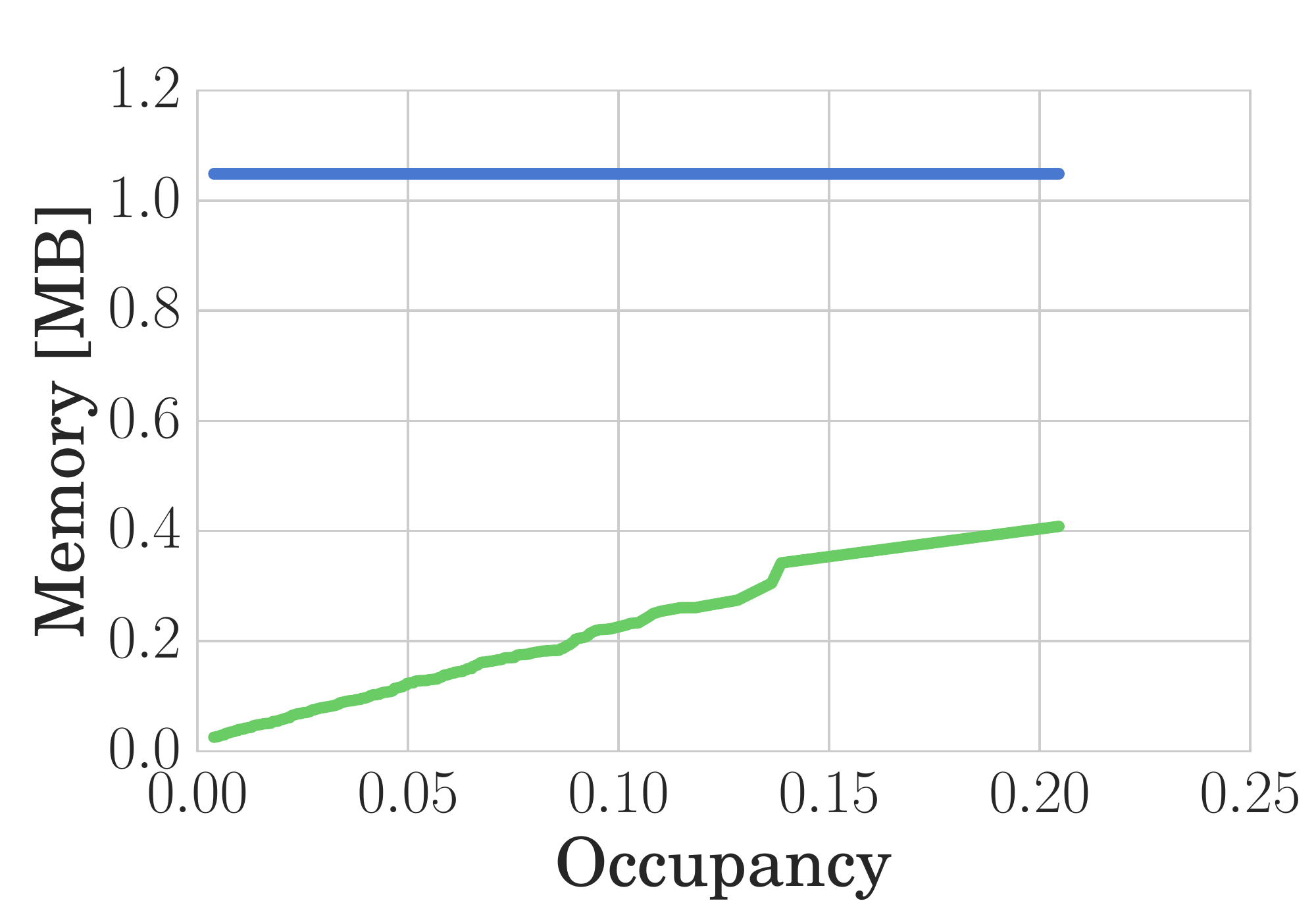}
  \caption*{{$64^3$}}
\end{subfigure}
\begin{subfigure}[b]{0.32\linewidth}
  \includegraphics[width=\linewidth]{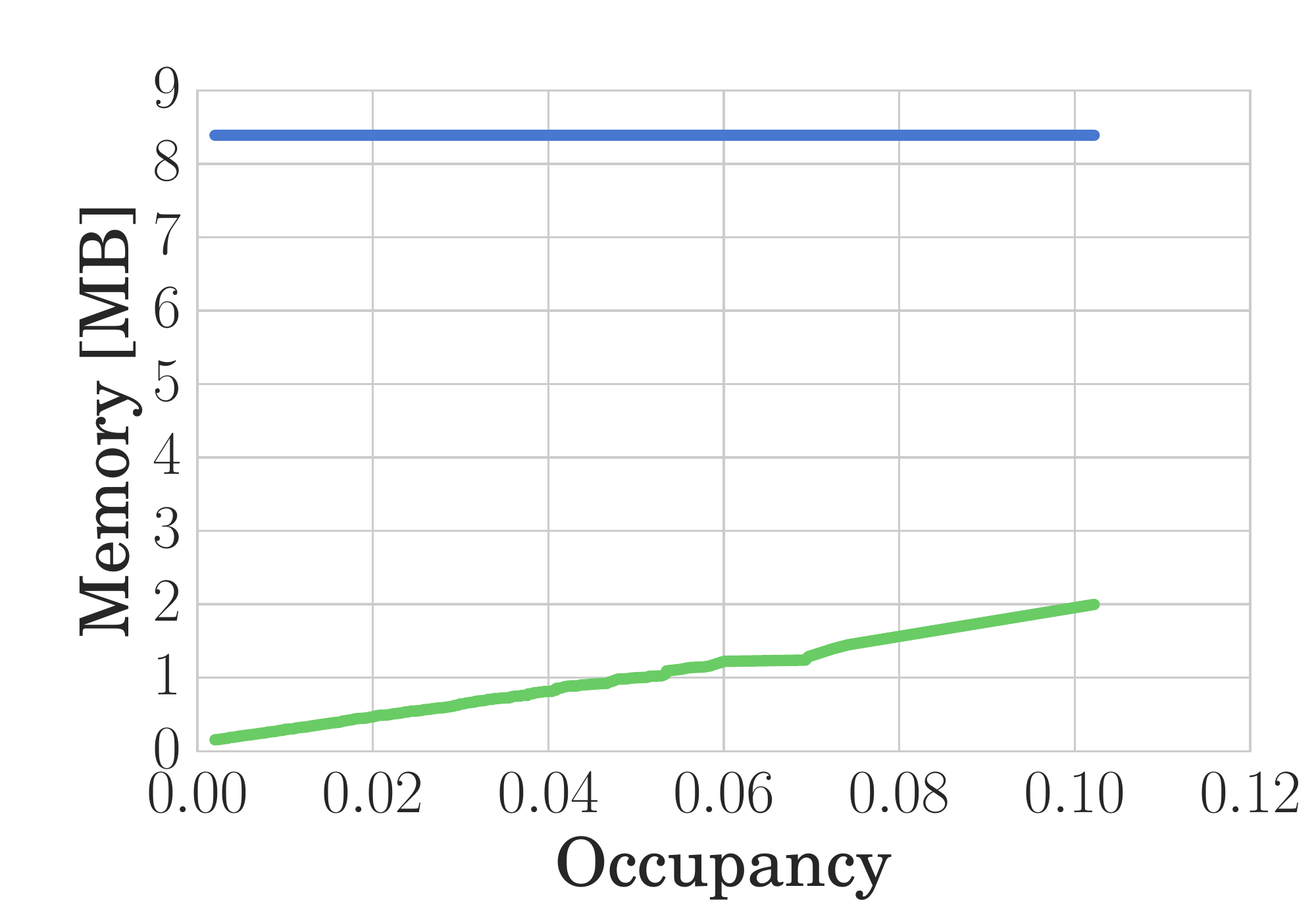}
  \caption*{{$128^3$}}
\end{subfigure}
\begin{subfigure}[b]{0.32\linewidth}
  \includegraphics[width=\linewidth]{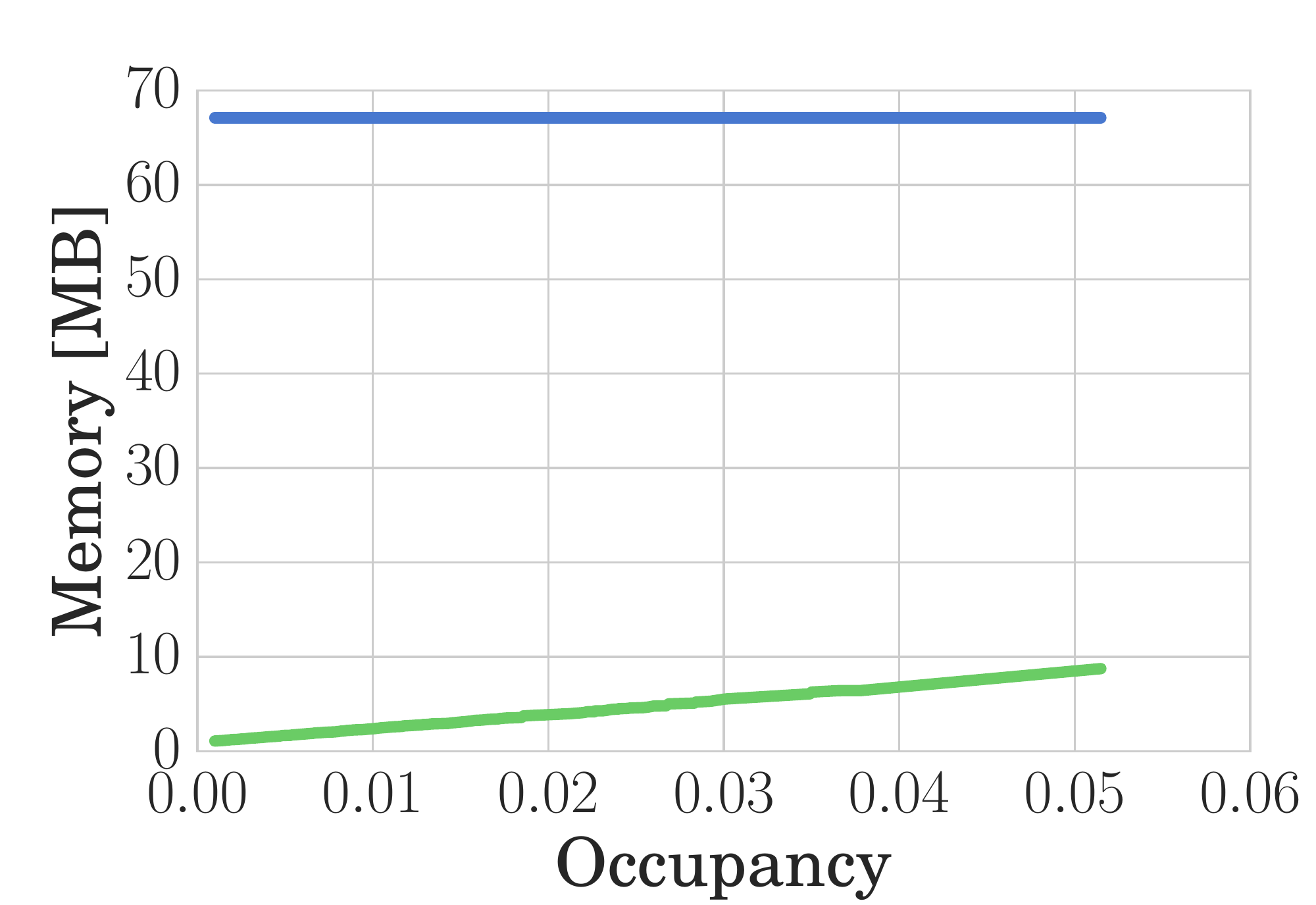}
  \caption*{{$256^3$}}
\end{subfigure}
\caption{
{\bf Memory consumption vs.\ occupancy.}}
\label{fig:mem_sparsity}
\end{figure}

We further use this experiment to investigate the question at which level of sparsity OctNet becomes useful.
To answer this, we visualize the memory consumption of a dense representation vs.\ our data structure at different resolutions and occupancies by sampling 500 shapes from the ModelNet40 dataset (see Fig.~\ref{fig:mem_sparsity}).
It can be observed that even at very low resolutions ($8^3$), our data structure is superior compared to the dense representation, even up to an occupancy level of $50\%$.
As the voxel resolution increases, the occupancy levels (x-axis) decreases since the data becomes sparser.
Our OctNet exploits this sparsity to achieve a significant reduction in memory consumption. 

\subsection{3D Orientation Estimation}
To demonstrate that OctNet can handle input resolutions larger than $256^3$ we added results for the 3D orientation estimation experiment with an input resolution of $512^3$.
The results are presented in \figref{fig:mn_chair_512}.
As for the orientation experiment in the main paper, we can observe the trend that performance increases with increasing input resolution.

\begin{figure*}
  \center
  \begin{subfigure}[b]{0.19\linewidth}
    \center
    \includegraphics[height=0.1\textheight]{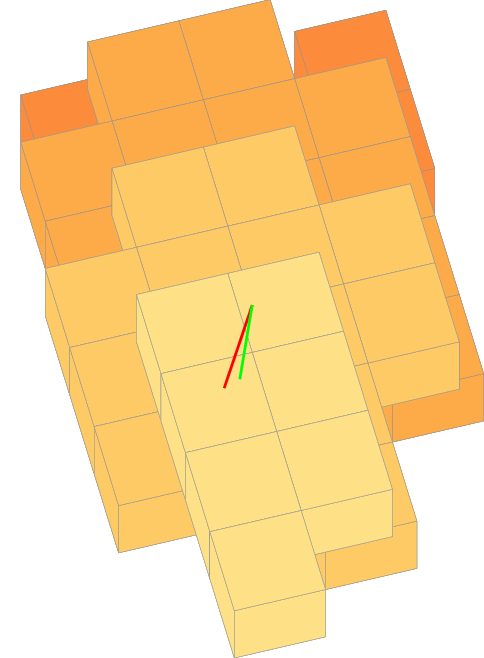}
    \includegraphics[height=0.1\textheight]{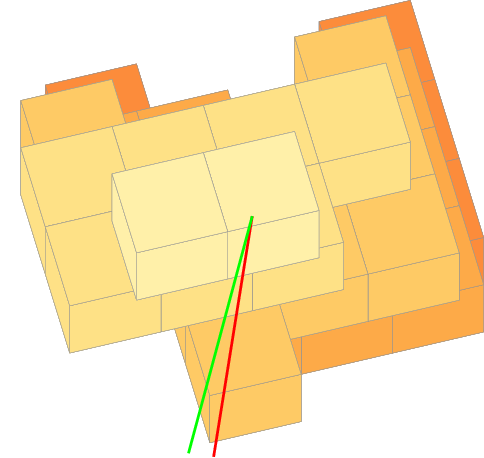}
    \includegraphics[height=0.1\textheight]{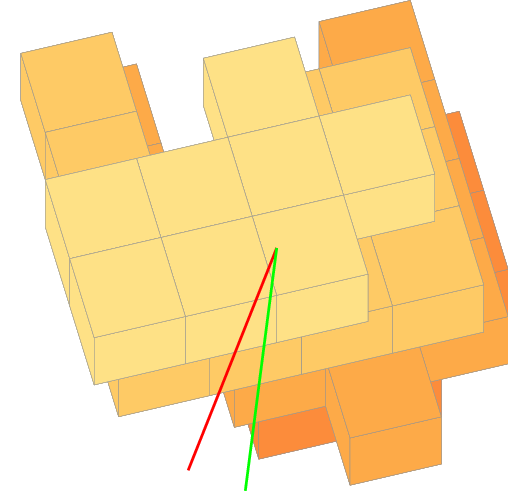}
    \caption{$8^3$}
  \end{subfigure}
  \begin{subfigure}[b]{0.19\linewidth}
    \center
    \includegraphics[height=0.1\textheight]{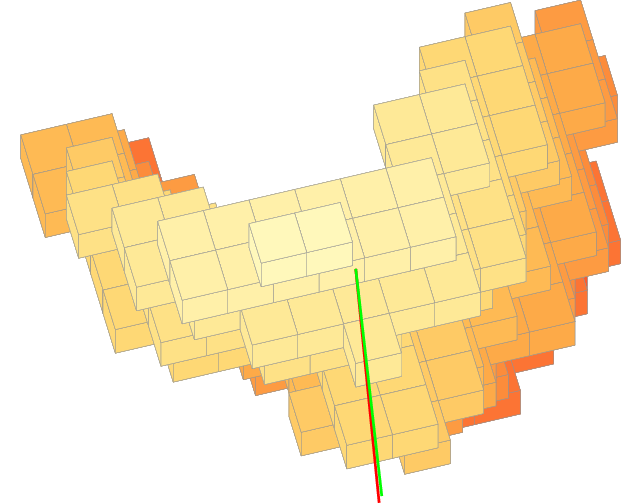}
    \includegraphics[height=0.1\textheight]{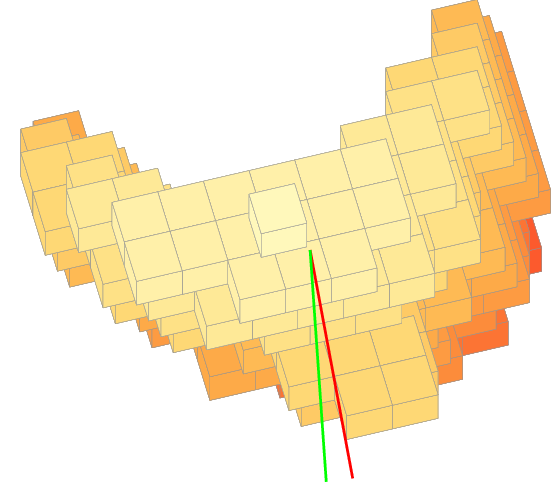}
    \includegraphics[height=0.1\textheight]{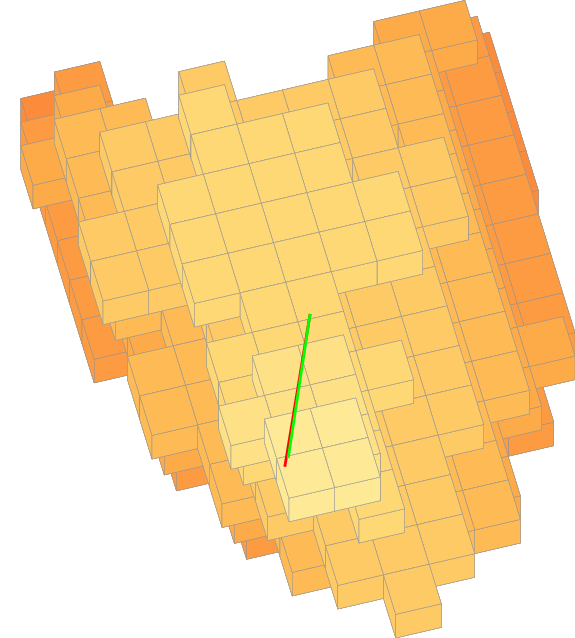}
    \caption{$16^3$}
  \end{subfigure}
  \begin{subfigure}[b]{0.19\linewidth}
    \center
    \includegraphics[height=0.1\textheight]{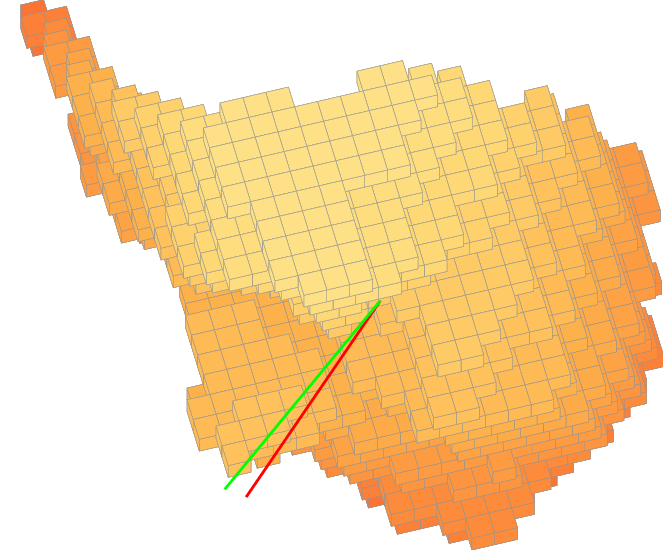}
    \includegraphics[height=0.1\textheight]{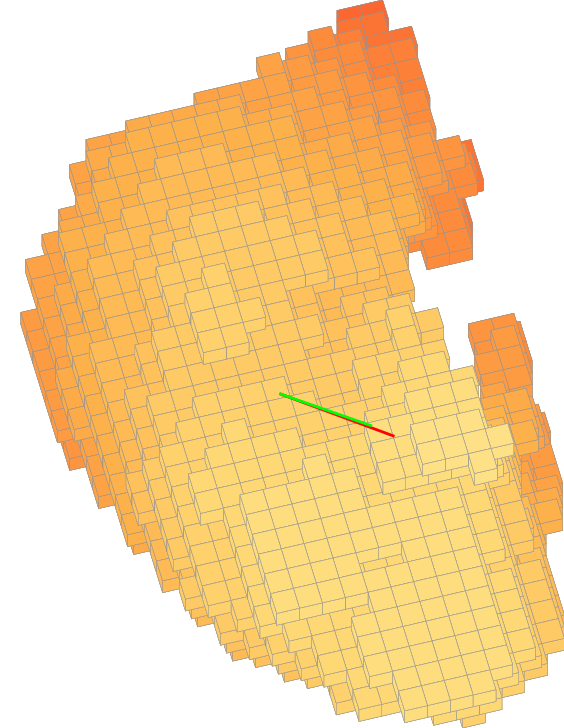}
    \includegraphics[height=0.1\textheight]{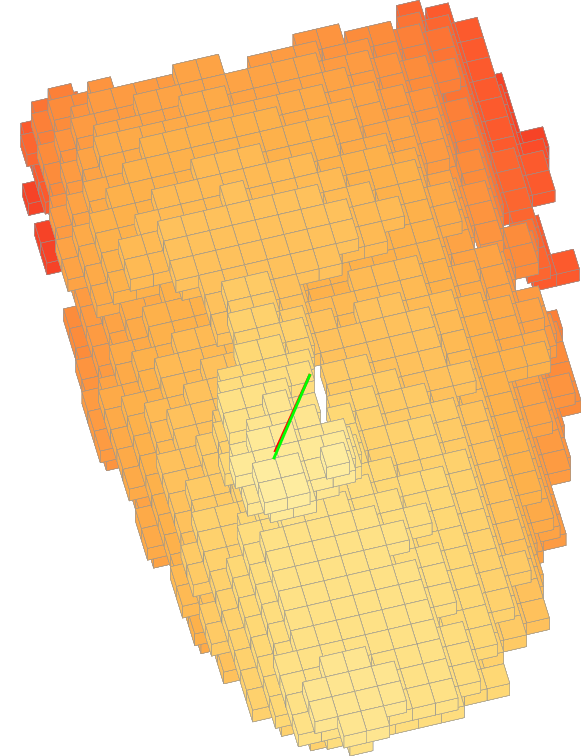}
    \caption{$32^3$}
  \end{subfigure}
  \begin{subfigure}[b]{0.19\linewidth}
    \center
    \includegraphics[height=0.1\textheight]{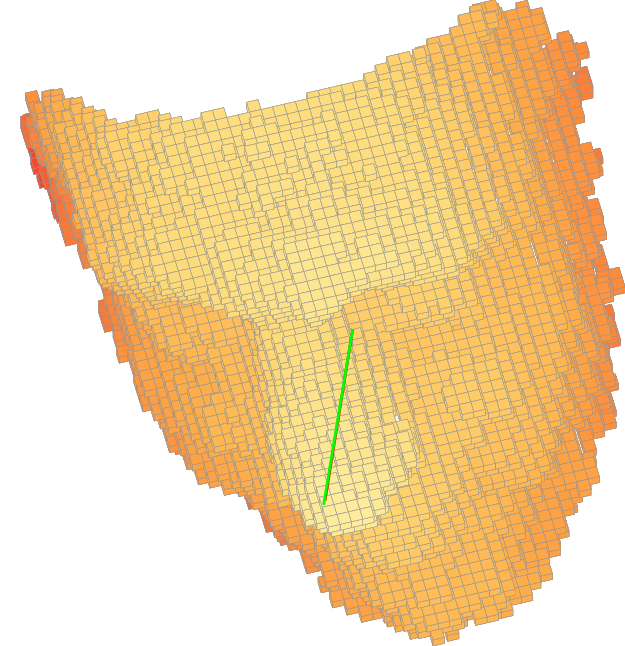}
    \includegraphics[height=0.1\textheight]{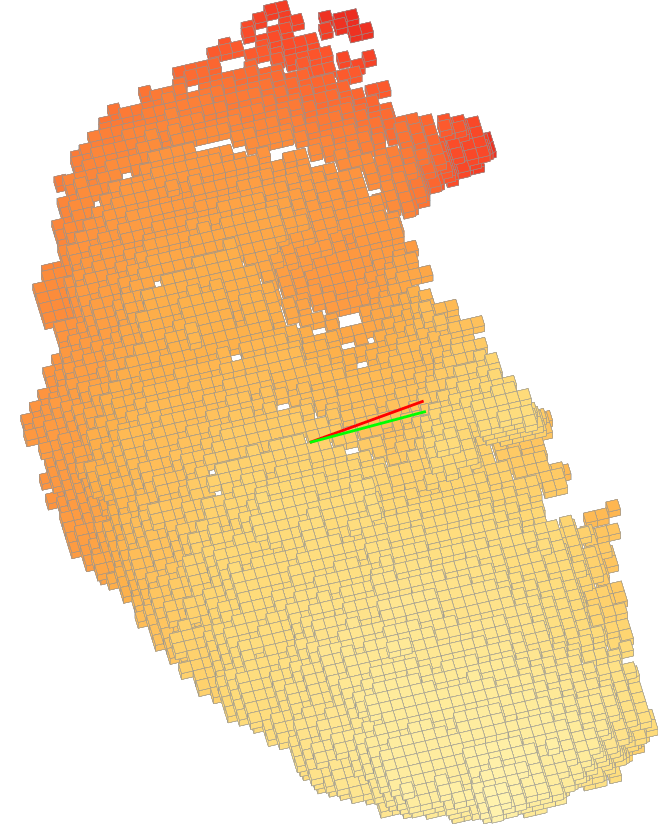}
    \includegraphics[height=0.1\textheight]{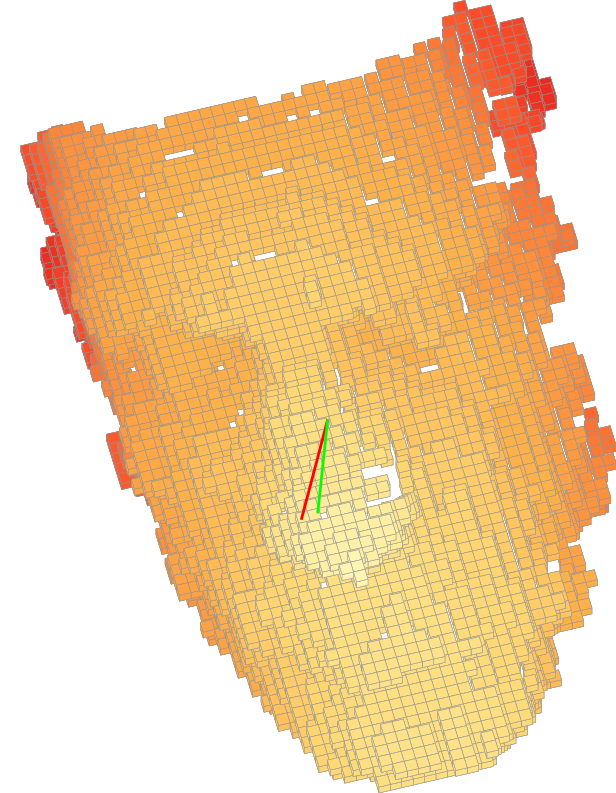}
    \caption{$64^3$}
  \end{subfigure}
  \begin{subfigure}[b]{0.19\linewidth}
    \center
    \includegraphics[height=0.1\textheight]{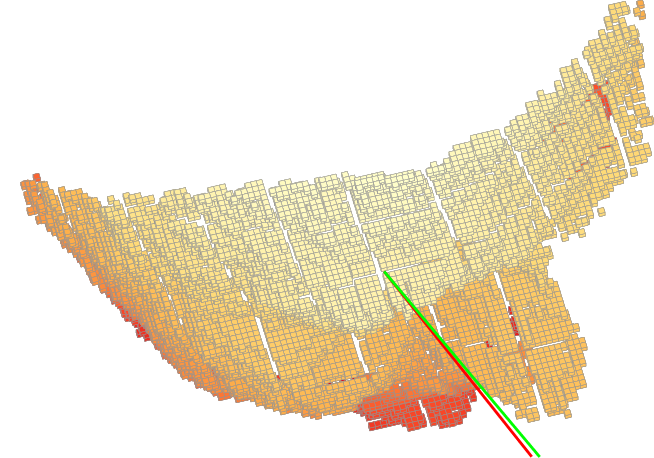}
    \includegraphics[height=0.1\textheight]{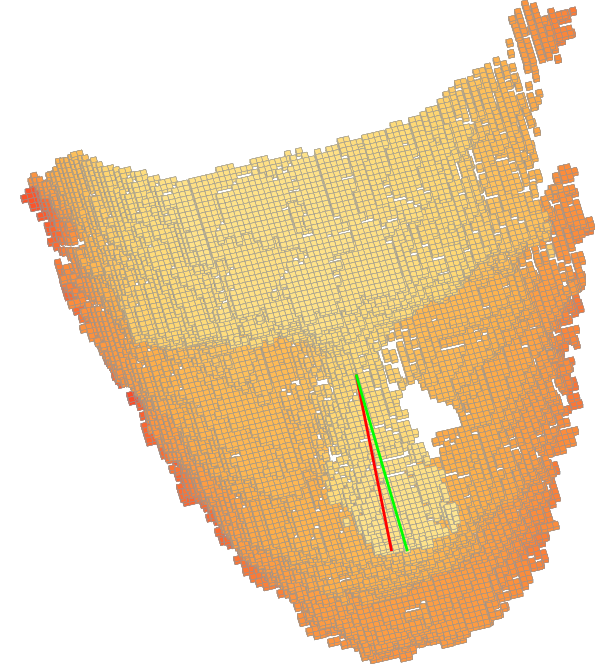}
    \includegraphics[height=0.1\textheight]{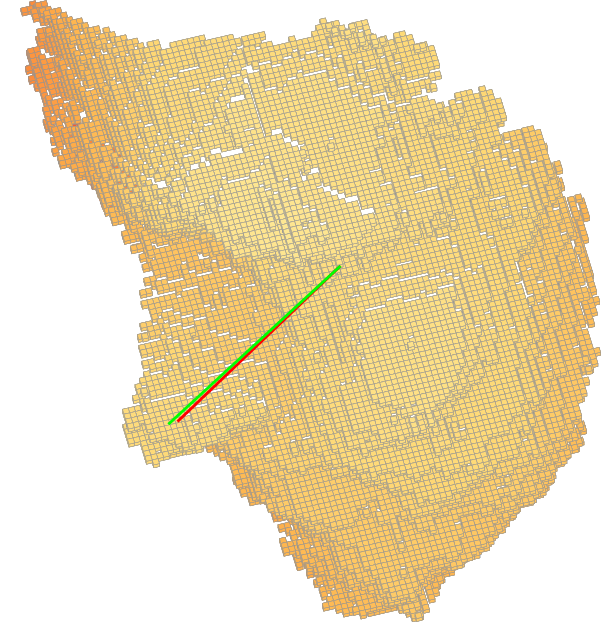}
    \caption{$128^3$}
  \end{subfigure}
  \caption{{\bf Qualitative Results for Head Pose Estimation.}}
  \label{fig:qualitative_headpose}
\end{figure*}

We evaluated our OctNet also on the Biwi Kinect Head Pose Database~\cite{fanelli_IJCV} as an additional experiment on 3D pose estimation.
The dataset consists of $24$ sequences of $20$ individuals sitting in front of a Kinect depth sensor.
For each frame the head center and the head pose in terms of its 3D rotation is annotated. 
We split the dataset into a training set of $18$ individuals for training and $2$ individuals for testing and project the depth map to 3D points with the given camera parameters.
We then create the hybrid grid-octree structure from the 3D points that belong to the head (In this experiment we are only interested in 3D orientation estimation, as the head can be reliable detected in the color images).
As in the previous experiment we parameterize the orientation with unit quaternions and train our OctNet using the same settings as in the previous 3D orientation estimation experiment.
\figref{fig:headpose} shows the quantitative results over varying input resolutions. 
We see a reasonable improvement of accuracy from $8^3$ up to $64^3$. 
Beyond this input resolution the octree resolution becomes finer than the resolution of the 3D point cloud.
Thus, further improvements can not be expected.
In \figref{fig:qualitative_headpose} we show some qualitative results.

\begin{minipage}{0.49\textwidth}
\begin{figure}[H]
  \center
  \includegraphics[width=0.49\linewidth]{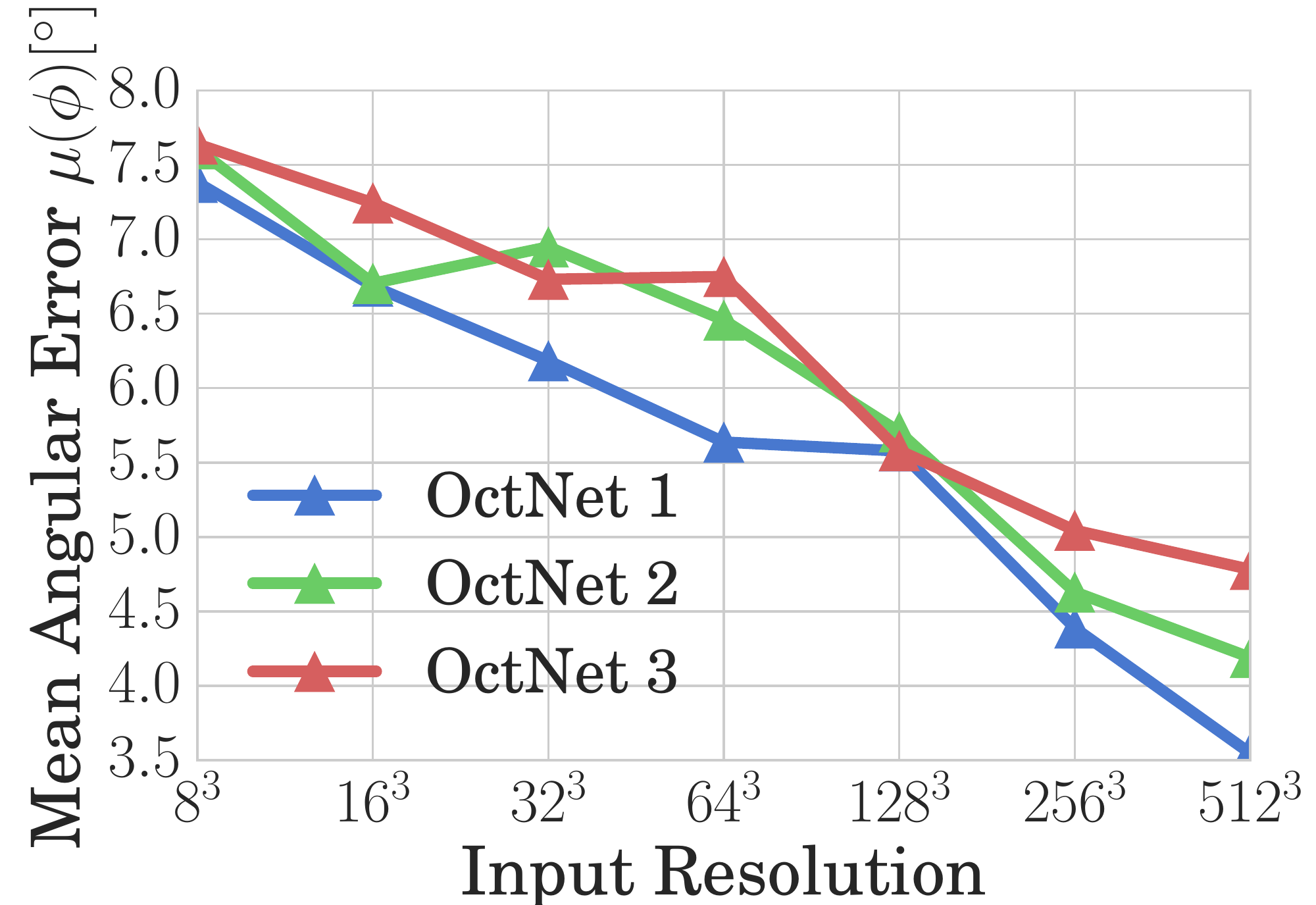}
  \caption{
    {\bf Additional Chair Orientation Results.}
  }
  \label{fig:mn_chair_512}
\end{figure}
\end{minipage}%
\hfill
\begin{minipage}{0.49\textwidth}
\begin{figure}[H]
  \center
  \begin{subfigure}[b]{0.49\linewidth}
    \includegraphics[width=\linewidth]{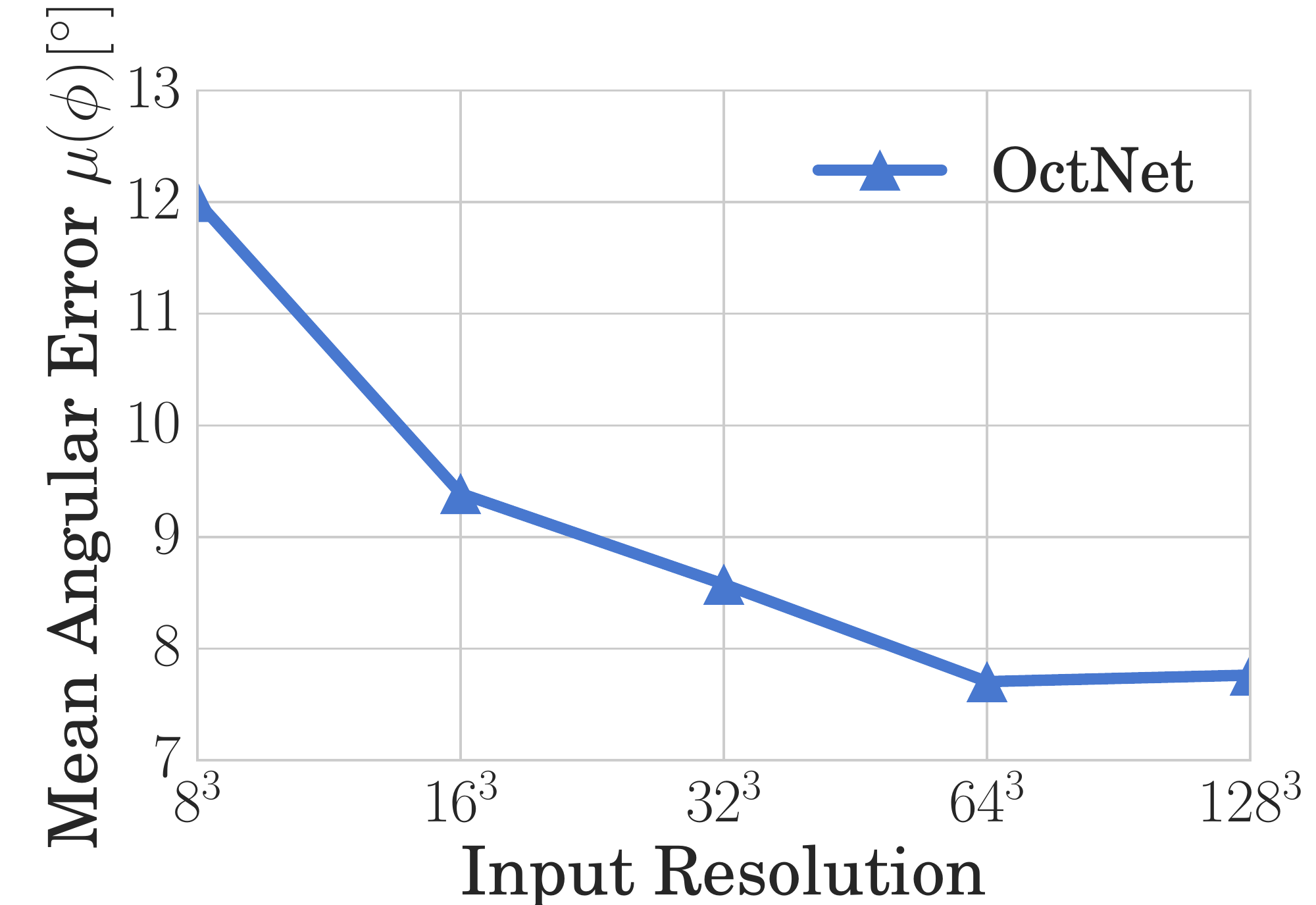}
  \end{subfigure}
  \begin{subfigure}[b]{0.49\linewidth}
    \includegraphics[width=\linewidth]{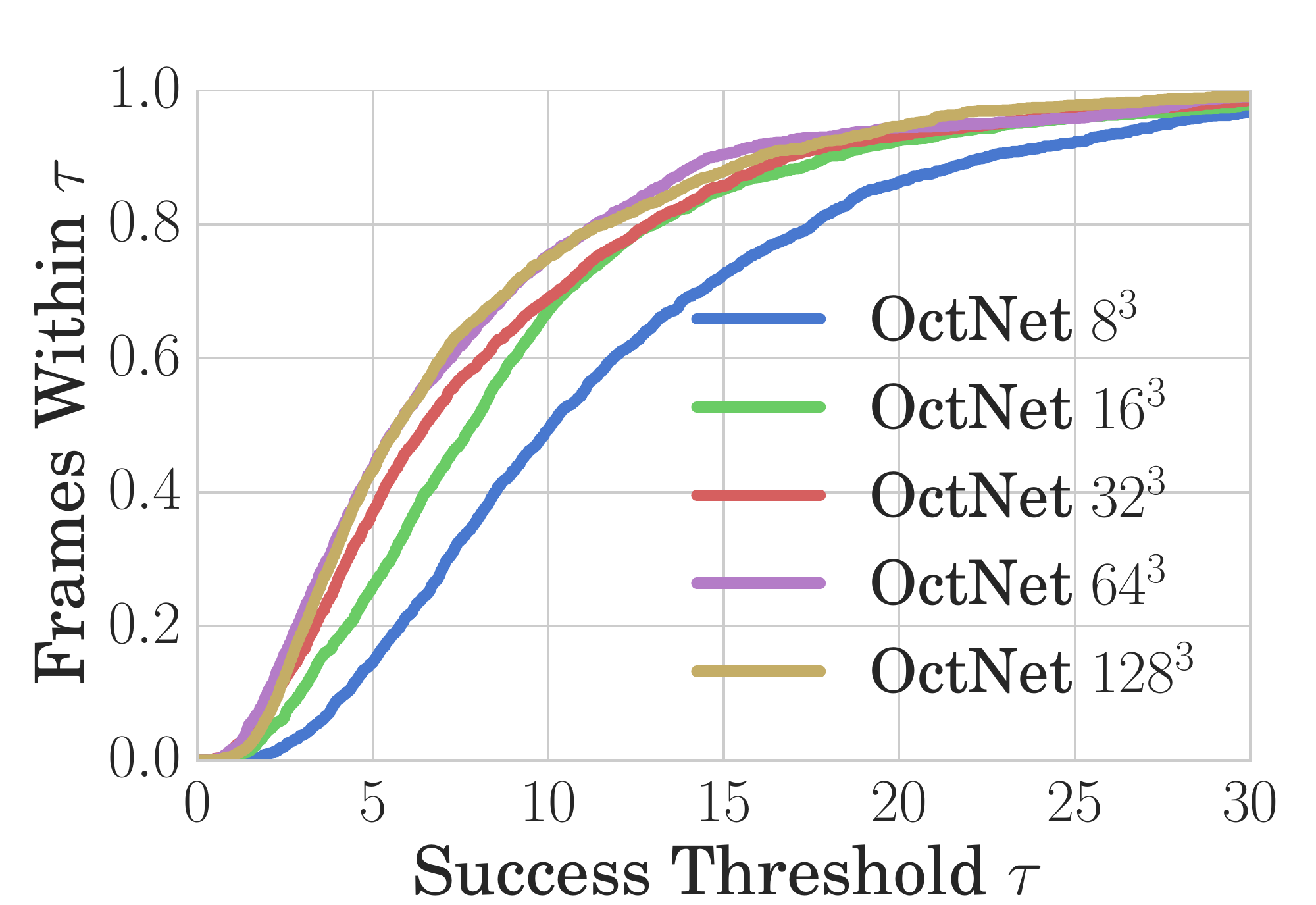}
  \end{subfigure}
  \caption{{\bf Head Pose Results.}}
  \label{fig:headpose}
\end{figure}
\end{minipage}

\subsection{3D Semantic Segmentation}
In this Section we present additional qualitative results of the semantic 3D point labeling task in \figref{fig:qualitative_varcity_01} to \ref{fig:qualitative_varcity_04}.
In these visualizations we show the color part of the voxelized input, the result of the labeling in the voxel representation, and the result back-projected to the 3D point cloud for different houses in the test set of~\cite{Riemenschneider2014ECCV}.

\newcommand{\FigQualResVarcity}[1]{
\begin{figure*}
  \center
  \rotatebox{90}{\hspace{1.3cm} \small{$256^3$} \hspace{1.6cm} \small{$128^3$} \hspace{1.5cm} \small{$64^3$}}~
  \begin{subfigure}[b]{0.24\linewidth}
    \includegraphics[width=\linewidth]{semantic/varcity_add/resized/#1_64rgbin.png}
    \includegraphics[width=\linewidth]{semantic/varcity_add/resized/#1_128rgbin.png}
    \includegraphics[width=\linewidth]{semantic/varcity_add/resized/#1_256rgbin.png}
    \caption{Voxelized Input}
  \end{subfigure}
  \begin{subfigure}[b]{0.24\linewidth}
    \includegraphics[width=\linewidth]{semantic/varcity_add/resized/#1_64es.png}
    \includegraphics[width=\linewidth]{semantic/varcity_add/resized/#1_128es.png}
    \includegraphics[width=\linewidth]{semantic/varcity_add/resized/#1_256es.png}
    \caption{Voxel Estimates}
  \end{subfigure}
  \begin{subfigure}[b]{0.24\linewidth}
     \includegraphics[width=\linewidth]{semantic/varcity_add/resized/#1_64pcles.png}
     \includegraphics[width=\linewidth]{semantic/varcity_add/resized/#1_128pcles.png}
     \includegraphics[width=\linewidth]{semantic/varcity_add/resized/#1_256pcles.png}
    \caption{Estimated Point Cloud}
  \end{subfigure}
  \begin{subfigure}[b]{0.24\linewidth}
     \includegraphics[width=\linewidth]{semantic/varcity_add/resized/#1_pclgt.png}
     \includegraphics[width=\linewidth]{semantic/varcity_add/resized/#1_pclgt.png}
     \includegraphics[width=\linewidth]{semantic/varcity_add/resized/#1_pclgt.png}
     \caption{Ground Truth Point Cloud}
  \end{subfigure}
  \caption{{\bf Facade Labeling Results.} Zoom in for details.}
  \label{fig:qualitative_varcity_#1}
\end{figure*}
}

\FigQualResVarcity{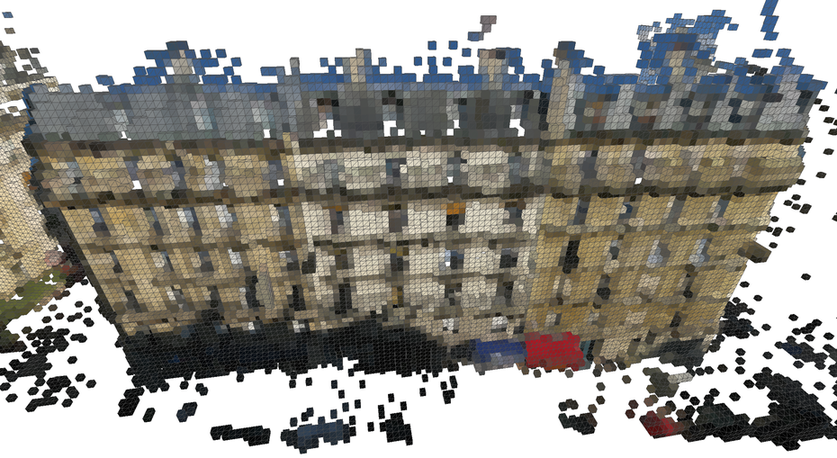}
\FigQualResVarcity{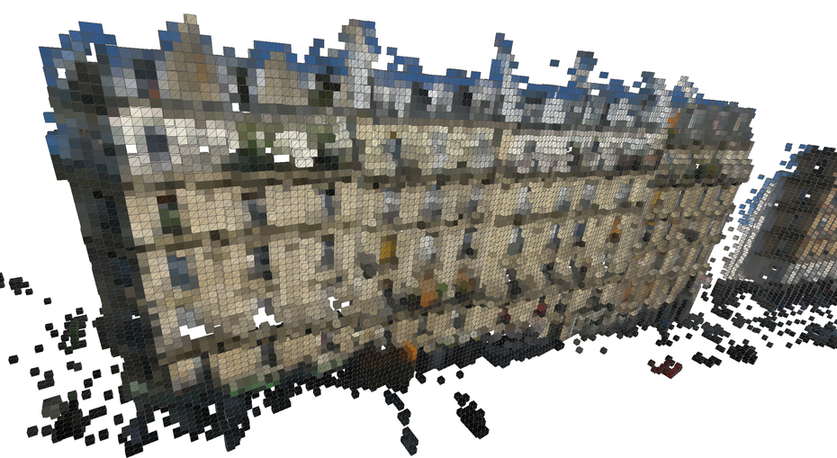}
\FigQualResVarcity{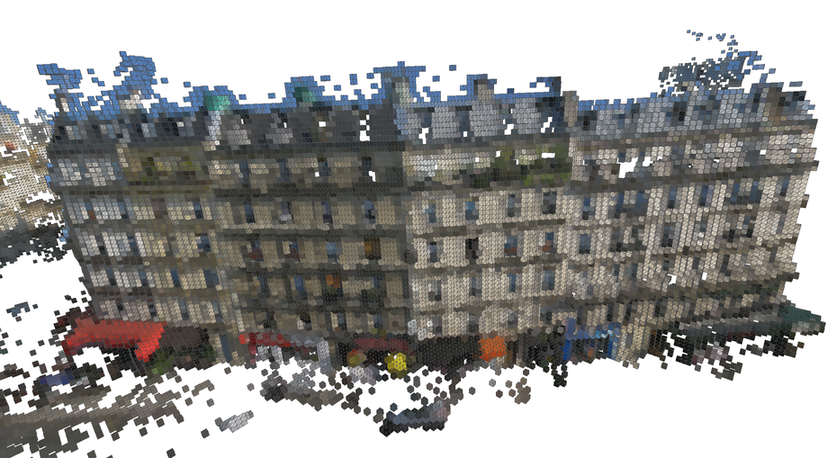}

\suppsection{Network Architecture Details}
\label{sec:network_architectures}
In this Section we detail the network architectures used throughout our experimental evaluations.
We use the following notation for brevity:
$\textrm{conv}(x,y)$ denotes a $3^3$ convolutional layer with $x$ input feature maps and $y$ output feature maps.
Similarly, $\textrm{maxpool}(f)$ is a max-pooling operation that decreases dimensionality by a factor of $f$ along each axis.
All convolutional and fully-connected layers, except the very last one, are followed by a ReLU as activation function.

In the first two experiments, 3D classification and 3D orientation estimation, we show two different classes of architectures. 
In the first one we keep the number of convolution layers per block fixed and add blocks depending on the input resolution of the network. 
We call those networks OctNet1, OctNet2, and OctNet3, depending on the number of convolution layers per block.
Therefore, the number of parameters increases along with the input resolution. 
The detailed architectures are depicted in Table \ref{tab:na_classification_ocnet1}, \ref{tab:na_classification_ocnet2}, and \ref{tab:na_classification_ocnet3} for the classification task and in Table \ref{tab:na_orientation_ocnet1}, \ref{tab:na_orientation_ocnet2}, and \ref{tab:na_orientation_ocnet3} for the orientation estimation tasks, respectively.
Second, we trained network architectures where we keep the number of parameters fixed, independently of the input resolution.
The detailed network architectures for those experiments are presented in Table~\ref{tab:na_classification_fw} and \ref{tab:na_orientation_fw}.

Finally, for semantic 3D point labeling, we use the U-Net type architecture~\cite{Badrinarayanan2015ARXIV,Cicek2016ARXIV} shown in \tabref{tab:na_semantic}.
We use a concatenation layer $\textrm{concat}(\cdot, \cdot)$ to combine the outputs from the decoder and encoder parts of the networks to preserve details.

\begin{table*}
  \center
  {\small
  \begin{tabular}{| c | c | c | c | c | c |}
    \hline 
    $\mathbf{8^3}$           & $\mathbf{16^3}$       & $\mathbf{32^3}$        & $\mathbf{64^3}$        & $\mathbf{128^3}$       & $\mathbf{256^3}$        \\  \hline
    $\textrm{conv}(1,8)$     & $\textrm{conv}(1,8)$  & $\textrm{conv}(1,8)$   & $\textrm{conv}(1,8)$   & $\textrm{conv}(1,8)$   & $\textrm{conv}(1,8)$    \\
                             & $\textrm{maxpool}(2)$ & $\textrm{maxpool}(2)$  & $\textrm{maxpool}(2)$  & $\textrm{maxpool}(2)$  & $\textrm{maxpool}(2)$   \\ \hline
                             & $\textrm{conv}(8,16)$ & $\textrm{conv}(8,16)$  & $\textrm{conv}(8,16)$  & $\textrm{conv}(8,16)$  & $\textrm{conv}(8,16)$   \\
                             &                       & $\textrm{maxpool}(2)$  & $\textrm{maxpool}(2)$  & $\textrm{maxpool}(2)$  & $\textrm{maxpool}(2)$   \\ \hline
                             &                       & $\textrm{conv}(16,24)$ & $\textrm{conv}(16,24)$ & $\textrm{conv}(16,24)$ & $\textrm{conv}(16,24)$   \\
                             &                       &                        & $\textrm{maxpool}(2)$  & $\textrm{maxpool}(2)$  & $\textrm{maxpool}(2)$   \\ \hline
                             &                       &                        & $\textrm{conv}(24,32)$ & $\textrm{conv}(24,32)$ & $\textrm{conv}(24,32)$   \\
                             &                       &                        &                        & $\textrm{maxpool}(2)$  & $\textrm{maxpool}(2)$   \\ \hline
                             &                       &                        &                        & $\textrm{conv}(32,40)$ & $\textrm{conv}(32,40)$   \\
                             &                       &                        &                        &                        & $\textrm{maxpool}(2)$   \\ \hline
                             &                       &                        &                        &                        & $\textrm{conv}(40,48)$   \\ \hline
    \multicolumn{6}{|c|}{$\textrm{Dropout}(0.5)$} \\ \hline
    \multicolumn{6}{|c|}{$\textrm{fully-connected}(1024)$} \\ \hline
    \multicolumn{6}{|c|}{$\textrm{fully-connected}(10)$} \\ \hline
    \multicolumn{6}{|c|}{$\textrm{SoftMax}$} \\ \hline
  \end{tabular}
  }
  \caption{{\bf Network Architectures ModelNet10 Classification: OctNet1}}
  \label{tab:na_classification_ocnet1}
\end{table*}

\begin{table*}
  \center
  {\small
  \begin{tabular}{| c | c | c | c | c | c |}
    \hline 
    $\mathbf{8^3}$          & $\mathbf{16^3}$        & $\mathbf{32^3}$        & $\mathbf{64^3}$        & $\mathbf{128^3}$       & $\mathbf{256^3}$        \\  \hline
    $\textrm{conv}(1,8)$    & $\textrm{conv}(1,8)$   & $\textrm{conv}(1,8)$   & $\textrm{conv}(1,8)$   & $\textrm{conv}(1,8)$   & $\textrm{conv}(1,8)$   \\
    $\textrm{conv}(8,8)$    & $\textrm{conv}(8,8)$   & $\textrm{conv}(8,8)$   & $\textrm{conv}(8,8)$   & $\textrm{conv}(8,8)$   & $\textrm{conv}(8,8)$   \\
                            & $\textrm{maxpool}(2)$  & $\textrm{maxpool}(2)$  & $\textrm{maxpool}(2)$  & $\textrm{maxpool}(2)$  & $\textrm{maxpool}(2)$  \\ \hline
                            & $\textrm{conv}(8,16)$  & $\textrm{conv}(8,16)$  & $\textrm{conv}(8,16)$  & $\textrm{conv}(8,16)$  & $\textrm{conv}(8,16)$  \\
                            & $\textrm{conv}(16,16)$ & $\textrm{conv}(16,16)$ & $\textrm{conv}(16,16)$ & $\textrm{conv}(16,16)$ & $\textrm{conv}(16,16)$  \\
                            &                        & $\textrm{maxpool}(2)$  & $\textrm{maxpool}(2)$  & $\textrm{maxpool}(2)$  & $\textrm{maxpool}(2)$  \\ \hline
                            &                        & $\textrm{conv}(16,24)$ & $\textrm{conv}(16,24)$ & $\textrm{conv}(16,24)$ & $\textrm{conv}(16,24)$  \\
                            &                        & $\textrm{conv}(24,24)$ & $\textrm{conv}(24,24)$ & $\textrm{conv}(24,24)$ & $\textrm{conv}(24,24)$  \\
                            &                        &                        & $\textrm{maxpool}(2)$  & $\textrm{maxpool}(2)$  & $\textrm{maxpool}(2)$  \\ \hline
                            &                        &                        & $\textrm{conv}(24,32)$ & $\textrm{conv}(24,32)$ & $\textrm{conv}(24,32)$  \\
                            &                        &                        & $\textrm{conv}(32,32)$ & $\textrm{conv}(32,32)$ & $\textrm{conv}(32,32)$  \\
                            &                        &                        &                        & $\textrm{maxpool}(2)$  & $\textrm{maxpool}(2)$  \\ \hline
                            &                        &                        &                        & $\textrm{conv}(32,40)$ & $\textrm{conv}(32,40)$  \\
                            &                        &                        &                        & $\textrm{conv}(40,40)$ & $\textrm{conv}(40,40)$  \\
                            &                        &                        &                        &                        & $\textrm{maxpool}(2)$  \\ \hline
                            &                        &                        &                        &                        & $\textrm{conv}(40,48)$  \\
                            &                        &                        &                        &                        & $\textrm{conv}(48,48)$  \\ \hline
    \multicolumn{6}{|c|}{$\textrm{Dropout}(0.5)$} \\ \hline
    \multicolumn{6}{|c|}{$\textrm{fully-connected}(1024)$} \\ \hline
    \multicolumn{6}{|c|}{$\textrm{fully-connected}(10)$} \\ \hline
    \multicolumn{6}{|c|}{$\textrm{SoftMax}$} \\ \hline
  \end{tabular}
  }
  \caption{{\bf Network Architectures ModelNet10 Classification: OctNet2}}
  \label{tab:na_classification_ocnet2}
\end{table*}

\begin{table*}
  \center
  {\small
  \begin{tabular}{| c | c | c | c | c | c |}
    \hline 
    $\mathbf{8^3}$         & $\mathbf{16^3}$        & $\mathbf{32^3}$        & $\mathbf{64^3}$        & $\mathbf{128^3}$       & $\mathbf{256^3}$        \\  \hline
    $\textrm{conv}(1,8)$   & $\textrm{conv}(1,8)$   & $\textrm{conv}(1,8)$   & $\textrm{conv}(1,8)$   & $\textrm{conv}(1,8)$   & $\textrm{conv}(1,8)$  \\
    $\textrm{conv}(8,8)$   & $\textrm{conv}(8,8)$   & $\textrm{conv}(8,8)$   & $\textrm{conv}(8,8)$   & $\textrm{conv}(8,8)$   & $\textrm{conv}(8,8)$  \\
    $\textrm{conv}(8,8)$   & $\textrm{conv}(8,8)$   & $\textrm{conv}(8,8)$   & $\textrm{conv}(8,8)$   & $\textrm{conv}(8,8)$   & $\textrm{conv}(8,8)$  \\
                           & $\textrm{maxpool}(2)$  & $\textrm{maxpool}(2)$  & $\textrm{maxpool}(2)$  & $\textrm{maxpool}(2)$  & $\textrm{maxpool}(2)$ \\ \hline
                           & $\textrm{conv}(8,16)$  & $\textrm{conv}(8,16)$  & $\textrm{conv}(8,16)$  & $\textrm{conv}(8,16)$  & $\textrm{conv}(8,16)$ \\
                           & $\textrm{conv}(16,16)$ & $\textrm{conv}(16,16)$ & $\textrm{conv}(16,16)$ & $\textrm{conv}(16,16)$ & $\textrm{conv}(16,16)$ \\
                           & $\textrm{conv}(16,16)$ & $\textrm{conv}(16,16)$ & $\textrm{conv}(16,16)$ & $\textrm{conv}(16,16)$ & $\textrm{conv}(16,16)$ \\
                           &                        & $\textrm{maxpool}(2)$  & $\textrm{maxpool}(2)$  & $\textrm{maxpool}(2)$  & $\textrm{maxpool}(2)$ \\ \hline
                           &                        & $\textrm{conv}(16,24)$ & $\textrm{conv}(16,24)$ & $\textrm{conv}(16,24)$ & $\textrm{conv}(16,24)$ \\
                           &                        & $\textrm{conv}(24,24)$ & $\textrm{conv}(24,24)$ & $\textrm{conv}(24,24)$ & $\textrm{conv}(24,24)$ \\
                           &                        & $\textrm{conv}(24,24)$ & $\textrm{conv}(24,24)$ & $\textrm{conv}(24,24)$ & $\textrm{conv}(24,24)$ \\
                           &                        &                        & $\textrm{maxpool}(2)$  & $\textrm{maxpool}(2)$  & $\textrm{maxpool}(2)$ \\ \hline
                           &                        &                        & $\textrm{conv}(24,32)$ & $\textrm{conv}(24,32)$ & $\textrm{conv}(24,32)$ \\
                           &                        &                        & $\textrm{conv}(32,32)$ & $\textrm{conv}(32,32)$ & $\textrm{conv}(32,32)$ \\
                           &                        &                        & $\textrm{conv}(32,32)$ & $\textrm{conv}(32,32)$ & $\textrm{conv}(32,32)$ \\
                           &                        &                        &                        & $\textrm{maxpool}(2)$  & $\textrm{maxpool}(2)$ \\ \hline
                           &                        &                        &                        & $\textrm{conv}(32,40)$ & $\textrm{conv}(32,40)$ \\
                           &                        &                        &                        & $\textrm{conv}(40,40)$ & $\textrm{conv}(40,40)$ \\
                           &                        &                        &                        & $\textrm{conv}(40,40)$ & $\textrm{conv}(40,40)$ \\
                           &                        &                        &                        &                        & $\textrm{maxpool}(2)$ \\ \hline
                           &                        &                        &                        &                        & $\textrm{conv}(40,48)$ \\
                           &                        &                        &                        &                        & $\textrm{conv}(48,48)$ \\
                           &                        &                        &                        &                        & $\textrm{conv}(48,48)$ \\ \hline
    \multicolumn{6}{|c|}{$\textrm{Dropout}(0.5)$} \\ \hline
    \multicolumn{6}{|c|}{$\textrm{fully-connected}(1024)$} \\ \hline
    \multicolumn{6}{|c|}{$\textrm{fully-connected}(10)$} \\ \hline
    \multicolumn{6}{|c|}{$\textrm{SoftMax}$} \\ \hline
  \end{tabular}
  }
  \caption{{\bf Network Architectures ModelNet10 Classification: OctNet3.}}
  \label{tab:na_classification_ocnet3}
\end{table*}

\begin{table*}
  \center
  {\small
  \begin{tabular}{| c | c | c | c | c | c |}
    \hline
    $\mathbf{8^3}$         & $\mathbf{16^3}$        & $\mathbf{32^3}$        & $\mathbf{64^3}$        & $\mathbf{128^3}$       & $\mathbf{256^3}$      \\ \hline
    $\textrm{conv}(1,8)$   & $\textrm{conv}(1,8)$   & $\textrm{conv}(1,8)$   & $\textrm{conv}(1,8)$   & $\textrm{conv}(1,8)$   & $\textrm{conv}(1,8)$  \\
    $\textrm{conv}(8,14)$  & $\textrm{conv}(8,14)$  & $\textrm{conv}(8,14)$  & $\textrm{conv}(8,14)$  & $\textrm{conv}(8,14)$  & $\textrm{conv}(8,14)$ \\
                           &                        &                        &                        &                        & $\textrm{maxpool}(2)$   \\ \hline
    $\textrm{conv}(14,14)$ & $\textrm{conv}(14,14)$ & $\textrm{conv}(14,14)$ & $\textrm{conv}(14,14)$ & $\textrm{conv}(14,14)$ & $\textrm{conv}(14,14)$ \\
    $\textrm{conv}(14,20)$ & $\textrm{conv}(14,20)$ & $\textrm{conv}(14,20)$ & $\textrm{conv}(14,20)$ & $\textrm{conv}(14,20)$ & $\textrm{conv}(14,20)$ \\
                           &                        &                        &                        & $\textrm{maxpool}(2)$  & $\textrm{maxpool}(2)$   \\ \hline
    $\textrm{conv}(20,20)$ & $\textrm{conv}(20,20)$ & $\textrm{conv}(20,20)$ & $\textrm{conv}(20,20)$ & $\textrm{conv}(20,20)$ & $\textrm{conv}(20,20)$ \\
    $\textrm{conv}(20,26)$ & $\textrm{conv}(20,26)$ & $\textrm{conv}(20,26)$ & $\textrm{conv}(20,26)$ & $\textrm{conv}(20,26)$ & $\textrm{conv}(20,26)$ \\
                           &                        &                        & $\textrm{maxpool}(2)$  & $\textrm{maxpool}(2)$  & $\textrm{maxpool}(2)$   \\ \hline
    $\textrm{conv}(26,26)$ & $\textrm{conv}(26,26)$ & $\textrm{conv}(26,26)$ & $\textrm{conv}(26,26)$ & $\textrm{conv}(26,26)$ & $\textrm{conv}(26,26)$ \\
    $\textrm{conv}(26,32)$ & $\textrm{conv}(26,32)$ & $\textrm{conv}(26,32)$ & $\textrm{conv}(26,32)$ & $\textrm{conv}(26,32)$ & $\textrm{conv}(26,32)$ \\
                           &                        & $\textrm{maxpool}(2)$  & $\textrm{maxpool}(2)$  & $\textrm{maxpool}(2)$  & $\textrm{maxpool}(2)$   \\ \hline
    $\textrm{conv}(32,32)$ & $\textrm{conv}(32,32)$ & $\textrm{conv}(32,32)$ & $\textrm{conv}(32,32)$ & $\textrm{conv}(32,32)$ & $\textrm{conv}(32,32)$ \\
    $\textrm{conv}(32,32)$ & $\textrm{conv}(32,32)$ & $\textrm{conv}(32,32)$ & $\textrm{conv}(32,32)$ & $\textrm{conv}(32,32)$ & $\textrm{conv}(32,32)$ \\
                           & $\textrm{maxpool}(2)$  & $\textrm{maxpool}(2)$  & $\textrm{maxpool}(2)$  & $\textrm{maxpool}(2)$  & $\textrm{maxpool}(2)$   \\ \hline
    \multicolumn{6}{|c|}{$\textrm{Dropout}(0.5)$} \\ \hline
    \multicolumn{6}{|c|}{$\textrm{fully-connected}(512)$} \\ \hline
    \multicolumn{6}{|c|}{$\textrm{fully-connected}(10)$} \\ \hline
    \multicolumn{6}{|c|}{$\textrm{SoftMax}$} \\ \hline
  \end{tabular}
  }
  \caption{{\bf Network Architectures ModelNet10 Classification.}}
  \label{tab:na_classification_fw}
\end{table*}

\begin{table*}
  \center
  {\small
  \begin{tabular}{| c | c | c | c | c | c |}
    \hline 
    $\mathbf{8^3}$           & $\mathbf{16^3}$       & $\mathbf{32^3}$        & $\mathbf{64^3}$        & $\mathbf{128^3}$       & $\mathbf{256^3}$        \\  \hline
    $\textrm{conv}(1,8)$     & $\textrm{conv}(1,8)$  & $\textrm{conv}(1,8)$   & $\textrm{conv}(1,8)$   & $\textrm{conv}(1,8)$   & $\textrm{conv}(1,8)$    \\
                             & $\textrm{maxpool}(2)$ & $\textrm{maxpool}(2)$  & $\textrm{maxpool}(2)$  & $\textrm{maxpool}(2)$  & $\textrm{maxpool}(2)$   \\ \hline
                             & $\textrm{conv}(8,16)$ & $\textrm{conv}(8,16)$  & $\textrm{conv}(8,16)$  & $\textrm{conv}(8,16)$  & $\textrm{conv}(8,16)$   \\
                             &                       & $\textrm{maxpool}(2)$  & $\textrm{maxpool}(2)$  & $\textrm{maxpool}(2)$  & $\textrm{maxpool}(2)$   \\ \hline
                             &                       & $\textrm{conv}(16,24)$ & $\textrm{conv}(16,24)$ & $\textrm{conv}(16,24)$ & $\textrm{conv}(16,24)$   \\
                             &                       &                        & $\textrm{maxpool}(2)$  & $\textrm{maxpool}(2)$  & $\textrm{maxpool}(2)$   \\ \hline
                             &                       &                        & $\textrm{conv}(24,32)$ & $\textrm{conv}(24,32)$ & $\textrm{conv}(24,32)$   \\
                             &                       &                        &                        & $\textrm{maxpool}(2)$  & $\textrm{maxpool}(2)$   \\ \hline
                             &                       &                        &                        & $\textrm{conv}(32,40)$ & $\textrm{conv}(32,40)$   \\
                             &                       &                        &                        &                        & $\textrm{maxpool}(2)$   \\ \hline
                             &                       &                        &                        &                        & $\textrm{conv}(40,48)$   \\ \hline
    \multicolumn{6}{|c|}{$\textrm{Dropout}(0.5)$} \\ \hline
    \multicolumn{6}{|c|}{$\textrm{fully-connected}(1024)$} \\ \hline
    \multicolumn{6}{|c|}{$\textrm{fully-connected}(4)$} \\ \hline
    \multicolumn{6}{|c|}{$\textrm{Normalize}$} \\ \hline
  \end{tabular}
  }
  \caption{{\bf Network Architectures Orientation Estimation: OctNet1}}
  \label{tab:na_orientation_ocnet1}
\end{table*}

\begin{table*}
  \center
  {\small
  \begin{tabular}{| c | c | c | c | c | c |}
    \hline 
    $\mathbf{8^3}$          & $\mathbf{16^3}$        & $\mathbf{32^3}$        & $\mathbf{64^3}$        & $\mathbf{128^3}$       & $\mathbf{256^3}$        \\  \hline
    $\textrm{conv}(1,8)$    & $\textrm{conv}(1,8)$   & $\textrm{conv}(1,8)$   & $\textrm{conv}(1,8)$   & $\textrm{conv}(1,8)$   & $\textrm{conv}(1,8)$   \\
    $\textrm{conv}(8,8)$    & $\textrm{conv}(8,8)$   & $\textrm{conv}(8,8)$   & $\textrm{conv}(8,8)$   & $\textrm{conv}(8,8)$   & $\textrm{conv}(8,8)$   \\
                            & $\textrm{maxpool}(2)$  & $\textrm{maxpool}(2)$  & $\textrm{maxpool}(2)$  & $\textrm{maxpool}(2)$  & $\textrm{maxpool}(2)$  \\ \hline
                            & $\textrm{conv}(8,16)$  & $\textrm{conv}(8,16)$  & $\textrm{conv}(8,16)$  & $\textrm{conv}(8,16)$  & $\textrm{conv}(8,16)$  \\
                            & $\textrm{conv}(16,16)$ & $\textrm{conv}(16,16)$ & $\textrm{conv}(16,16)$ & $\textrm{conv}(16,16)$ & $\textrm{conv}(16,16)$  \\
                            &                        & $\textrm{maxpool}(2)$  & $\textrm{maxpool}(2)$  & $\textrm{maxpool}(2)$  & $\textrm{maxpool}(2)$  \\ \hline
                            &                        & $\textrm{conv}(16,24)$ & $\textrm{conv}(16,24)$ & $\textrm{conv}(16,24)$ & $\textrm{conv}(16,24)$  \\
                            &                        & $\textrm{conv}(24,24)$ & $\textrm{conv}(24,24)$ & $\textrm{conv}(24,24)$ & $\textrm{conv}(24,24)$  \\
                            &                        &                        & $\textrm{maxpool}(2)$  & $\textrm{maxpool}(2)$  & $\textrm{maxpool}(2)$  \\ \hline
                            &                        &                        & $\textrm{conv}(24,32)$ & $\textrm{conv}(24,32)$ & $\textrm{conv}(24,32)$  \\
                            &                        &                        & $\textrm{conv}(32,32)$ & $\textrm{conv}(32,32)$ & $\textrm{conv}(32,32)$  \\
                            &                        &                        &                        & $\textrm{maxpool}(2)$  & $\textrm{maxpool}(2)$  \\ \hline
                            &                        &                        &                        & $\textrm{conv}(32,40)$ & $\textrm{conv}(32,40)$  \\
                            &                        &                        &                        & $\textrm{conv}(40,40)$ & $\textrm{conv}(40,40)$  \\
                            &                        &                        &                        &                        & $\textrm{maxpool}(2)$  \\ \hline
                            &                        &                        &                        &                        & $\textrm{conv}(40,48)$  \\
                            &                        &                        &                        &                        & $\textrm{conv}(48,48)$  \\ \hline
    \multicolumn{6}{|c|}{$\textrm{Dropout}(0.5)$} \\ \hline
    \multicolumn{6}{|c|}{$\textrm{fully-connected}(1024)$} \\ \hline
    \multicolumn{6}{|c|}{$\textrm{fully-connected}(4)$} \\ \hline
    \multicolumn{6}{|c|}{$\textrm{Normalize}$} \\ \hline
  \end{tabular}
  }
  \caption{{\bf Network Architectures Orientation Estimation: OctNet2}}
  \label{tab:na_orientation_ocnet2}
\end{table*}

\begin{table*}
  \center
  {\small
  \begin{tabular}{| c | c | c | c | c | c |}
    \hline 
    $\mathbf{8^3}$         & $\mathbf{16^3}$        & $\mathbf{32^3}$        & $\mathbf{64^3}$        & $\mathbf{128^3}$       & $\mathbf{256^3}$        \\  \hline
    $\textrm{conv}(1,8)$   & $\textrm{conv}(1,8)$   & $\textrm{conv}(1,8)$   & $\textrm{conv}(1,8)$   & $\textrm{conv}(1,8)$   & $\textrm{conv}(1,8)$  \\
    $\textrm{conv}(8,8)$   & $\textrm{conv}(8,8)$   & $\textrm{conv}(8,8)$   & $\textrm{conv}(8,8)$   & $\textrm{conv}(8,8)$   & $\textrm{conv}(8,8)$  \\
    $\textrm{conv}(8,8)$   & $\textrm{conv}(8,8)$   & $\textrm{conv}(8,8)$   & $\textrm{conv}(8,8)$   & $\textrm{conv}(8,8)$   & $\textrm{conv}(8,8)$  \\
                           & $\textrm{maxpool}(2)$  & $\textrm{maxpool}(2)$  & $\textrm{maxpool}(2)$  & $\textrm{maxpool}(2)$  & $\textrm{maxpool}(2)$ \\ \hline
                           & $\textrm{conv}(8,16)$  & $\textrm{conv}(8,16)$  & $\textrm{conv}(8,16)$  & $\textrm{conv}(8,16)$  & $\textrm{conv}(8,16)$ \\
                           & $\textrm{conv}(16,16)$ & $\textrm{conv}(16,16)$ & $\textrm{conv}(16,16)$ & $\textrm{conv}(16,16)$ & $\textrm{conv}(16,16)$ \\
                           & $\textrm{conv}(16,16)$ & $\textrm{conv}(16,16)$ & $\textrm{conv}(16,16)$ & $\textrm{conv}(16,16)$ & $\textrm{conv}(16,16)$ \\
                           &                        & $\textrm{maxpool}(2)$  & $\textrm{maxpool}(2)$  & $\textrm{maxpool}(2)$  & $\textrm{maxpool}(2)$ \\ \hline
                           &                        & $\textrm{conv}(16,24)$ & $\textrm{conv}(16,24)$ & $\textrm{conv}(16,24)$ & $\textrm{conv}(16,24)$ \\
                           &                        & $\textrm{conv}(24,24)$ & $\textrm{conv}(24,24)$ & $\textrm{conv}(24,24)$ & $\textrm{conv}(24,24)$ \\
                           &                        & $\textrm{conv}(24,24)$ & $\textrm{conv}(24,24)$ & $\textrm{conv}(24,24)$ & $\textrm{conv}(24,24)$ \\
                           &                        &                        & $\textrm{maxpool}(2)$  & $\textrm{maxpool}(2)$  & $\textrm{maxpool}(2)$ \\ \hline
                           &                        &                        & $\textrm{conv}(24,32)$ & $\textrm{conv}(24,32)$ & $\textrm{conv}(24,32)$ \\
                           &                        &                        & $\textrm{conv}(32,32)$ & $\textrm{conv}(32,32)$ & $\textrm{conv}(32,32)$ \\
                           &                        &                        & $\textrm{conv}(32,32)$ & $\textrm{conv}(32,32)$ & $\textrm{conv}(32,32)$ \\
                           &                        &                        &                        & $\textrm{maxpool}(2)$  & $\textrm{maxpool}(2)$ \\ \hline
                           &                        &                        &                        & $\textrm{conv}(32,40)$ & $\textrm{conv}(32,40)$ \\
                           &                        &                        &                        & $\textrm{conv}(40,40)$ & $\textrm{conv}(40,40)$ \\
                           &                        &                        &                        & $\textrm{conv}(40,40)$ & $\textrm{conv}(40,40)$ \\
                           &                        &                        &                        &                        & $\textrm{maxpool}(2)$ \\ \hline
                           &                        &                        &                        &                        & $\textrm{conv}(40,48)$ \\
                           &                        &                        &                        &                        & $\textrm{conv}(48,48)$ \\
                           &                        &                        &                        &                        & $\textrm{conv}(48,48)$ \\ \hline
    \multicolumn{6}{|c|}{$\textrm{Dropout}(0.5)$} \\ \hline
    \multicolumn{6}{|c|}{$\textrm{fully-connected}(1024)$} \\ \hline
    \multicolumn{6}{|c|}{$\textrm{fully-connected}(4)$} \\ \hline
    \multicolumn{6}{|c|}{$\textrm{Normalize}$} \\ \hline
  \end{tabular}
  }
  \caption{{\bf Network Architectures Orientation Estimation: OctNet3.}}
  \label{tab:na_orientation_ocnet3}
\end{table*}

\begin{table*}
  \center
  {\small
  \begin{tabular}{| c | c | c | c | c | c |}
    \hline
    $\mathbf{8^3}$         & $\mathbf{16^3}$        & $\mathbf{32^3}$        & $\mathbf{64^3}$        & $\mathbf{128^3}$       & $\mathbf{256^3}$      \\ \hline
    $\textrm{conv}(1,8)$   & $\textrm{conv}(1,8)$   & $\textrm{conv}(1,8)$   & $\textrm{conv}(1,8)$   & $\textrm{conv}(1,8)$   & $\textrm{conv}(1,8)$  \\
    $\textrm{conv}(8,14)$  & $\textrm{conv}(8,14)$  & $\textrm{conv}(8,14)$  & $\textrm{conv}(8,14)$  & $\textrm{conv}(8,14)$  & $\textrm{conv}(8,14)$ \\
                           &                        &                        &                        &                        & $\textrm{maxpool}(2)$   \\ \hline
    $\textrm{conv}(14,14)$ & $\textrm{conv}(14,14)$ & $\textrm{conv}(14,14)$ & $\textrm{conv}(14,14)$ & $\textrm{conv}(14,14)$ & $\textrm{conv}(14,14)$ \\
    $\textrm{conv}(14,20)$ & $\textrm{conv}(14,20)$ & $\textrm{conv}(14,20)$ & $\textrm{conv}(14,20)$ & $\textrm{conv}(14,20)$ & $\textrm{conv}(14,20)$ \\
                           &                        &                        &                        & $\textrm{maxpool}(2)$  & $\textrm{maxpool}(2)$   \\ \hline
    $\textrm{conv}(20,20)$ & $\textrm{conv}(20,20)$ & $\textrm{conv}(20,20)$ & $\textrm{conv}(20,20)$ & $\textrm{conv}(20,20)$ & $\textrm{conv}(20,20)$ \\
    $\textrm{conv}(20,26)$ & $\textrm{conv}(20,26)$ & $\textrm{conv}(20,26)$ & $\textrm{conv}(20,26)$ & $\textrm{conv}(20,26)$ & $\textrm{conv}(20,26)$ \\
                           &                        &                        & $\textrm{maxpool}(2)$  & $\textrm{maxpool}(2)$  & $\textrm{maxpool}(2)$   \\ \hline
    $\textrm{conv}(26,26)$ & $\textrm{conv}(26,26)$ & $\textrm{conv}(26,26)$ & $\textrm{conv}(26,26)$ & $\textrm{conv}(26,26)$ & $\textrm{conv}(26,26)$ \\
    $\textrm{conv}(26,32)$ & $\textrm{conv}(26,32)$ & $\textrm{conv}(26,32)$ & $\textrm{conv}(26,32)$ & $\textrm{conv}(26,32)$ & $\textrm{conv}(26,32)$ \\
                           &                        & $\textrm{maxpool}(2)$  & $\textrm{maxpool}(2)$  & $\textrm{maxpool}(2)$  & $\textrm{maxpool}(2)$   \\ \hline
    $\textrm{conv}(32,32)$ & $\textrm{conv}(32,32)$ & $\textrm{conv}(32,32)$ & $\textrm{conv}(32,32)$ & $\textrm{conv}(32,32)$ & $\textrm{conv}(32,32)$ \\
    $\textrm{conv}(32,32)$ & $\textrm{conv}(32,32)$ & $\textrm{conv}(32,32)$ & $\textrm{conv}(32,32)$ & $\textrm{conv}(32,32)$ & $\textrm{conv}(32,32)$ \\
                           & $\textrm{maxpool}(2)$  & $\textrm{maxpool}(2)$  & $\textrm{maxpool}(2)$  & $\textrm{maxpool}(2)$  & $\textrm{maxpool}(2)$   \\ \hline
    \multicolumn{6}{|c|}{$\textrm{Dropout}(0.5)$} \\ \hline
    \multicolumn{6}{|c|}{$\textrm{fully-connected}(512)$} \\ \hline
    \multicolumn{6}{|c|}{$\textrm{fully-connected}(4)$} \\ \hline
    \multicolumn{6}{|c|}{$\textrm{Normalize}$} \\ \hline
  \end{tabular}
  }
  \caption{{\bf Network Architectures Orientation Estimation.}}
  \label{tab:na_orientation_fw}
\end{table*}

\begin{table*}
  \center
  {\small
  \begin{tabular}{| c | c | c}
    \hline
    Output name    & Operation                \\ \hline
                   & $\textrm{conv}(8,8)$     \\
    Enc1           & $\textrm{conv}(8,16)$    \\ 
                   & $\textrm{maxpool}(2)$    \\ \hline
                   & $\textrm{conv}(16,16)$   \\ 
    Enc2           & $\textrm{conv}(16,32)$   \\ 
                   & $\textrm{maxpool}(2)$    \\ \hline
                   & $\textrm{conv}(32,32)$   \\ 
    Enc3           & $\textrm{conv}(32,64)$   \\ 
                   & $\textrm{maxpool}(2)$    \\ \hline
                   & $\textrm{conv}(64,64)$   \\ 
    Enc4           & $\textrm{conv}(64,128)$  \\ 
                   & $\textrm{maxpool}(2)$    \\ \hline
                   & $\textrm{conv}(128,128)$ \\ 
                   & $\textrm{conv}(128,128)$ \\ 
                   & $\textrm{conv}(128,128)$ \\ 
    Dec4           & $\textrm{unpool}(2)$     \\ \hline
                   & $\textrm{concat}(\textrm{Enc4,Dec4})$     \\
                   & $\textrm{conv}(256,128)$ \\ 
                   & $\textrm{conv}(128,64)$  \\ 
    Dec3           & $\textrm{unpool}(2)$     \\ \hline
                   & $\textrm{concat}(\textrm{Enc3,Dec3})$     \\
                   & $\textrm{conv}(128,64)$ \\ 
                   & $\textrm{conv}(64,32)$  \\ 
    Dec2           & $\textrm{unpool}(2)$     \\ \hline
                   & $\textrm{concat}(\textrm{Enc2,Dec2})$     \\
                   & $\textrm{conv}(64,32)$ \\ 
                   & $\textrm{conv}(32,16)$  \\ 
    Dec1           & $\textrm{unpool}(2)$     \\ \hline
                   & $\textrm{concat}(\textrm{Enc1,Dec1})$     \\
                   & $\textrm{conv}(32,32)$ \\ 
                   & $\textrm{conv}(32,8)$  \\ 
    Output         & $\textrm{SoftMax}$     \\
    \hline    
  \end{tabular}
  }
  \caption{{\bf Network Architecture Semantic 3D Point Cloud Labeling.}}
  \label{tab:na_semantic}
\end{table*}

\end{document}